\documentclass[preprint,12pt]{elsarticle}
\usepackage{amsmath}
\usepackage{amssymb}
\usepackage{subfigure}
\usepackage{cases}
\usepackage{url}
\usepackage{threeparttable}
\usepackage{booktabs}
\usepackage{longtable}
\usepackage{colortbl}
\usepackage{url}
\graphicspath{{./Figures}}
\usepackage{graphicx}
\usepackage{epstopdf}
\usepackage{algorithmic}
\usepackage[ruled]{algorithm2e}
\usepackage{amssymb,amsmath,amsthm}
\newtheorem{theorem}{Theorem}
\newtheorem{prop}{Proposition}
\usepackage{color}

\usepackage[colorlinks=true,
linkcolor=black,
anchorcolor=black,
citecolor=black,
urlcolor=blue
]{hyperref}

\begin{document}

\newpage

Title: Twin support vector quantile regression\\
\\
Author names and affiliations:\\
\\
Yafen Ye\\
School of Economics, Zhejiang University of Technology, Hangzhou, 310023, P.R.China\\
Institute for Industrial System Modernization, Zhejiang University of Technology, Hangzhou, 310023, P.R.China\\
yafenye@163.com\\
\\
Zhihu Xu\\
School of Economics, Zhejiang University of Technology, Hangzhou, 310023, P.R.China\\
dianxianquan@126.com\\
\\
Jinhua Zhang\\
School of Economics, Zhejiang University of Technology, Hangzhou, 310023, P.R.China\\
celliazjh@zjut.edu.cn\\
\\
Weijie Chen\\
School of Economics, Zhejiang University of Technology, Hangzhou, 310023, P.R.China\\
Zhijiang College, Zhejiang University of Technology, Hangzhou, 310023, P.R.China\\
wjcper2008@126.com\\
\\
Yuanhai Shao\\
Management School, Hainan University, Haikou, 570228, P. R. China\\
shaoyuanhai21@163.com\\
\\
Corresponding Author:\\
Yuanhai Shao\\
Management School, Hainan University, Haikou, 570228, P. R. China\\
shaoyuanhai21@163.com

\begin{frontmatter}

\title{Twin support vector quantile regression}

\author[1,2]{Yafen Ye}
\address[1]{School of Economics, Zhejiang University of Technology, Hangzhou, 310023,
P.R.China}
\address[2]{Institute for Industrial System Modernization, Zhejiang University of Technology, Hangzhou, 310023, P.R.China}

\author[1]{Zhihu Xu}

\author[1]{Jinhua Zhang}

\author[1,3]{Weijie Chen}
\address[3]{Zhijiang College, Zhejiang University of Technology, Hangzhou, 310023,
P.R.China}

\author[4]{Yuanhai Shao \corref{cor1}}
\address[4] {Management School, Hainan University, Haikou, 570228, P. R. China} \cortext[cor1]{Corresponding author:
shaoyuanhai21@163.com}

\begin{abstract}
We propose a twin support vector quantile regression (TSVQR) to capture the heterogeneous and asymmetric information in modern data. Using a quantile parameter, TSVQR effectively depicts the heterogeneous distribution information with respect to all portions of data points. Correspondingly, TSVQR constructs two smaller sized quadratic programming problems (QPPs) to generate two nonparallel planes to measure the distributional asymmetry between the lower and upper bounds at each quantile level. The QPPs in TSVQR are smaller and easier to solve than those in previous quantile regression methods. Moreover, the dual coordinate descent algorithm for TSVQR also accelerates the training speed. Experimental results on six artificial data sets, five benchmark data sets, two large scale data sets, two time-series data sets, and two imbalanced data sets indicate that the TSVQR outperforms previous quantile regression methods in terms of the effectiveness of completely capturing the heterogeneous and asymmetric information and
the efficiency of the learning process.

\end{abstract}


\end{frontmatter}

\section{Introduction}
Heterogeneity, as one of the major statistical features of modern data (\citeauthor{wang2017heterogeneous}, \citeyear{wang2017heterogeneous}),
has commonly emerged in diverse fields, such as economics (\citeauthor{tran2019local}, \citeyear{tran2019local}; \citeauthor{ye2022financial}, \citeyear{ye2022financial}), finance (\citeauthor{dirick2021hierarchical}, \citeyear{dirick2021hierarchical}), environment (\citeauthor{yang2021does}, \citeyear{yang2021does}), genetics (\citeauthor{miller2021characterizing}, \citeyear{miller2021characterizing}), and medicine (\citeauthor{khan2021precision}, \citeyear{khan2021precision}). Moreover, in some real-world applications,
the heterogeneity problem has always been accompanied by an asymmetry problem ( \citeauthor{kolasa2009structural}, \citeyear{kolasa2009structural}; \citeauthor{kundaje2012ubiquitous}, \citeyear{kundaje2012ubiquitous}; \citeauthor{baigh2021existence}, \citeyear{baigh2021existence}).
In the medicine field, for instance, the impact of temperature on COVID-19 in Indian states is heterogeneous and asymmetric (\citeauthor{irfan2022asymmetric}, \citeyear{irfan2022asymmetric}).

Regression technology, as an effective data analysis tool, presents challenges in dealing with the problem of heterogeneity and asymmetry. Ordinary least squares (OLS) regression (\citeauthor{dempster1977simulation}, \citeyear{dempster1977simulation}) is one of the commonly used methods for the estimation of the optimal regression function because its estimators have the smallest variance among the class of linear unbiased estimators.
Unfortunately, least squares regression is a conditional mean model that only reflects the mean of the conditional distribution of data points.
Least squares regression is not suitable for the problem of distributional heterogeneity. To deal with the distributional heterogeneity problem, Koenker and Bassett (\citeyear{koenker1978regression}) proposed quantile regression (QR) which uses a quantile parameter to compute several different regression curves corresponding to various percentage points of the distributions.
As the quantile parameter increases, the estimated lines move up through the data,
retaining in most cases a slope reasonably close to that of the family of true conditional quantile functions. QR offers a comprehensive strategy for obtaining a more complete picture of the data points, and has been an effective model to address the problem of distributional heterogeneity. Therefore, numerous extensions and variants of QR have been proposed (\citeauthor{wang2023generative}, \citeyear{wang2023generative}).

In the field of machine learning, support vector regression (SVR) (\citeauthor{drucker1997support}, \citeyear{drucker1997support}; \citeauthor{burges1998tutorial}, \citeyear{burges1998tutorial}; \citeauthor{wu2022iterative}, \citeyear{wu2022iterative}; \citeauthor{wang2023new}, \citeyear{wang2023new}) in the framework of statistical learning theory is an effective regression model. SVR adopts the $L_2$-norm regularization term for the estimation of the optimal regression function, effectively overcoming the overfitting problem. Takeuchi et al. (\citeyear{takeuchi2006nonparametric}) brought the technique of QR into the SVR framework, and proposed nonparametric quantile regression (NPQR). Subsequently, support vector quantile regression (SVQR) (\citeauthor{li2007quantile}, \citeyear{li2007quantile}; \citeauthor{park2011quantile}, \citeyear{park2011quantile}), support vector machine with generalized quantile loss (RQSVR) (\citeauthor{yang2019robust},\citeyear{yang2019robust}), $\varepsilon$-insensitive support vector quantile regression ($\varepsilon$-SVQR) (\citeauthor{anand2020new}, \citeyear{anand2020new}), and online support vector quantile regression (Online-SVQR) (\citeauthor{ye2021online}, \citeyear{ye2021online}) were proposed. These models follow the spirit of QR and use the quantile parameter to measure the various percentage points of the distributions. However, all of them use two parallel hyperplanes to find the decision function, which may lose the asymmetric information for all data points.

Twin support vector regression (TSVR) (\citeauthor{peng2010tsvr}, \citeyear{peng2010tsvr}; \citeauthor{gupta2019improved}, \citeyear{gupta2019improved}; \citeauthor{gupta2021regularization}, \citeyear{gupta2021regularization}; \citeauthor{shi2023robust}, \citeyear{shi2023robust}; \citeauthor{gu2023incremental}, \citeyear{gu2023incremental}) aims to generate two nonparallel functions such that each function determines the $\varepsilon$-insensitive lower or upper bounds of the unknown regressor, which can reflect the asymmetric information in data points. Shao et al. (\citeyear{shao2013varepsilon}) added a regularization term into TSVR, and proposed $\varepsilon$-twin support vector regression ($\varepsilon$-TSVR).
$\varepsilon$-TSVR yields the dual problems to be stable positive definite quadratic programming problems, and improves the performance of regression. To improve the generalization performance, an asymmetric $\nu$-twin support vector regression with pinball loss (Asy-$\nu$-TSVR) (\citeauthor{xu2018asymmetric}, \citeyear{xu2018asymmetric}) was proposed. Subsequently, an asymmetric Lagrangian $\nu$-twin with pinball loss (URALTSVR) (\citeauthor{gupta2021robust}, \citeyear{gupta2021robust}) was proposed to control the fitting error inside the asymmetric tube. Although two unparallel hyperplanes in the framework of TSVR effectively reflect the asymmetric information,
the use of this approach may lead to the loss of the unobserved heterogeneous information for all data points.

To capture the unobservable heterogeneous and asymmetric information in data points simultaneously,
we bring the spirit of QR into $\varepsilon$-TSVR,
and then propose a twin support vector quantile regression (TSVQR). On the one hand, TSVQR uses a quantile parameter to estimate the relationships between variables for all portions of a probability distribution.
TSVQR provides multiple trends of the distribution,
capturing the disparity in trends caused by heterogeneity.
On the other hand, TSVQR aims to generate two nonparallel functions at each quantile level such that each function determines the $\varepsilon$-insensitive lower or upper bounds of the unknown regressor,
which effectively captures the asymmetric information for each quantile level of the distribution. Therefore, TSVQR has the ability to capture the heterogeneous and asymmetric information in all data points simultaneously. Moreover, TSVQR solves two smaller-sized quadratic programming problems (QPPs), and each QPP has only one group of constraints for all data points.
The strategy of solving two smaller-sized QPPs makes TSVQR work quickly. In addition, dual coordinate descent algorithm is adopted to solve the QPPs,
which also accelerates the learning speed. The experimental results of six artificial data sets and five benchmark data sets demonstrate that TSVQR is much faster than SVQR, $\varepsilon$-SVQR, Online-SVQR, and a group penalized pseudo quantile regression (GPQR) (\citeauthor{ouhourane2022group}, \citeyear{ouhourane2022group}). Furthermore, TSVQR captures the heterogeneous and asymmetric information at each quantile level and offers a more complete view of the statistical landscape and the relationships among variables than SVQR, $\varepsilon$-SVQR, Online-SVQR, and GPQR. To sum up, the main contributions of this paper are summarized as follows:

 (\romannumeral1) TSVQR uses the quantile parameter to minimize an asymmetric version of errors, which yields a family of regression curves corresponding to different quantile levels and thus captures the heterogeneous information in data points.

 (\romannumeral2) TSVQR generates two nonparallel planes to measure the asymmetric distribution between the lower and upper bounds of each given quantile level,
 thus effectively capturing the asymmetric information at each given quantile level.

 (\romannumeral3) TSVQR solves two smaller quadratic programming problems and its each QPP has one group of constraints, which reduces the computational complexity. Moreover, the dual coordinate descent algorithm is adopted to solve its each QPP, leading to an accelerated learning speed.

 (\romannumeral4) The numerical experimental results and the statistical test results show that TSVQR outperforms
SVQR, $\varepsilon$-SVQR, GPQR, and Online-SVQR in terms of the effectiveness
of completely capturing the heterogeneous and asymmetric information. As the efficiency of the learning process, the training speed of TSVQR and URALTSVR is significantly faster than SVQR, $\varepsilon$-SVQR, GPQR, and Online-SVQR.

The rest of this paper is organized as follows. Section \ref{Sec:2} briefly introduces related works. In Section \ref{Sec:3}, we propose TSVQR in detail. Section \ref{Sec:4} describes the artificial and benchmark data set experiments, and section \ref{Sec:5} shows the application of TSVQR to big data sets. Section \ref{Sec:6} concludes this paper.

\section{Background}\label{Sec:2}
We briefly review SVR, SVQR, TSVR, and $\varepsilon$-TSVR that are closely related to twin support vector quantile regression (TSVQR). For simplicity, it is only concerned with the linear version. Table \ref{notation} lists the corresponding notations of these methods.
\begin{table}
\scriptsize
 \begin{center}
  \caption{List of notations.}
  \label{notation}
  \begin{tabular}{@{}ll|ll@{}}
   \hline
   Notation                              & Description                  &    Notation                & Description    \\
   \hline
    $\mathbf{A}$              &  input                      &        $\mathbf{Y}$          &   the response column vector\\
    $(\mathbf{A}_i,y_i)$                 &  the $i$-th training sample  & $(\mathbf{A},\mathbf{Y})$     & the training set \\
    $\mathbf{\xi}$ and $\mathbf{\xi^*}$  & slack variables & $\mathbf{e}$ & a vector of ones \\
    $\mathbf{w}$                         & the coefficients of the decision function & $b$ & an intercept \\
    $\mathbf{w_1}$                       & the coefficients of lower-bound function & $\mathbf{w_2}$ & the coefficients of up-bound function\\
    $b_1$                                & an intercept of lower-bound function & $b_2$ & an intercept of up-bound function\\
    $\tau $                              & a quantile parameter &   $\mathbf{K}$ & kernel matrix   \\
    $C, C_1, C_2$                        & parameters           & $\varepsilon, \varepsilon_1, \varepsilon_2$ & insensitive parameters \\
  \hline
  \end{tabular}
 \end{center}
\end{table}

\subsection{Support vector regression}
The optimal regression function of SVR is constructed as follows
\begin{equation}\label{SVRf}
f(x)=\mathbf{w^T}x+b,
\end{equation}
where $\mathbf{w}\in R^n$ denotes the coefficients of the decision function, and $b\in R$ is an intercept.

In standard SVR (\citeauthor{drucker1997support}, \citeyear{drucker1997support}), parameters $\mathbf{w}$ and $b$ in the regression
function (\ref{SVRf}) are estimated by solving the following
optimization problem
\begin{equation}\label{SVR}
\begin{split}
\underset{\mathbf{w}, b, \mathbf{\xi}, \mathbf{\xi^*}}{\min} & \frac{1}{2}||\mathbf{w}||^2+C \mathbf{e^T}(\mathbf{\xi}+\mathbf{\xi^*}) \\
 s.t.&~~  \mathbf{Y}-(\mathbf{Aw}+b \mathbf{e})\leq\varepsilon \mathbf{e}+\mathbf{\xi}, ~\mathbf{\xi}\geq \mathbf{0},\\
& (\mathbf{Aw}+b \mathbf{e})-\mathbf{Y} \leq
 \varepsilon \mathbf{e}+\mathbf{\xi^*},~\mathbf{\xi^*} \geq \mathbf{0},\\
 \end{split}
\end{equation}
where $|| \cdot ||^2$ represents the $L_2$-norm, $C>0$ is a
parameter determining the trade-off between the regularization term
and empirical risk, $\mathbf{\xi}$ and $\mathbf{\xi^*}$ are slack variables, $\mathbf{e}$ is
a vector of ones of appropriate dimensions, and \textbf{$\varepsilon > 0$ } is an insensitive parameter. Note that, the constraints in (\ref{SVR}) are adopted to generate
two parallel planes that locate more training samples locate in the flat region.
Therefore, SVR is not suitable for solving the existence of asymmetry
since two parallel planes cannot reflect the asymmetric information of the training samples.

\subsection{Support vector quantile regression}
In standard SVQR, $\mathbf{w}$ and $b$ in the regression function (\ref{SVRf}) are estimated by minimizing
\begin{eqnarray}\label{SVQR2}
\begin{aligned}
 \underset{\mathbf{w},b}{\min}~&\frac{1}{2}\|\mathbf{w}\|^2 + C (1-\tau) {\bf e}^\top \mathbf{\xi} + C \tau {\bf e}^\top \mathbf{\xi^*}\\
 s.t.&~~  \mathbf{Y}-(\mathbf{Aw}+b \mathbf{e}{+ \xi})\leq \mathbf{\xi}, ~\mathbf{\xi}\geq \mathbf{0},\\
& (\mathbf{Aw}+b \mathbf{e}+{\xi^*})-\mathbf{Y} \leq
 \mathbf{\xi^*},~\mathbf{\xi^*} \geq \mathbf{0},\\
\end{aligned}
\end{eqnarray}
where $|| \cdot ||^2$ represents the $L_2$-norm, $C>0$ is a
parameter determining the trade-off between the regularization term
and empirical risk, $\mathbf{\xi}$ and $\mathbf{\xi^*}$ are slack variables, $\mathbf{e}$ is
a vector of ones of appropriate dimensions, and $\tau~(0 <\tau < 1)$ is a quantile parameter.
SVQR minimizes the pinball loss function along with $L_2$-norm regularization term for the estimation of the optimal regression function. The pinball loss function in SVQR captures the information of training samples at different quantiles.
Thus, SVQR is suitable for solving the heterogeneous problem.
The basic idea of SVQR is to find the final decision function by maximizing the margin between two parallel hyperplanes.
However, using two parallel hyperplanes to find the decision function is not suitable for the existence of asymmetric information in the training data set.

\subsection{Twin support vector regression}
Different from SVR and SVQR, linear TSVR (\citeauthor{peng2010tsvr}, \citeyear{peng2010tsvr}) seeks a pair of lower-bound and upper-bound functions
\begin{equation*}\label{f1}
f_1(x)=\mathbf{w_1^T}x+b_1,
\end{equation*}
and
\begin{equation*}\label{f2}
f_2(x)=\mathbf{w_2^T}x+b_2,
\end{equation*}
where $\mathbf{w_1} \in R^n$, $\mathbf{w_2} \in R^n$, $b_1\in R$, and $b_2\in R$.

The linear TSVR can be formulated as the following minimization problems
\begin{equation*}\label{TSVR1}
\begin{split}
\underset{\mathbf{w_1}, b_1, \mathbf{\xi}}{\min} & \frac{1}{2}[\mathbf{Y}-\mathbf{e} \varepsilon_1 - (\mathbf{Aw_1}+b_1 \mathbf{e})]^T[\mathbf{Y}-\mathbf{e} \varepsilon_1 - (\mathbf{Aw_1}+b_1 \mathbf{e})] + C_1 \mathbf{e}^T \mathbf{\xi}  \\
 s.t.&~~  \mathbf{Y}-(\mathbf{Aw_1}+b_1 \mathbf{e})\geq \varepsilon_1 \mathbf{e}-\mathbf{\xi}, \mathbf{\xi}\geq \mathbf{0},\\
 \end{split}
\end{equation*}
and
\begin{equation*}\label{TSVR2}
\begin{split}
\underset{\mathbf{w_2}, b_2, \mathbf{\xi^*}}{\min} & \frac{1}{2}[\mathbf{Y}+\mathbf{e} \varepsilon_2 - (\mathbf{Aw_2}+b_2 \mathbf{e})]^T[\mathbf{Y}+\mathbf{e} \varepsilon_2 - (\mathbf{Aw_2}+b_2 \mathbf{e})] + C_2 \mathbf{e}^T \mathbf{\xi^*} \\
 s.t.&~~  (\mathbf{Aw_2}+b_2 \mathbf{e}) - \mathbf{Y} \geq \varepsilon_2 \mathbf{e}-\mathbf{\xi^*}, \mathbf{\xi^*} \geq \mathbf{0},\\
 \end{split}
\end{equation*}
where $C_1>0$, $C_2>0$, \textbf{$\varepsilon_1 > 0$ and $\varepsilon_2 > 0$} are parameters, and $\mathbf{\xi}$ and $\mathbf{\xi^*}$ are slack variables.

Shao et al. (\citeyear{shao2013varepsilon}) brought the structural risk minimization principle
into TSVR, and proposed $\varepsilon$-twin support vector regression ($\varepsilon$-TSVR) as follows:
\begin{equation*}\label{ETSVR1}
\begin{split}
\underset{\mathbf{w_1}, b_1, \mathbf{\xi}}{\min} & \frac{1}{2} C_3 (\|\mathbf{w_1}\|_2^2 +b_1^2)+ \frac{1}{2} \xi^T \xi + C_1 \mathbf{e}^T \mathbf{\xi}\\
 s.t.&~~ \mathbf{Y}- (\mathbf{Aw_1}+b_1 \mathbf{e})= \xi, \\
     &~~ \mathbf{Y}-(\mathbf{Aw_1}+b_1 \mathbf{e})\geq - \varepsilon_1 \mathbf{e}-\mathbf{\xi}, \mathbf{\xi}\geq \mathbf{0},\\
 \end{split}
\end{equation*}
and
\begin{equation*}\label{ETSVR2}
\begin{split}
\underset{\mathbf{w_2}, b_2, \mathbf{\xi^*}}{\min} & \frac{1}{2} C_4 (\|\mathbf{w_2}\|_2^2 +b_2^2)+ \frac{1}{2} {\xi^*}^T \xi^* + C_2 \mathbf{e}^T \mathbf{\xi^*}\\
s.t.&~~ (\mathbf{Aw_2}+b_2 \mathbf{e}) - \mathbf{Y}= \xi^*, \\
     &~~ (\mathbf{Aw_2}+b_2 \mathbf{e})-\mathbf{Y} \geq - \varepsilon_2 \mathbf{e}-\mathbf{\xi^*}, \mathbf{\xi^*}\geq \mathbf{0},\\
 \end{split}
\end{equation*}
where $C_3>0$, and $C_4>0$.

It can be seen that the function $f_1(x)$ determines the $\varepsilon_1$-insensitive lower-bound regressor,
and the function $f_2(x)$ determines the $\varepsilon_2$-insensitive upper-bound regressor.
Obviously, TSVR and $\varepsilon$-TSVR yield two nonparallel functions $f_1(x)$ and $f_2(x)$ to measure the asymmetric information of the training data set. However, TSVR and $\varepsilon$-TSVR are not suitable for the existence of heterogenous information in the training data points.

\section{Twin support vector quantile regression }\label{Sec:3}
In this section, we bring the spirit of quantile regression into twin support vector regression,
and then propose a twin support vector quantile regression (TSVQR) that takes unobserved heterogeneity and asymmetry into consideration.
In the following, we will
first propose a linear TSVQR, and then discuss the model property of TSVQR. Finally, we extend linear TSVQR into a nonlinear version.

\subsection{Linear twin support vector quantile regression}
\subsubsection{Problem formulation}
The goal of TSVQR is to find the following pair of lower-bound and upper-bound functions
\begin{equation*}\label{ff1}
f_1(x)=\mathbf{w_1^T}x+b_1,
\end{equation*}
and
\begin{equation*}\label{ff2}
f_2(x)=\mathbf{w_2^T}x+b_2.
\end{equation*}
The final decision function is constructed as
\begin{eqnarray*}\label{ff}
\begin{array}{ll}
 f(\mathbf{x})= \frac{1}{2}(f_1(\mathbf{x})+f_2(\mathbf{x})).\\
 \end{array}
\end{eqnarray*}

The minimization problems of TSVQR are as follows
\begin{eqnarray}\label{TSVQR1}
\begin{aligned}
 \underset{\mathbf{w_1},b_1,\mathbf{\xi}}{\min}~&\frac{1}{2}(\|\mathbf{w_1}\|_2^2 +b_1^2) + C_1\mathbf{e^T}\mathbf{\xi} + C_1\tau\mathbf{e^T}(\mathbf{Y}-\mathbf{Aw_1}-b_1\mathbf{e})\\
 \text{s.t.\ } & \mathbf{Y}-(\mathbf{Aw_1+b_1e})\geq \varepsilon_1 \mathbf{e}-\mathbf{\xi}, ~\mathbf{\xi}\geq \mathbf{0},\\
\end{aligned}
\end{eqnarray}
and
\begin{eqnarray}\label{TSVQR2}
\begin{aligned}
\underset{\mathbf{w_2},b_2,\mathbf{\xi}^*}{\min}~&\frac{1}{2}(\|\mathbf{w_2}\|_2^2 + b_2^2)+ C_2\mathbf{e^T}\mathbf{\xi}^* + C_2(1-\tau) \mathbf{e^T}(\mathbf{Aw_2}+b_2\mathbf{e}-\mathbf{Y})\\
\text{s.t.\ } & (\mathbf{Aw_2+b_2\mathbf{e}})-\mathbf{Y}\geq \varepsilon_2 \mathbf{e}- \mathbf{\xi}^*, ~\mathbf{\xi}^* \geq \mathbf{0},\\
\end{aligned}
\end{eqnarray}
where $C_1>0$, $C_2>0$, $\xi$ and $\xi^*$ are the slack variables, \textbf{$\varepsilon_1 > 0$ and $\varepsilon_2 > 0$ }are parameters, and $\tau~(0 <\tau < 1)$ is a quantile parameter.

It can be seen from (\ref{TSVQR1}) and (\ref{TSVQR2}) that the proposed TSVQR controls the negative influence of the training error on the final decision function by adjusting the quantile parameter $\tau$. As $\tau$ increases from $0$ to $1$, the final decision function $f(x)$ moves up through the training data. Therefore, TSVQR effectively depicts the heterogeneous information at each quantile level of the training data points. Moreover, TSVQR comprises a pair of quadratic programming problems such that each QPP determines the either of the lower-bound or upper-bound functions
by using one group of constraints. Specifically,
given the quantile parameter and the training data set, function $f_1(x)$ determines the $\varepsilon_1$-insensitive lower-bound regressor,
while function $f_2(x)$
determines the $\varepsilon_2$-insensitive upper-bound regressor. Obviously, $f_1(x)$ and $f_2(x)$ are nonparallel. Therefore, TSVQR effectively depicts the asymmetric information at each quantile level. TSVQR only solves two small-sized QPPs and each QPP has one group of constraints which reduces the computational complexity.

\subsubsection{Problem solution}
Define $\mathbf{G}=[\mathbf{A}~\mathbf{e}]$, $\mathbf{u_1}=[\mathbf{w_1^T} ~b_1]^T$, and  $\mathbf{u_2}=[\mathbf{w_2^T}~ b_2]^T$.
The Lagrangian functions for the problems (\ref{TSVQR1}) and (\ref{TSVQR2}) are as follows
\begin{eqnarray*}\label{L1}
\begin{aligned}
 L(\mathbf{u_1},\mathbf{\xi}, \mathbf{\alpha}, \mathbf{\beta}) = &\frac{1}{2}\|\mathbf{u_1}\|_2^2 + C_1\mathbf{e^T}\mathbf{\xi}
 + C_1\tau\mathbf{e^T}(\mathbf{Y}-\mathbf{u_1^TG}) \\
& -\mathbf{\alpha}^T (\mathbf{Y}-\mathbf{u_1^TG}-\varepsilon_1\mathbf{e}+\mathbf{\xi})- \mathbf{\beta}^T \mathbf{\xi},\\
\end{aligned}
\end{eqnarray*}
and
\begin{eqnarray*}\label{L2}
\begin{aligned}
 L(\mathbf{u_2},\mathbf{\xi^*}, \mathbf{\alpha^*}, \mathbf{\beta^*}) = &\frac{1}{2}\|\mathbf{u_2}\|_2^2 + C_2\mathbf{e^T}\mathbf{\xi^*}
 + C_2(1-\tau)\mathbf{e^T}(\mathbf{u_2^TG - \mathbf{Y}}) \\
& -\mathbf{\alpha^*}^T (\mathbf{u_2^TG}-\mathbf{Y}-\varepsilon_2\mathbf{e}+\mathbf{\xi^*})- \mathbf{\beta^*}^T \mathbf{\xi^*},\\
\end{aligned}
\end{eqnarray*}
where $\mathbf{\alpha}$, $\mathbf{\beta}$, $\mathbf{\alpha^*}$, and $\mathbf{\beta^*}$  are the Lagrangian multiplier vectors.
The KKT necessary and sufficient optimality conditions
for the problems (\ref{TSVQR1}) and (\ref{TSVQR2}) are given by
\begin{eqnarray}\label{KKT1}
\begin{aligned}
& \mathbf{u_1}-C_1\tau \mathbf{G^T} \mathbf{e}+ \mathbf{G^T} \mathbf{\alpha}=0,\\
& C_1\mathbf{e}-\mathbf{\alpha}-\mathbf{\beta}=0, \\
& \mathbf{Y}-\mathbf{u_1^TG} \geq \varepsilon_1 \mathbf{e}-\mathbf{\xi}, ~ \mathbf{\xi} \geq \mathbf{0}, \\
& \mathbf{\alpha^T}(\mathbf{Y}-\mathbf{u_1^TG}-\varepsilon_1 \mathbf{e}+ \mathbf{\xi} )=0, ~ \mathbf{\alpha} \geq \mathbf{0}, \\
& \mathbf{\beta^T} \mathbf{\xi}= \mathbf{0}, ~\mathbf{\beta}\geq \mathbf{0},\\
\end{aligned}
\end{eqnarray}
and
\begin{eqnarray}\label{KKT2}
\begin{aligned}
& \mathbf{u_2}+C_2(1-\tau) \mathbf{G^T} \mathbf{e} - \mathbf{G^T} \mathbf{\alpha^*}=0,\\
& C_2\mathbf{e}-\mathbf{\alpha^*}-\mathbf{\beta^*}=0, \\
& \mathbf{u_2^TG}-\mathbf{Y} \geq \varepsilon_2 \mathbf{e}-\mathbf{\xi^*}, ~\mathbf{\xi^*} \geq \mathbf{0}, \\
& \mathbf{\alpha^*}^T(\mathbf{u_2^TG}-\mathbf{Y}-\varepsilon_2 \mathbf{e}+ \mathbf{\xi^*} )=0, ~\mathbf{\alpha^*} \geq \mathbf{0}, \\
& \mathbf{\beta^*}^T \mathbf{\xi^*}= \mathbf{0}, ~\mathbf{\beta^*} \geq \mathbf{0}.\\
\end{aligned}
\end{eqnarray}

Using the above KKT conditions, we obtain the following dual representations of optimization problems (\ref{TSVQR1}) and (\ref{TSVQR2})
\begin{eqnarray}\label{DTSVQR1}
\begin{aligned}
 \underset{\mathbf{\alpha}}{\min}~&\frac{1}{2}\mathbf{\alpha^T}\mathbf{G}\mathbf{G^T}\mathbf{\alpha}-C_1\tau \mathbf{e^T}\mathbf{G}\mathbf{G^T}\mathbf{\alpha}+\mathbf{Y^T}\mathbf{\alpha}-\varepsilon_1 \mathbf{e^T}\mathbf{\alpha}\\
 \text{s.t.\ } & \mathbf{0} \leq \mathbf{\alpha} \leq C_1 \mathbf{e},\\
\end{aligned}
\end{eqnarray}
and
\begin{eqnarray}\label{DTSVQR2}
\begin{aligned}
 \underset{\mathbf{\alpha^*}}{\min}~&\frac{1}{2}\mathbf{{\alpha^*}^T}\mathbf{G}\mathbf{G^T}\mathbf{\alpha^*}-C_2(1-\tau) \mathbf{e^T}\mathbf{G}\mathbf{G^T}\mathbf{\alpha^*}-\mathbf{Y^T}\mathbf{\alpha^*}-\varepsilon_2 \mathbf{e^T}\mathbf{\alpha^*}\\
 \text{s.t.\ } &  \mathbf{0} \leq \mathbf{\alpha^*} \leq C_2 \mathbf{e}.\\
\end{aligned}
\end{eqnarray}

We adopt the dual coordinate descent method (DCDM) (\citeauthor{hsieh2008dual}, \citeyear{hsieh2008dual}) to solve problems (\ref{DTSVQR1}) and (\ref{DTSVQR2}).
Define $\bar{\mathbf{G}}=\mathbf{G}\mathbf{G^T}$, $\mathbf{d_1}= C_1\tau\mathbf{G}\mathbf{G^T}\mathbf{e}-\mathbf{Y}+\varepsilon_1\mathbf{e} $, and
$\mathbf{d_2}=C_2(1-\tau)\mathbf{G}\mathbf{G^T}\mathbf{e}+\mathbf{Y}+\varepsilon_2\mathbf{e}$, (\ref{DTSVQR1}) and (\ref{DTSVQR2}) can be rewritten as
\begin{eqnarray*}\label{RDTSVQR1}
\begin{aligned}
 \underset{\mathbf{\alpha}}{\min}~~f(\mathbf{\alpha})=~&\frac{1}{2}\mathbf{\alpha^T}\bar{\mathbf{G}}\mathbf{\alpha}-\mathbf{d_1}^T\mathbf{\alpha}\\
 \text{s.t.\ } & \mathbf{0} \leq \mathbf{\alpha} \leq C_1 \mathbf{e},\\
\end{aligned}
\end{eqnarray*}
and
\begin{eqnarray*}\label{RDTSVQR2}
\begin{aligned}
 \underset{\mathbf{\alpha^*}}{\min}~~f^*(\mathbf{\alpha^*})=~&\frac{1}{2}\mathbf{\alpha^*}^T\bar{\mathbf{G}}\mathbf{\alpha^*}-\mathbf{d_2}^T\mathbf{\alpha^*}\\
 \text{s.t.\ } & \mathbf{0} \leq \mathbf{\alpha^*} \leq C_2 \mathbf{e}.\\
\end{aligned}
\end{eqnarray*}
The DCDM starts with a random initial points $\mathbf{\alpha}^0$ and $\mathbf{\alpha^*}^0$, and for each iteration one of the variables is selected to minimize the following problems while the other variables are kept as constant solution
\begin{eqnarray}\label{RRDTSVQR1}
\begin{aligned}
 & \underset{\mathbf{t}}{\min}~~f(\mathbf{\alpha_i}+tI_i)\\
 & \text{s.t.\ }  0 \leq \mathbf{\alpha_i}+t \leq C_1,~i=1, 2, \cdots, l,\\
\end{aligned}
\end{eqnarray}
and
\begin{eqnarray}\label{RRDTSVQR2}
\begin{aligned}
 & \underset{\mathbf{t^*}}{\min}~~f^*(\mathbf{\alpha^{*}_i}+t^*I_i)\\
& \text{s.t.\ } 0 \leq \mathbf{\alpha^{*}_i} + t^*\leq C_2,~i=1, 2, \cdots, l,\\
\end{aligned}
\end{eqnarray}
where $I_i$ denotes the vector with $1$ in the $i$-th coordinate and $0$ elsewhere.
The objective functions of (\ref{RRDTSVQR1}) and (\ref{RRDTSVQR2}) are simplified as
\begin{eqnarray*}\label{SRDTSVQR1}
\begin{aligned}
 & f(\mathbf{\alpha_i}+tI_i)=\frac{1}{2}\bar{\mathbf{G}}_{ii}t^2+ \nabla_if(\alpha_i)t+f(\alpha_i) ,~i=1, 2, \cdots, l,\\
\end{aligned}
\end{eqnarray*}
and
\begin{eqnarray*}\label{SRDTSVQR2}
\begin{aligned}
f^*(\mathbf{\alpha^{*}_i}+t^*I_i)=\frac{1}{2}\bar{\mathbf{G}}_{ii}{t^*}^2+ \nabla_if^*(\alpha^*_i)t^*+f^*(\alpha^*_i),~i=1, 2, \cdots, l,\\
\end{aligned}
\end{eqnarray*}
where $\nabla_if$ and $\nabla_if^*$ are the $i$-th components of the gradients $\nabla f$ and $\nabla f^*$, respectively.
Combining the bounded constraints $0 \leq \mathbf{\alpha_i}+t \leq C_1$ and $0 \leq \mathbf{\alpha^{*}_i} + t^*\leq C_2$,
 the minimizers of (\ref{RRDTSVQR1}) and (\ref{RRDTSVQR2}) lead to the following bounded constraint solutions
 \begin{eqnarray*}\label{STSVQR1}
\begin{aligned}
 & \alpha^{new}_i= min(max(\alpha_i-\frac{\nabla_i f(\alpha_i)}{\mathbf{\bar{G}}_{ii}},0), C_1),\\
\end{aligned}
\end{eqnarray*}
and
 \begin{eqnarray*}\label{STSVQR2}
\begin{aligned}
 & \alpha^{*new}_i= min(max(\alpha^*_i-\frac{\nabla_i f^*(\alpha^*_i)}{\mathbf{\bar{G}}_{ii}},0), C_2).\\
\end{aligned}
\end{eqnarray*}
The corresponding algorithm of linear TSVQR is summarized in Algorithm \ref{Alg:1}.
\begin{algorithm}
\SetAlgoNoLine
\KwIn{Training set $(\mathbf{A},\mathbf{Y})$; Parameters $C_1$, $C_2$, $\tau$, $\varepsilon_1$ and $\varepsilon_2$;}
\KwOut{Updated $\mathbf{\alpha}$ and $\mathbf{\alpha^*}$;}
\Begin{Set $\mathbf{\alpha}=\mathbf{0}$ and $\mathbf{\alpha^*}=\mathbf{0}$;\\
Compute $\mathbf{d}= C_1\tau\mathbf{G}\mathbf{G^T}\mathbf{e}-\mathbf{Y}+\varepsilon_1\mathbf{e}$,
$\mathbf{d^*}=C_2(1-\tau)\mathbf{G}\mathbf{G^T}\mathbf{e}+\mathbf{Y}+\varepsilon_2\mathbf{e}$, and $\bar{\mathbf{G}}=\mathbf{G}\mathbf{G^T}$;\\
\If {$\mathbf{\alpha}$ not converge;}
{\textbf{do for $i=1,2,\cdots, l$ do} \\
$\nabla_i f(\alpha_i)\leftarrow \alpha^T \bar{\mathbf{G}}_{ii}I_i-d^TI_i$\\
$\alpha_i^{old} \leftarrow \alpha_i$;\\
$\alpha_i \leftarrow min(max(\alpha_i-\frac{\nabla_i f(\alpha_i)}{\mathbf{\bar{G}}_{ii}},0), C_1)$;}
\If {$\mathbf{\alpha^*}$ not converge;}
{\textbf{do for $i=1,2,\cdots, l$ do} \\
$\nabla_i f(\alpha^*_i)\leftarrow \alpha^{*T} \bar{\mathbf{G}}_{ii}I_i-d^{*T}I_i$\\
$\alpha^{*old}_i \leftarrow \alpha^*_i$;\\
$\alpha^{*}_i \leftarrow min(max(\alpha^*_i-\frac{\nabla_i f^*(\alpha^*_i)}{\mathbf{\bar{G}}_{ii}},0), C_2)$;}
}
\caption{ Dual coordinate descent algorithm for linear TSVQR.} \label{Alg:1}
\end{algorithm}

By introducing the transformation from $R^n$ to $ R^{n+1}:= \begin{bmatrix}
     x \\
     1
\end{bmatrix} $, the functions $f_1(x)$ and $f_2(x)$ can be expressed as
\begin{eqnarray*}\label{jf1}
\begin{array}{ll}
 f_1(\mathbf{x}) = \mathbf{u_1^{\top}x}=[\mathbf{w_1^T}, b_1]\begin{bmatrix}
     x \\
     1
\end{bmatrix},
 \end{array}
\end{eqnarray*}
and
\begin{eqnarray*}\label{jf2}
\begin{array}{ll}
 f_2(\mathbf{x}) = \mathbf{u_2^{\top}x}=[\mathbf{w_2^T}, b_1]\begin{bmatrix}
     x \\
     1
\end{bmatrix}.
 \end{array}
\end{eqnarray*}

Obtaining the solutions of $\mathbf{\alpha}$ and $\mathbf{\alpha^*}$ by Algorithm \ref{Alg:1},
the lower- and upper- bound functions $f_1(\mathbf{x})$ and $f_2(\mathbf{x})$ can be rewritten as
\begin{eqnarray*}\label{f1}
\begin{array}{ll}
 f_1(\mathbf{x})&= \mathbf{u_1^{\top}x}\\
 &=C_1\tau \mathbf{e}^T\mathbf{Gx}-\mathbf{\alpha^T}\mathbf{Gx},
 \end{array}
\end{eqnarray*}
and
\begin{eqnarray*}\label{f2}
\begin{array}{ll}
 f_2(\mathbf{x})&= \mathbf{u_2^{\top}x}\\
 & =-C_2(1-\tau) \mathbf{e}^T\mathbf{Gx}+\mathbf{\alpha^{*T}}\mathbf{Gx}.
 \end{array}
\end{eqnarray*}

The final decision function is constructed as
\begin{eqnarray}\label{ff}
\begin{array}{ll}
 f(\mathbf{x})&= \frac{1}{2}(f_1(\mathbf{x})+f_2(\mathbf{x}))\\
 & = \frac{1}{2}(\mathbf{\alpha^*}-\mathbf{\alpha})^T\mathbf{Gx}+
 \frac{1}{2} [C_1\tau-C_2 (1-\tau)]\mathbf{e}^T\mathbf{Gx}.
 \end{array}
\end{eqnarray}
The first term in $f(\mathbf{x})$ is the sum of $\mathbf{\alpha}$ and $\mathbf{\alpha^*}$, reflecting the asymmetric information at each quantile location of the training data points.
The second term in $f(\mathbf{x})$, using the quantile parameter,
depicts the heterogeneous information at each quantile level of the training data points.
Therefore, TSVQR has the ability to solve the problems of heterogeneity and asymmetry.
\subsubsection{Geometric interpretation}
As the quantile parameter $\tau$ ranges from $0$ to $1$,
we can obtain a family of $f_1(x)$, $f_2(x)$, and $f(x)$, respectively. We define the
$\tau$-th lower-bound function as $f_{1,\tau}(\mathbf{x})$, the upper-bound function as $f_{2,\tau}(\mathbf{x})$, and final decision function as $f_{\tau}(\mathbf{x})$. For simplicity of exposition,
we focus on five cases where the values of $\tau$
are $0.10$, $0.25$, $0.50$, $0.75$, and $ 0.90$, respectively. Then, we obtain the following down-bound functions\\
(\romannumeral1)~~~when $\tau=0.10$, $f_1(x)$ becomes $f_{1,0.10}(\mathbf{x})=0.10 C_1\mathbf{e}^T\mathbf{Gx}-\mathbf{\alpha}^T\mathbf{Gx},$\\
(\romannumeral2)~~when $\tau=0.25$, $f_1(x)$ becomes $f_{1,0.25}(\mathbf{x})=0.25 C_1\mathbf{e}^T\mathbf{Gx}-\mathbf{\alpha}^T\mathbf{Gx},$\\
(\romannumeral3) when  $\tau=0.50$, $f_1(x)$ becomes $f_{1,0.50}(\mathbf{x})=0.50 C_1\mathbf{e}^T\mathbf{Gx}-\mathbf{\alpha}^T\mathbf{Gx},$\\
(\romannumeral4) when  $\tau=0.75$, $f_1(x)$ becomes $f_{1,0.75}(\mathbf{x})=0.75 C_1\mathbf{e}^T\mathbf{Gx}-\mathbf{\alpha}^T\mathbf{Gx}$,\\
(\romannumeral5)~~when $\tau=0.90$, $f_1(x)$ becomes $f_{1,0.90}(\mathbf{x})=0.90 C_1\mathbf{e}^T\mathbf{Gx}-\mathbf{\alpha}^T\mathbf{Gx}$,\\
and the following up-bound functions \\
(\romannumeral1)~~~when $\tau=0.10$, $f_2(x)$ becomes $f_{2,0.10}(\mathbf{x})=\mathbf{\alpha^*}^T \mathbf{Gx}-0.90 C_2\mathbf{e}^T\mathbf{Gx}$,\\
(\romannumeral2)~~when $\tau=0.25$, $f_2(x)$ becomes $f_{2,0.25}(\mathbf{x})=\mathbf{\alpha^*}^T \mathbf{Gx}-0.75 C_2\mathbf{e}^T\mathbf{Gx}$,\\
(\romannumeral3) when $\tau=0.50$, $f_2(x)$ becomes $f_{2,0.50}(\mathbf{x})=\mathbf{\alpha^*}^T \mathbf{Gx}-0.50 C_2\mathbf{e}^T\mathbf{Gx}$,\\
(\romannumeral4) when $\tau=0.75$, $f_2(x)$ becomes $f_{2,0.75}(\mathbf{x})=\mathbf{\alpha^*}^T \mathbf{Gx}-0.25 C_2\mathbf{e}^T\mathbf{Gx}$,\\
(\romannumeral5)~~when $\tau=0.90$, $f_2(x)$ becomes $f_{2,0.90}(\mathbf{x})=\mathbf{\alpha^*}^T \mathbf{Gx}-0.10 C_2\mathbf{e}^T\mathbf{Gx}$. \\

We calculate the mean of $f_{1,\tau}(\mathbf{x})$ and $f_{2,\tau}(\mathbf{x})$, and obtain the following cases of $f_{\tau}(\mathbf{x})$.

$\mathbf{Case ~1}$: When $\tau=0.10$, the final decision function (\ref{ff}) becomes
\begin{eqnarray*}\label{f0.1}
\begin{array}{ll}
 f_{\tau={0.10}}(\mathbf{x})= \frac{1}{2}(\mathbf{\alpha^*}-\mathbf{\alpha})^T\mathbf{Gx}+
 \frac{1}{2} (0.1C_1-0.9C_2)\mathbf{e}^T\mathbf{Gx}.
 \end{array}
\end{eqnarray*}
$f_{\tau={0.10}}(x)$ represents the conditional distribution of the training samples at the $0.1$th quantile level.

$\mathbf{Case ~2}$: When $\tau=0.25$, the final decision function (\ref{ff}) becomes
\begin{eqnarray*}\label{f0.25}
\begin{array}{ll}
 f_{\tau={0.25}}(\mathbf{x})= \frac{1}{2}(\mathbf{\alpha^*}-\mathbf{\alpha})^T\mathbf{Gx}+
 \frac{1}{2} (0.25C_1-0.75C_2)\mathbf{e}^T\mathbf{Gx}.
 \end{array}
\end{eqnarray*}
$f_{\tau={0.25}}(\mathbf{x})$ represents the conditional distribution of the training samples at first quantile.

$\mathbf{Case ~3}$: When $\tau=0.50$, the final decision function (\ref{ff}) becomes
\begin{eqnarray*}\label{f0.50}
\begin{array}{ll}
 f_{\tau={0.50}}(\mathbf{x})= \frac{1}{2}(\mathbf{\alpha^*}-\mathbf{\alpha})^T\mathbf{Gx}+
 \frac{1}{2} (0.5C_1-0.5C_2)\mathbf{e}^T\mathbf{Gx}.
 \end{array}
\end{eqnarray*}
$f_{\tau={0.50}}(\mathbf{x})$ represents the conditional distribution of the training samples at median level.

$\mathbf{Case ~4}$: When $\tau=0.75$, the final decision function (\ref{ff}) becomes
\begin{eqnarray*}\label{f0.75}
\begin{array}{ll}
 f_{\tau={0.75}}(\mathbf{x})= \frac{1}{2}(\mathbf{\alpha^*}-\mathbf{\alpha})^T\mathbf{Gx}+
 \frac{1}{2} (0.75C_1-0.25C_2)\mathbf{e}^T\mathbf{Gx}.
 \end{array}
\end{eqnarray*}
$f_{\tau={0.75}}(\mathbf{x})$ represents the conditional distribution of the training samples at third quantile level.

$\mathbf{Case ~5}$: When $\tau=0.90$, the final decision function (\ref{ff}) becomes
\begin{eqnarray*}\label{f0.9}
\begin{array}{ll}
 f_{\tau={0.90}}(\mathbf{x})= \frac{1}{2}(\mathbf{\alpha^*}-\mathbf{\alpha})^T\mathbf{Gx}+
 \frac{1}{2} (0.9C_1-0.1C_2)\mathbf{e}^T\mathbf{Gx}.
 \end{array}
\end{eqnarray*}
$f_{\tau={0.90}}(\mathbf{x})$ represents the conditional distribution of the training samples at the $0.9$th quantile level.

\begin{figure}[htbp]
\centering
\subfigure[$\tau=0.10$]{\includegraphics[width=0.35\textheight]{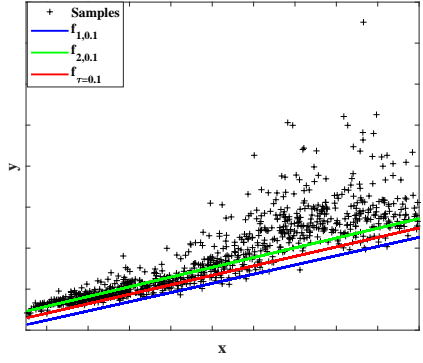}}
\subfigure[$\tau=0.25$]{\includegraphics[width=0.35\textheight]{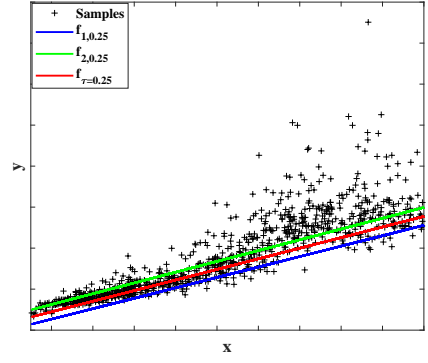}}
\subfigure[$\tau=0.50$]{\includegraphics[width=0.35\textheight]{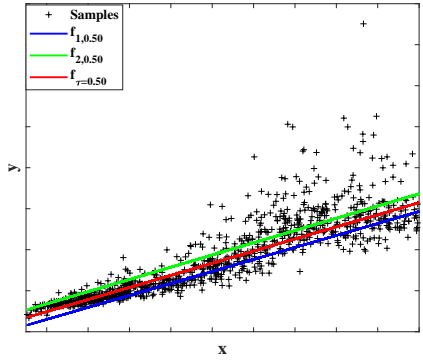}}
\subfigure[$\tau=0.75$]{\includegraphics[width=0.35\textheight]{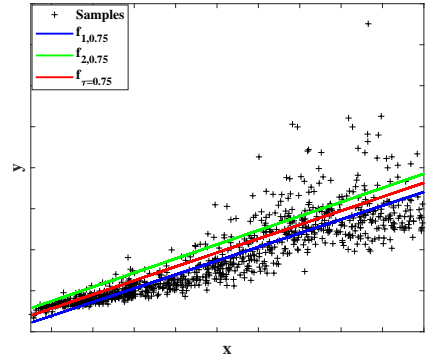}}
\subfigure[$\tau=0.90$]{\includegraphics[width=0.35\textheight]{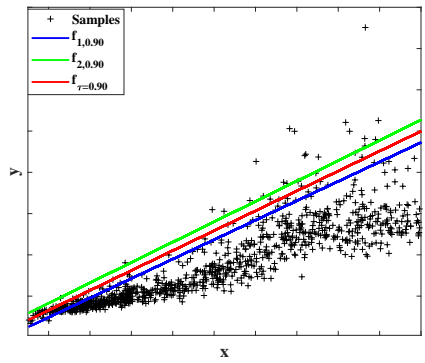}}
\caption{The geometric interpretation for TSVQR ($\tau=0.10, 0.25, 0.50, 0.75,$ and $0.90$).}\label{TSVQR interp}
\end{figure}

The intuitive geometric interpretation of TSVQR is shown in Fig. 1. The blue lines represent the lower-bound function $f_{1}(x)$ with different cases.
The green lines represent the upper-bound function $f_2(x)$ with different cases. From Fig.~\ref{TSVQR interp} (a)-(e),
we observe that $f_{1,\tau}(\mathbf{x})$ and $f_{2,\tau}(\mathbf{x})$ are nonparallel,
demonstrating the information up and down the $\tau$-th quantile location, respectively.
Moreover, $f_{1,\tau}(\mathbf{x})$ and $f_{2,\tau}(\mathbf{x})$ adjust the quantile parameter to depict the whole distribution information, especially
when unobservable heterogeneity exists in the training data points. The final decision function $f_{\tau}(\mathbf{x})$ in the middle of $f_{1,\tau}(\mathbf{x})$ and $f_{2,\tau}(\mathbf{x})$ is
 represented by the red lines in Fig.~\ref{TSVQR interp}. Final decision functions
 provide a more complete picture of heterogeneous and asymmetric information in the training data points.

\subsection{Model property}
\subsubsection{Support vector}
\begin{theorem}
Suppose that $\mathbf{\bar{\alpha}}=(\bar{\alpha}_1, \dots, \bar{\alpha}_l)^T$ is the solution to (\ref{DTSVQR1}) for $i=1, \dots, l$,
and $f_1(x)=\mathbf{\bar{w}_1^{\top}}x+\bar{b}_1$ is the optimal down-bound regression function.

(\romannumeral1) For $0<\bar{\alpha}_{i}< C_1$, the corresponding data point
$(x_i,y_i)$ lies on the $\varepsilon_1$-insensitive down-bound regressor.

(\romannumeral2) For $\bar{\alpha_i}=C_1$, the corresponding data point $(x_i,y_i)$ lies below
the $\varepsilon_1$-insensitive down-bound regressor.

(\romannumeral3) For $\bar{\alpha_i}=0$, the corresponding data point $(x_i,y_i)$ lies
above the $\varepsilon_1$-insensitive down-bound regressor.
\end{theorem}

\begin{proof}
By KKT conditions (\ref{KKT1}), we obtain
\begin{eqnarray*}\label{LL31}
\begin{aligned}
&\mathbf{\bar{\alpha}} \in [\mathbf{0}, C_1\mathbf{e}], ~ \mathbf{\bar{\alpha}^T}(\mathbf{Y}-\mathbf{A}\mathbf{\bar{w}_1}-\bar{b}_1 \mathbf{e}-\varepsilon_1 \mathbf{e}+ \mathbf{\bar{\xi}})=0.
\end{aligned}
\end{eqnarray*}

(\romannumeral1) If $\mathbf{0}<\mathbf{\bar{\alpha}}< C_1\mathbf{e}$,
then $ \mathbf{0} <\mathbf{\bar{\beta}}< C_1\mathbf{e} $, and $\mathbf{\bar{\xi}}=0$.
We obtain
$\mathbf{Y}=\mathbf{\bar{w}_1^{\top}}x+\bar{b}_1\mathbf{e}+\varepsilon_1 \mathbf{e}$.
For $0<\bar{\alpha}_i<C_1$, the corresponding data point
$(x_i,y_i)$ lies on the $\varepsilon_1$-insensitive down-bound regressor.

(\romannumeral2) If $\mathbf{\bar{\alpha}}=C_1 \mathbf{e}$,
then $ \mathbf{\bar{\beta}}= \mathbf{0} $, and $\mathbf{\bar{\xi}}>\mathbf{0}$.
We obtain $\mathbf{Y} < \mathbf{\bar{w}_1^{\top}}x+\bar{b}_1\mathbf{e}+\varepsilon_1 \mathbf{e}$.
For $\bar{\alpha_i}=C_1$, the corresponding data point $(x_i,y_i)$ lies below
the $\varepsilon_1$-insensitive down-bound regressor.

(\romannumeral3) If $\mathbf{\bar{\alpha}}= \mathbf{0} $,
then $\mathbf{\bar{\beta}}= C_1 \mathbf{e} $, and $\mathbf{\bar{\xi}}= \mathbf{0}$.
We obtain $\mathbf{Y} > \mathbf{\bar{w}_1^{\top}}x+\bar{b}_1\mathbf{e}+\varepsilon_1 \mathbf{e}$.
For $\bar{\alpha_i}=0$, the corresponding data point $(x_i,y_i)$ lies above
the $\varepsilon_1$-insensitive down-bound regressor.
\end{proof}

\begin{theorem}
Suppose that $\mathbf{\bar{\alpha}^*}=(\bar{\alpha}_1^*, \dots, \bar{\alpha}_l^*)^T$ is the solution to (\ref{DTSVQR2}) for $i=1, \dots, l$,
and $f_2(x)=\mathbf{\bar{w}_2^{\top}}x+\bar{b}_2$ is the optimal up-bound regression function.

(\romannumeral1) For $0<\bar{\alpha}_{i}^*< C_2$, the corresponding data point
$(x_i, y_i)$ lies on the $\varepsilon_2$-insensitive up-bound regressor.

(\romannumeral2) For $\bar{\alpha_i}^*=C_2$, the corresponding data point $(x_i, y_i)$ lies above
the $\varepsilon_2$-insensitive up-bound regressor.

(\romannumeral3) For $\bar{\alpha_i}^*=0$, the corresponding data point $(x_i, y_i)$ lies
below the $\varepsilon_2$-insensitive up-bound regressor.
\end{theorem}

\begin{proof}
By KKT conditions (\ref{KKT2}), we obtain
\begin{eqnarray*}\label{LL31}
\begin{aligned}
&  \mathbf{\bar{\alpha}^*} \in [\mathbf{0}, C_2\mathbf{e}], ~ \mathbf{\bar{\alpha}^{*T}}(\mathbf{\bar{w}_2^{\top}}x+\bar{b}_2\mathbf{e}-\varepsilon_2 \mathbf{e}-\mathbf{Y} + \mathbf{\bar{\xi}^*})=0.
\end{aligned}
\end{eqnarray*}

(\romannumeral1) If $\mathbf{0}<\mathbf{\bar{\alpha}^*}< C_2\mathbf{e}$,
then $ \mathbf{0} <\mathbf{\bar{\beta}^*}< C_2\mathbf{e} $, and $\mathbf{\bar{\xi}^*}=0$.
We obtain
$\mathbf{Y}=\mathbf{\bar{w}_2^{\top}}x+\bar{b}_2\mathbf{e}-\varepsilon_2 \mathbf{e}$.
For $0<\bar{\alpha}^*_i<C_2$, the corresponding data point
$(x_i, y_i)$ lies on the $\varepsilon_2$-insensitive up-bound regressor.

(\romannumeral2) If $\mathbf{\bar{\alpha}^*}=C_2 \mathbf{e}$,
then $ \mathbf{\bar{\beta}^*}= \mathbf{0} $, and $\mathbf{\bar{\xi}^*}>\mathbf{0}$.
We obtain $\mathbf{Y} > \mathbf{\bar{w}_2^{\top}}x +\bar{b}_2\mathbf{e}-\varepsilon_2 \mathbf{e}$.
For $\bar{\alpha^*_i}=C_2$, the corresponding data point $(x_i, y_i)$ lies above
the $\varepsilon_2$-insensitive up-bound regressor.

(\romannumeral3) If $\mathbf{\bar{\alpha}^*}= \mathbf{0} $,
then $\mathbf{\bar{\beta}^*}= C_2 \mathbf{e} $, and $\mathbf{\bar{\xi}^*}= \mathbf{0}$.
We obtain $\mathbf{Y} < \mathbf{\bar{w}_2^{\top}}x +\bar{b}_2\mathbf{e}-\varepsilon_2 \mathbf{e}$.
For $\bar{\alpha^*_i}=0$, the corresponding data point $(x_i, y_i)$ lies below
the $\varepsilon_2$-insensitive up-bound regressor.
\end{proof}

According to Theorems $1$ and $2$, we obtain
\begin{equation*}\label{linetheta}
\begin{cases}
y_i< \mathbf{\bar{w}_1^{\top}}x_i+\bar{b}_1+\varepsilon_1, & \alpha_i= C_1 ,\\
y_i= \mathbf{\bar{w}_1^{\top}}x_i+\bar{b}_1+\varepsilon_1, & 0 < \alpha_i < C_1,\\
\mathbf{\bar{w}_1^{\top}x_i}+\bar{b}_1+\varepsilon_1 < y_i < \mathbf{\bar{w}_2^{\top}}x_i+\bar{b}_2-\varepsilon_2, & \alpha_i=0, \alpha_i^*=0,\\
y_i=\mathbf{\bar{w}_2^{\top}}x_i+\bar{b}_2-\varepsilon_2, & 0 < \alpha_i^*< C_2,\\
y_i> \mathbf{\bar{w}_2^{\top}}x_i+\bar{b}_2-\varepsilon_2, & \alpha_i^*=C_2.
\end{cases}
\end{equation*}
The index set of support vector in the TSVQR can be defined as $I_{SV}=\{ i|0 < \alpha_i \leq C_1, 0 < \alpha_i^* \leq C_2\}$.

\subsubsection{Empirical Conditional Quantile Estimator}
\begin{prop}
Let p, N, and Z denote the number of positive, negative, and zero elements of the residual vector $y_i-f(x_i)$.
The minimizer of (\ref{TSVQR1}) and (\ref{TSVQR2}) satisfies:\\
(\romannumeral1)~N is bounded from above by $\tau l$.\\
(\romannumeral2)~p is bounded from above by $(1-\tau)l$.\\
(\romannumeral3)If $(x,y)$ is drawn iid from distribution $P(x,y)$, with $P(x,y)$ continuous and the expectation
of the modulus of absolute continuity of its density satisfying $lim_{\delta\rightarrow0}E[\varepsilon(\delta)]=0$.
With probability 1, asymptotically, $\frac{N}{l}$ equals $\tau$.
\end{prop}

\begin{proof}
For the
claims (\romannumeral1) and (\romannumeral2), we follow the proof process of Takeuchi et al. (\citeyear{takeuchi2006nonparametric}, Lemma 3) and Koenker(\citeyear{koenker2005quantile}, Theorem 2.2). Assume that we get optimal solutions of (\ref{TSVQR1}) and (\ref{TSVQR2}). Then, increasing the minimizer $f$ by $\delta f$
changes the objective by $[N(1-\tau)+Z(1-\tau)-p\tau]\delta f$. Likewise, decreasing the minimizer $f$ by $\delta f$ changes the objective by
$[-N(1-\tau)+\tau p+\tau Z]\delta f$. Requiring that both terms are nonnegative at optimality in conjunction with the fact that $N+p \leq  l$ and $l=N+p+Z$ prove the claims (\romannumeral1) and (\romannumeral2).

 For the claim (\romannumeral3), we follow the proof process of Sch{\"o}lkopf (\citeyear{scholkopf2000new}, Proposition 1). The condition on $P\{ y|x\}$ means that
 \begin{eqnarray}\label{Ep}
\begin{aligned}
\underset{f,t}{sup} E[P(|f(x)+t-y|< \gamma)] < \delta(\gamma).
\end{aligned}
\end{eqnarray}
 When $\gamma \rightarrow 0 $, function $\delta(\gamma)$ approaches $0$.
 We further get
 that for all $t$
 \begin{eqnarray*}\label{Ep2}
\begin{aligned}
P(\underset{f}{sup} (\hat{P}_l(|f(x)+t-y|< \gamma/2)<P(|f(x)+t-y|< \gamma))> \alpha)<c_1{c^{-l}_2},
\end{aligned}
\end{eqnarray*}
 where $\hat{P}_l$ is the sample-based estimate of $P$, and $c_1, c_2$ may depend on $\gamma$ and $\alpha$.
 Discretizing the values of $t$, taking the union bound, and applying equation (\ref{Ep}) show that the supremum over
 $f$ and $t$ of $\hat{P}_l(f(x)+t-y=0)$ converges to 0 in probability. Thus, the fraction of points on the edge of the tube
 almost surely converges to 0. With probability 1, asymptotically, $\frac{N}{l}$ equals $\tau$.
\end{proof}

\begin{figure*}[h!]
\centering
\subfigure[$x=-1$]{\includegraphics[width=0.35\textheight]{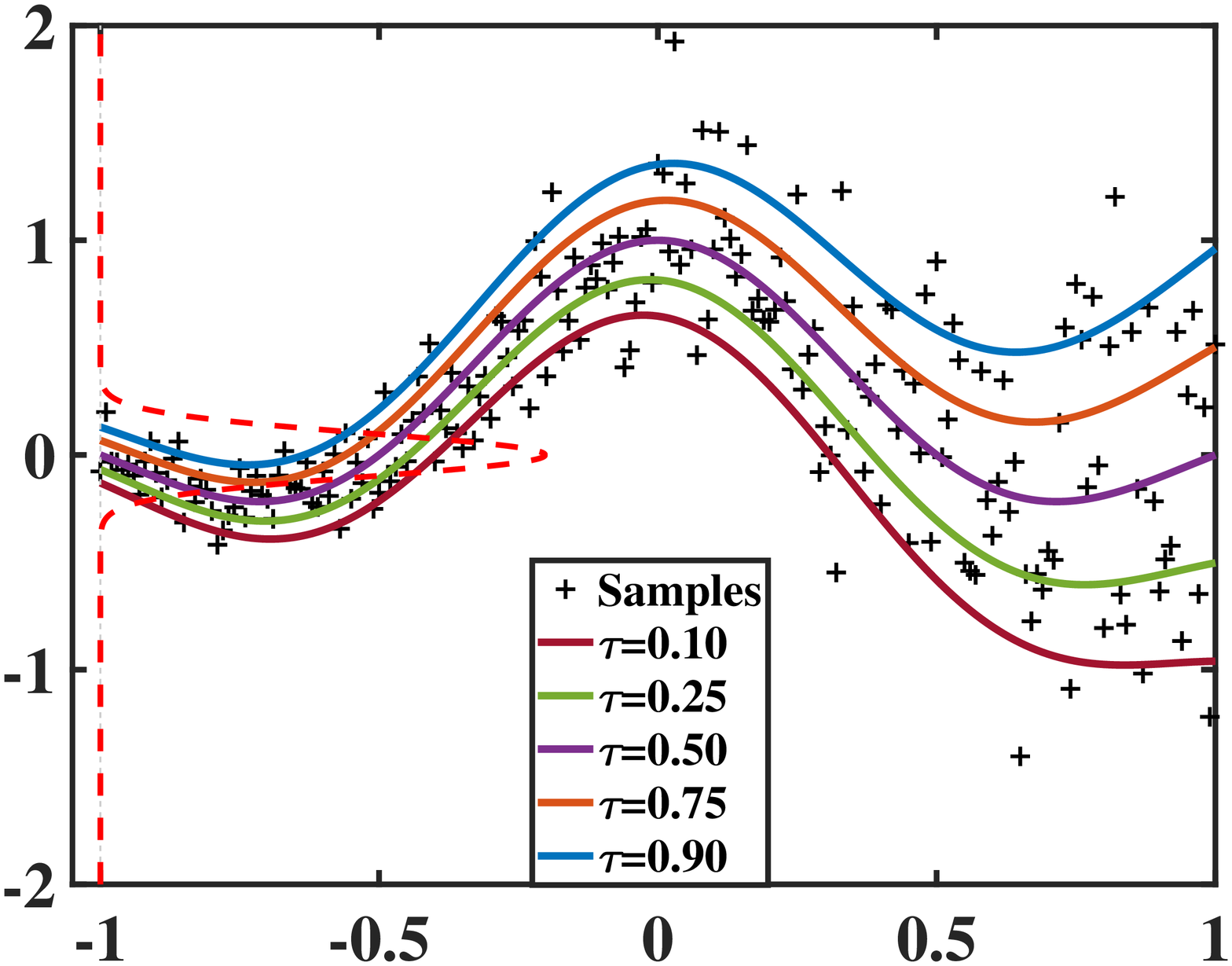}}
\subfigure[$x=-0.5$]{\includegraphics[width=0.35\textheight]{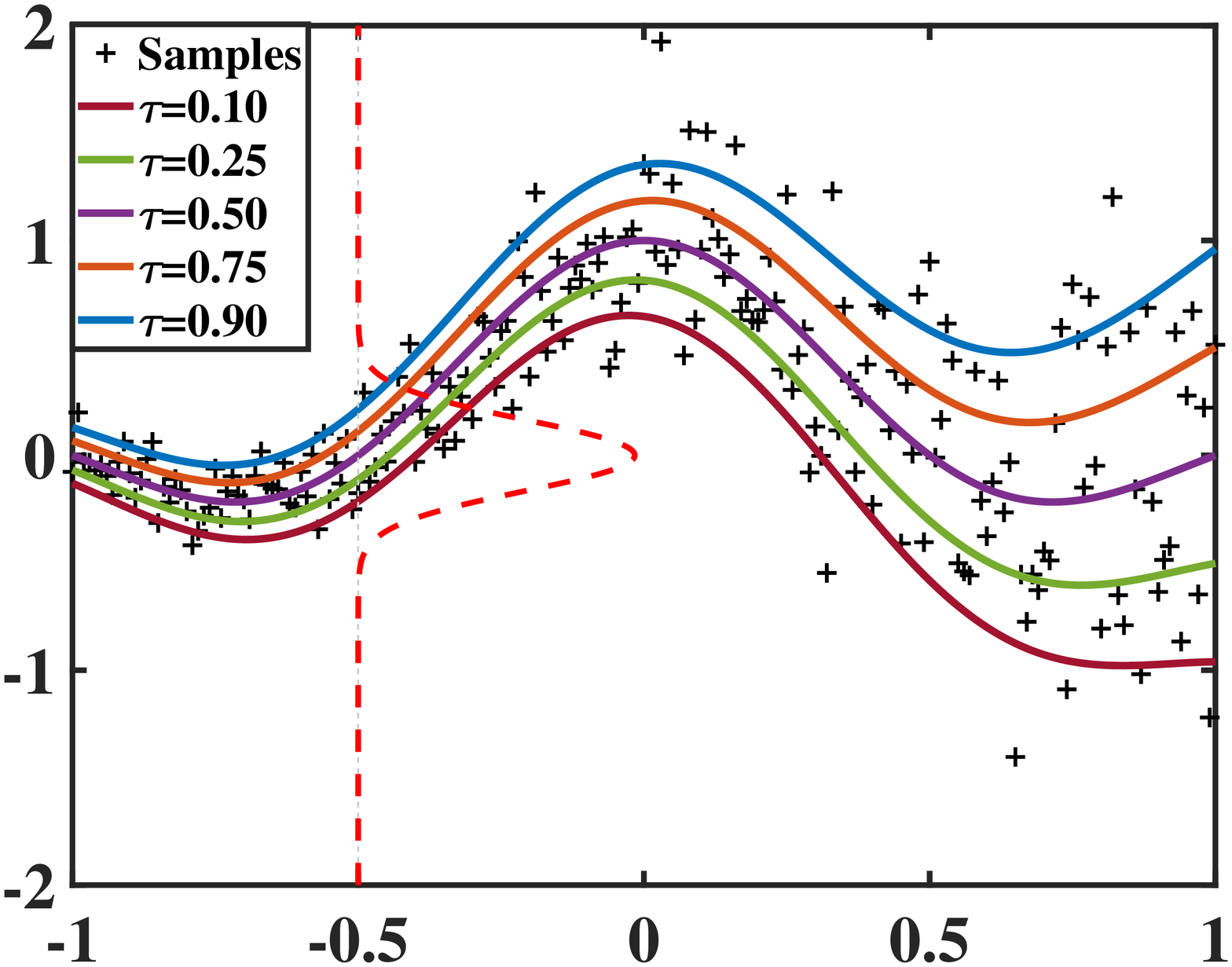}}
\subfigure[$x=0$]{\includegraphics[width=0.35\textheight]{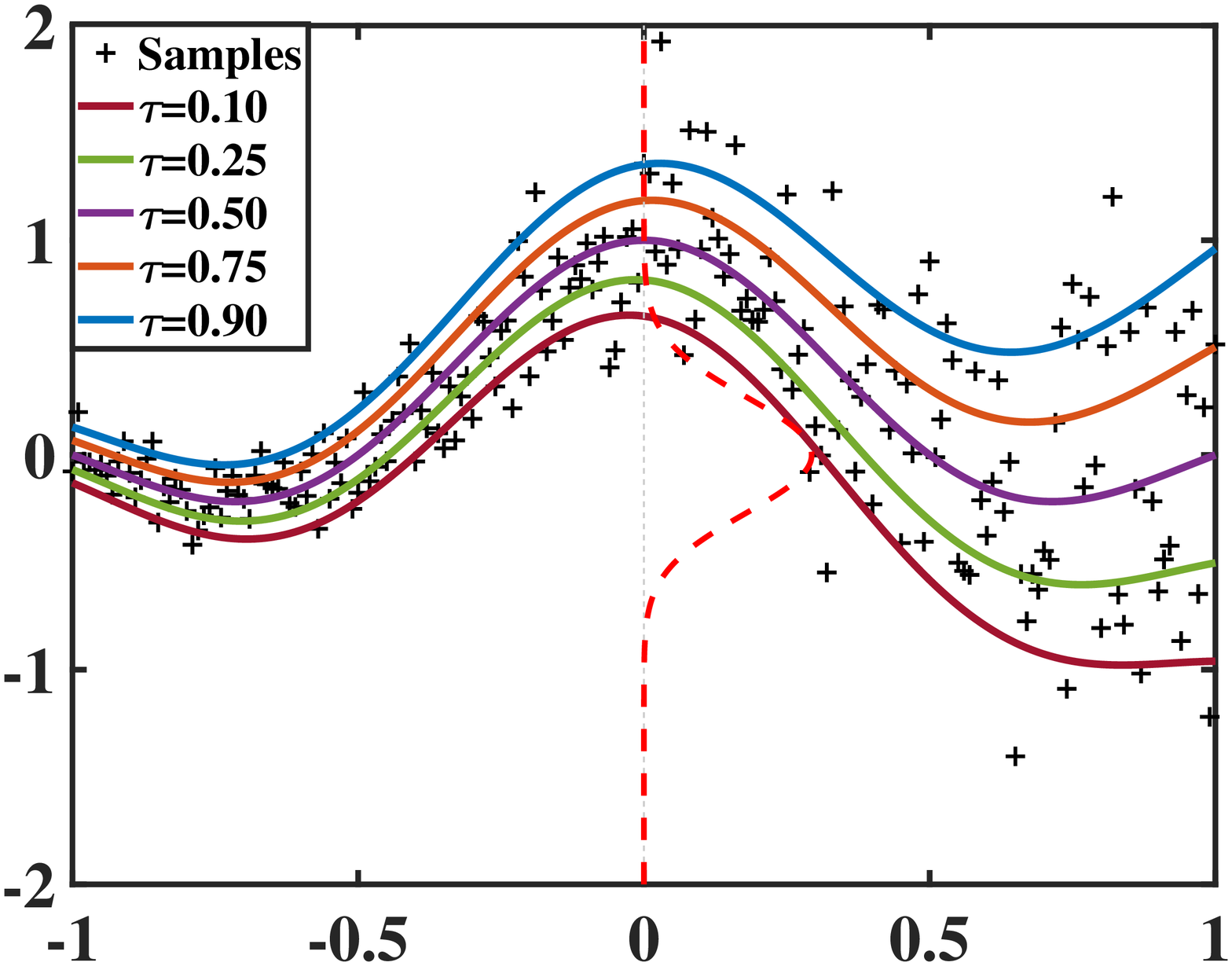}}
\subfigure[$x=-0.5$]{\includegraphics[width=0.35\textheight]{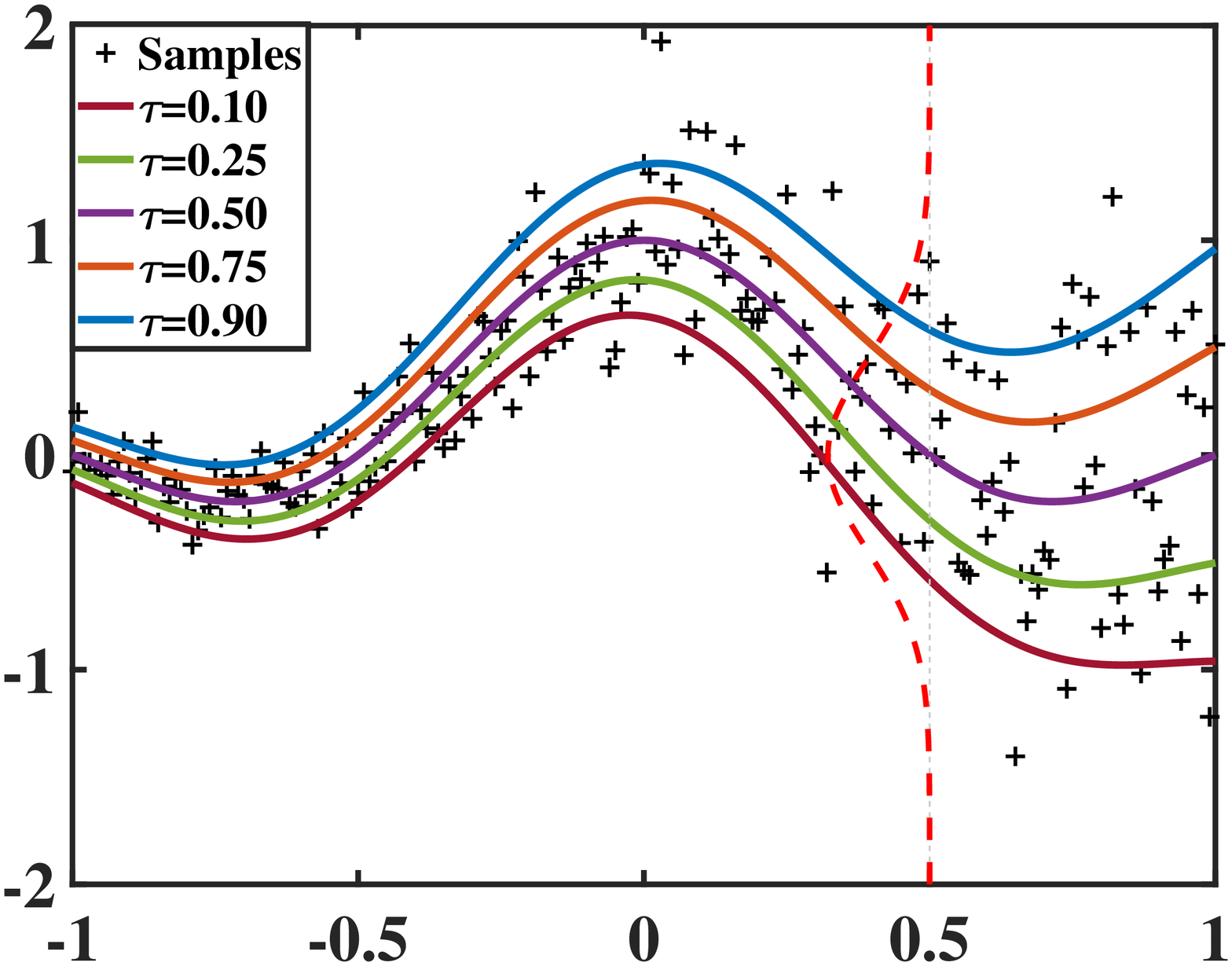}}
\subfigure[$x=1$]{\includegraphics[width=0.35\textheight]{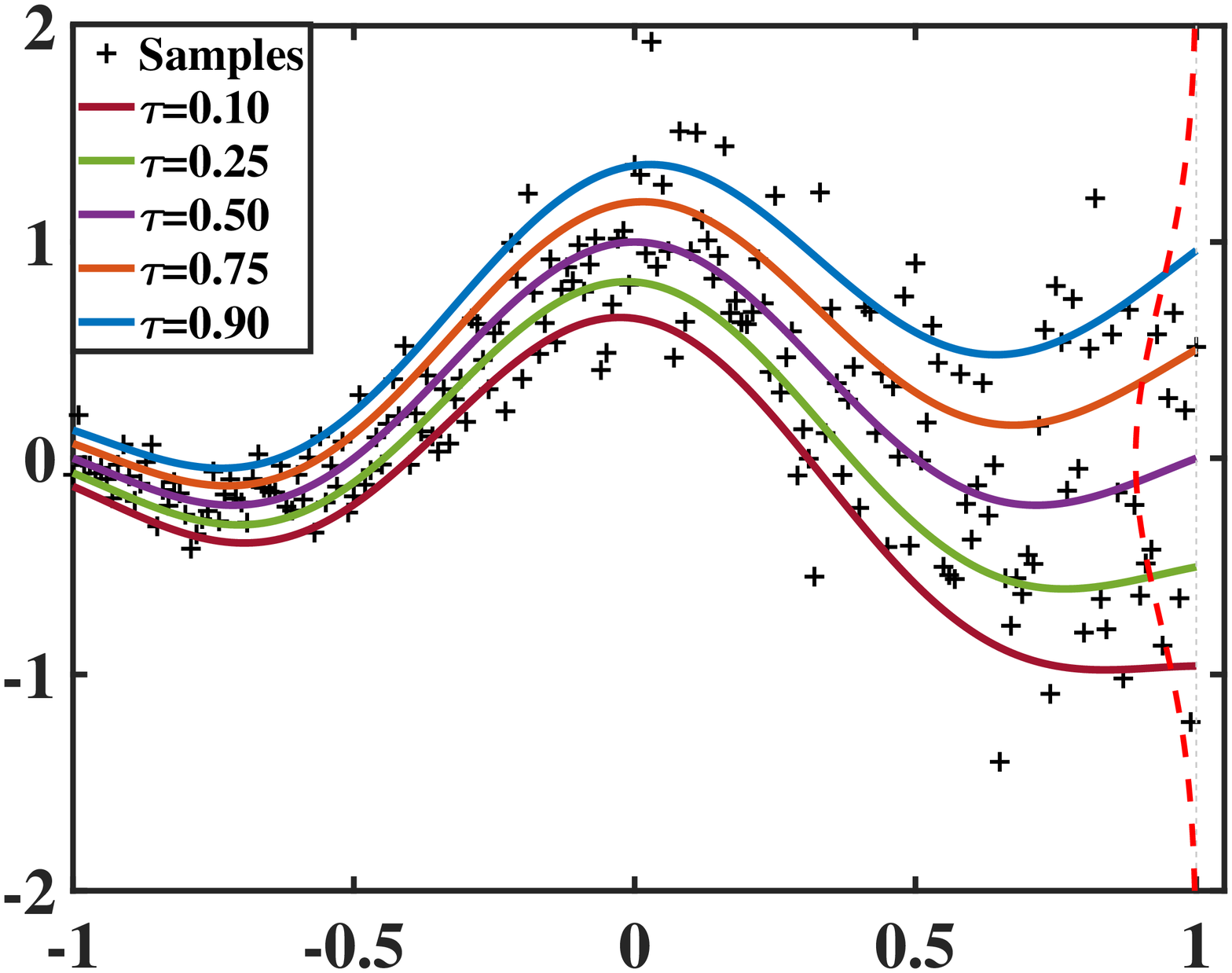}}
\caption{Conditional quantile functions and probability densities.}\label{cqf}
\end{figure*}

To illustrate regression analyses with the conditional quantile estimator, we provide the following example. Consider the relationship between $y$ and $x$ as
 \begin{eqnarray*}
\begin{aligned}
y(x) = \frac{sin(2 \pi x)}{2 \pi x}+\xi, \xi \sim N(0,{\sigma(x)}^2),\\
\end{aligned}
\end{eqnarray*}
 where $ \sigma(x)=0.1e^{1+x}$, and $x\in [-1,1].$
 Compute the $\tau$th quantiles by solving $P\{y\leq f |x \}= \tau$ explicitly. Since $\xi$ is normally distributed, the $\tau$th quantile of $\xi$
 is given by $\sigma(x)\phi^{-1}(\tau)$, where $\phi$ is the cumulative distribution function of the normal distribution with unit variance. We further get
 \begin{eqnarray*}
\begin{aligned}
y_{\tau}(x) = \frac{sin(2 \pi x)}{2 \pi x}+0.1e^{1+x}\phi^{-1}(\tau).\\
\end{aligned}
\end{eqnarray*}
 Fig. \ref{cqf} shows the conditional quantile functions of (\ref{Ep}) when $\tau=0.10$, $0.25$, $0.50$, $0.75$, and $0.90$. The probability densities $p(y|x=-1)$, $p(y|x=-0.5)$, $p(y|x=0)$, $p(y|x=0.5)$, and $p(y|x=1)$ are also illustrated in Fig. \ref{cqf}. The $\tau$th conditional quantile function is obtained by connecting the $\tau$th quantile of the conditional distribution $p(y|x)$ for all $x$.
 From Fig. \ref{cqf}, we find that $\tau=0.10$, $0.25$, $0.50$, $0.75$, and $0.90$ cases track the lower, first quantile, median, third quantile, and upper envelope of the data points, respectively.

\subsection{Nonlinear twin support vector quantile regression}
For the nonlinear TSVQR, we first introduce a nonlinear mapping from the original space to the feature space: $\Phi$ $\rightarrow$ $\chi = \mathbf{\Phi(X)}$.
The nonlinear regression functions are defined as follows

\begin{equation*}\label{Nf1}
f_1(x)=\mathbf{w_1^T}\mathbf{\Phi}(x)+b_1,
\end{equation*}
and
\begin{equation*}\label{Nf2}
f_2(x)=\mathbf{w_2^T}\mathbf{\Phi}(x)+b_2.
\end{equation*}

Similar to the linear case, the primal problems of the nonlinear TSVQR are as follows
\begin{eqnarray}\label{NTSVQR1}
\begin{aligned}
 \underset{\mathbf{w_1},b_1,\mathbf{\xi}}{\min}~&\frac{1}{2}(\|\mathbf{w_1}\|_2^2+ b_1^2)+ C_1\mathbf{e^T}\mathbf{\xi} + C_1\tau \mathbf{e^T}[\mathbf{Y}-\mathbf{w_1}^T \mathbf{\Phi(A)}-b_1\mathbf{e}]\\
 \text{s.t.\ } & \mathbf{Y}-\mathbf{w_1}^T \mathbf{\Phi(A)}-b_1\mathbf{e} \geq \varepsilon_1 \mathbf{e}-\mathbf{\xi}, ~\mathbf{\xi}\geq \mathbf{0},\\
\end{aligned}
\end{eqnarray}
and
\begin{eqnarray}\label{NTSVQR2}
\begin{aligned}
\underset{\mathbf{w_2},b_2,\mathbf{\xi}^*}{\min}~&\frac{1}{2}(\|\mathbf{w_2}\|_2^2+ \| b_2 \|^2)+ C_2\mathbf{e^T}\mathbf{\xi}^* + C_2(1-\tau){\mathbf{e^T}} [\mathbf{w_2}^T \mathbf{\Phi(A)}+b_2\mathbf{e}-\mathbf{Y}]\\
\text{s.t.\ } & \mathbf{w_2}^T \mathbf{\Phi(A)}+b_2\mathbf{e}-\mathbf{Y}\geq \varepsilon_2 \mathbf{e} - \mathbf{\xi}^*,~ \mathbf{\xi}^* \geq \mathbf{0}.\\
\end{aligned}
\end{eqnarray}

Define $H=[\mathbf{\Phi(A)}~\mathbf{e}]$, $\mathbf{u_1}=[\mathbf{w_1^T} ~b_1]^T$, $\mathbf{u_2}=[\mathbf{w_2^T}~ b_2]^T$.
Then, we get the following Lagrangian functions of (\ref{NTSVQR1}) and (\ref{NTSVQR2})
\begin{eqnarray*}\label{NL1}
\begin{aligned}
 L(\mathbf{u_1},\mathbf{\xi}, \mathbf{\alpha}, \mathbf{\beta}) = &\frac{1}{2}\|\mathbf{u_1}\|_2^2 + C_1\mathbf{e^T}\mathbf{\xi}
 + C_1\tau\mathbf{e^T}(\mathbf{Y}-\mathbf{u_1^TH}) \\
& -\mathbf{\alpha}^T (\mathbf{Y}-\mathbf{u_1^TH}-\varepsilon_1 \mathbf{e}+\mathbf{\xi})- \mathbf{\beta}^T \mathbf{\xi},\\
\end{aligned}
\end{eqnarray*}
and
\begin{eqnarray*}\label{NL2}
\begin{aligned}
 L(\mathbf{u_2},\mathbf{\xi^*}, \mathbf{\alpha^*}, \mathbf{\beta^*}) = &\frac{1}{2}\|\mathbf{u_2}\|_2^2 + C_2\mathbf{e^T}\mathbf{\xi^*}
 + C_2(1-\tau)\mathbf{e^T}(\mathbf{u_2^TH} - \mathbf{Y}) \\
& -\mathbf{\alpha^*}^T (\mathbf{u_2^TH}-\mathbf{Y}-\varepsilon_2 \mathbf{e}+\mathbf{\xi^*})- \mathbf{\beta^*}^T \mathbf{\xi^*},\\
\end{aligned}
\end{eqnarray*}
where $\mathbf{\alpha}$, $\mathbf{\beta}$, $\mathbf{\alpha^*}$, and $\mathbf{\beta^*}$  are the Lagrangian multiplier vectors.
The KKT necessary and sufficient optimality conditions
for the problems (\ref{NTSVQR1}) and (\ref{NTSVQR2}) are given by
\begin{eqnarray*}\label{NKKT1}
\begin{aligned}
& \mathbf{u_1}-C_1\tau \mathbf{H^T} \mathbf{e}+ \mathbf{H^T} \mathbf{\alpha}=0,\\
& C_1\mathbf{e}-\mathbf{\alpha}-\mathbf{\beta}=0, \\
& \mathbf{Y}-\mathbf{u_1^TH} \geq \varepsilon_1 \mathbf{e}-\mathbf{\xi}, ~ \mathbf{\xi} \geq \mathbf{0}, \\
& \mathbf{\alpha^T}(\mathbf{Y}-\mathbf{u_1^TH}-\varepsilon_1 \mathbf{e}+ \mathbf{\xi} )=0, ~ \mathbf{\alpha} \geq \mathbf{0}, \\
& \mathbf{\beta}^T \mathbf{\xi}= \mathbf{0}, ~\mathbf{\beta}\geq \mathbf{0},\\
\end{aligned}
\end{eqnarray*}
and
\begin{eqnarray*}\label{NKKT2}
\begin{aligned}
& \mathbf{u_2}+C_2(1-\tau) \mathbf{H^T} \mathbf{e} - \mathbf{H^T} \mathbf{\alpha^*}=0,\\
& C_2\mathbf{e}-\mathbf{\alpha^*}-\mathbf{\beta^*}=0, \\
& \mathbf{u_2^TH}-\mathbf{Y} \geq \varepsilon_2 \mathbf{e}-\mathbf{\xi^*}, ~\mathbf{\xi^*} \geq \mathbf{0}, \\
& \mathbf{\alpha^*}^T(\mathbf{u_2^TH}-\mathbf{Y}-\varepsilon_2 \mathbf{e}+ \mathbf{\xi^*} )=0, ~\mathbf{\alpha^*} \geq \mathbf{0}, \\
& \mathbf{\beta^*}^T \mathbf{\xi^*}= \mathbf{0}, ~\mathbf{\beta^*} \geq \mathbf{0}.\\
\end{aligned}
\end{eqnarray*}

Using the above KKT conditions, we derive the dual problems of (\ref{NTSVQR1}) and (\ref{NTSVQR2})
\begin{eqnarray}\label{DNTSVQR1}
\begin{aligned}
 \underset{\mathbf{\alpha}}{\min}~&\frac{1}{2}\mathbf{\alpha^T}\mathbf{H}\mathbf{H^T}\mathbf{\alpha}-C_1\tau \mathbf{e^T}\mathbf{H}\mathbf{H^T}\mathbf{\alpha}+\mathbf{Y^T}\mathbf{\alpha}-\varepsilon_1 \mathbf{e^T}\mathbf{\alpha}\\
 \text{s.t.\ } & \mathbf{0} \leq \mathbf{\alpha} \leq C_1 \mathbf{e},\\
\end{aligned}
\end{eqnarray}
and
\begin{eqnarray}\label{DNTSVQR2}
\begin{aligned}
 \underset{\mathbf{\alpha^*}}{\min}~&\frac{1}{2}\mathbf{{\alpha^*}^T}\mathbf{H}\mathbf{H^T}\mathbf{\alpha^*}-C_2(1-\tau) \mathbf{e^T}\mathbf{H}\mathbf{H^T}\mathbf{\alpha^*}-\mathbf{Y^T}\mathbf{\alpha^*}-\varepsilon_2 \mathbf{e^T}\mathbf{\alpha^*}\\
 \text{s.t.\ } &  \mathbf{0} \leq \mathbf{\alpha^*} \leq C_2 \mathbf{e}.\\
\end{aligned}
\end{eqnarray}

Similarly, we us DCDM to solve problems (\ref{DNTSVQR1}) and (\ref{DNTSVQR2}).
Define $\bar{\mathbf{H}}=\mathbf{H}\mathbf{H^T}=\mathbf{K(A,A^T)}+\mathbf{e}\mathbf{e}^T$,
$\mathbf{d_1}= C_1\tau\mathbf{H}\mathbf{H^T}\mathbf{e}-\mathbf{Y}+\varepsilon_1\mathbf{e} $, and
$\mathbf{d_2}=C_2(1-\tau)\mathbf{H}\mathbf{H^T}\mathbf{e}+\mathbf{Y}+\varepsilon_2\mathbf{e}$.
$\mathbf{K(A,A^T)}=\mathbf{\Phi(A)}\mathbf{\Phi(A^T)}$ is the kernel function. Then, (\ref{DNTSVQR1}) and (\ref{DNTSVQR2}) can be rewritten as
\begin{eqnarray*}\label{RDTSVQR1}
\begin{aligned}
 \underset{\mathbf{\alpha}}{\min}~~f(\mathbf{\alpha})=~&\frac{1}{2}\mathbf{\alpha^T}\bar{\mathbf{H}}\mathbf{\alpha}-\mathbf{d^T_1}\mathbf{\alpha}\\
 \text{s.t.\ } & \mathbf{0} \leq \mathbf{\alpha} \leq C_1 \mathbf{e},\\
\end{aligned}
\end{eqnarray*}
and
\begin{eqnarray*}\label{RDTSVQR2}
\begin{aligned}
 \underset{\mathbf{\alpha^*}}{\min}~~f^*(\mathbf{\alpha^*})=~&\frac{1}{2}\mathbf{\alpha^*}^T\bar{\mathbf{H}}\mathbf{\alpha^*}-\mathbf{d^T_2}\mathbf{\alpha^*}\\
 \text{s.t.\ } & \mathbf{0} \leq \mathbf{\alpha^*} \leq C_2 \mathbf{e}.\\
\end{aligned}
\end{eqnarray*}
DCDM starts with a random initial points $\mathbf{\alpha}^0$ and $\mathbf{\alpha^*}^0$. For updating $\mathbf{\alpha}$ and $\mathbf{\alpha^*}$,
we solve the following one-variable sub-problems
\begin{eqnarray}\label{RRDNTSVQR1}
\begin{aligned}
 & \underset{\mathbf{t}}{\min}~~f(\mathbf{\alpha_i}+tI_i)\\
 & \text{s.t.\ }  0 \leq \mathbf{\alpha_i}+t \leq C_1,~i=1, 2, \cdots, l,\\
\end{aligned}
\end{eqnarray}
and
\begin{eqnarray}\label{RRDNTSVQR2}
\begin{aligned}
 & \underset{\mathbf{t^*}}{\min}~~f^*(\mathbf{\alpha^{*}_i}+t^*I_i)\\
& \text{s.t.\ } 0 \leq \mathbf{\alpha^{*}_i} + t^*\leq C_2,~i=1, 2, \cdots, l.\\
\end{aligned}
\end{eqnarray}

\begin{figure}[htbp]
\centering
\includegraphics[width=0.6\textheight]{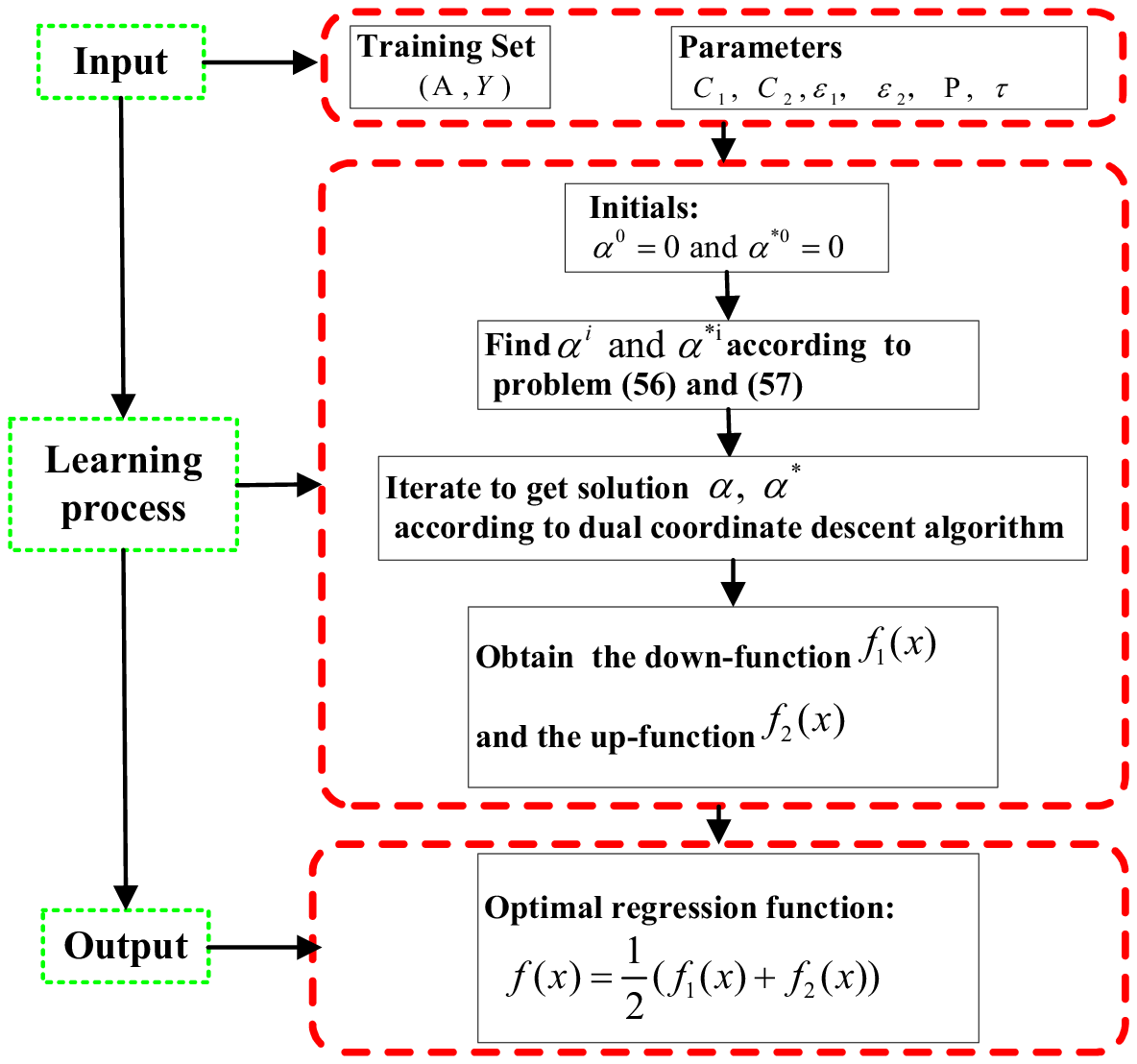}\\
\caption{Flowchart of the solution of nonlinear TSVQR.} \label{flowchart}
\end{figure}

The objective functions of (\ref{RRDNTSVQR1}) and (\ref{RRDNTSVQR2}) are simplified as
\begin{eqnarray*}\label{SRDTSVQR1}
\begin{aligned}
 & f(\mathbf{\alpha_i}+tI_i)=\frac{1}{2}\bar{\mathbf{H}}_{ii}t^2+ \nabla_if(\alpha_i)t+f(\alpha_i) ,~i=1, 2, \cdots, l,\\
\end{aligned}
\end{eqnarray*}
and
\begin{eqnarray*}\label{SRDTSVQR2}
\begin{aligned}
f^*(\mathbf{\alpha^{*}_i}+t^*I_i)=\frac{1}{2}\bar{\mathbf{H}}_{ii}{t^*}^2+ \nabla_if^*(\alpha^*_i)t^*+f^*(\alpha^*_i),~i=1, 2, \cdots, l.\\
\end{aligned}
\end{eqnarray*}
We combine the bounded constraints $0 \leq \mathbf{\alpha_i}+t \leq C_1$ and $0 \leq \mathbf{\alpha^{*}_i} + t^*\leq C_2$,
and get the following bounded constraint solutions of (\ref{RRDNTSVQR1}) and (\ref{RRDNTSVQR2})
 \begin{eqnarray*}\label{STSVQR1}
\begin{aligned}
 & \alpha^{new}_i= min(max(\alpha_i-\frac{\nabla_i f(\alpha_i)}{\mathbf{\bar{H}}_{ii}},0), C_1),\\
\end{aligned}
\end{eqnarray*}
and
 \begin{eqnarray*}\label{STSVQR2}
\begin{aligned}
  & \alpha^{*new}_i= min(max(\alpha^*_i-\frac{\nabla_i f^*(\alpha^*_i)}{\mathbf{\bar{H}}_{ii}},0), C_2).\\
\end{aligned}
\end{eqnarray*}
Then, we obtain Algorithm \ref{Alg:2} to solve nonlinear TSVQR in detail.
Finally, we obtain the down- and up-bound functions $f_1(x)$ and $f_2(x)$, and then construct the final decision function $f(x)= \frac{1}{2}[f_1(x)+f_2(x)]$. The flowchart of the solution of nonlinear TSVQR is summarized in Fig. \ref{flowchart}.
\begin{algorithm}
\SetAlgoNoLine
\KwIn{Training set $(\mathbf{A},\mathbf{Y})$; Parameters $C_1$, $C_2$, $\tau$, $\varepsilon_1$ and $\varepsilon_2$, and the kernel parameter $P$;}
\KwOut{Updated $\mathbf{\alpha}$ and $\mathbf{\alpha^*}$;}
\Begin{Set $\mathbf{\alpha}=\mathbf{0}$ and $\mathbf{\alpha^*}=\mathbf{0}$;\\
Compute $\mathbf{d}= C_1\tau\mathbf{H}\mathbf{H^T}\mathbf{e}-\mathbf{Y}+\varepsilon_1\mathbf{e}$,
$\mathbf{d^*}=C_2(1-\tau)\mathbf{H}\mathbf{H^T}\mathbf{e}+\mathbf{Y}+\varepsilon_2\mathbf{e}$, and $\bar{\mathbf{H}}=\mathbf{H}\mathbf{H^T}$;\\
\If {$\mathbf{\alpha}$ not converge;}
{do for $i=1,2,\cdots, l$ do \\
$\nabla_i f(\alpha_i)\leftarrow \alpha^T \bar{\mathbf{H}}_{ii}I_i-d^TI_i$\\
$\alpha_i^{old} \leftarrow \alpha_i$;\\
$\alpha_i \leftarrow min(max(\alpha_i-\frac{\nabla_i f(\alpha_i)}{\mathbf{\bar{H}}_{ii}},0), C_1)$;}
\If {$\mathbf{\alpha^*}$ not converge;}
{do for $i=1,2,\cdots, l$ do \\
$\nabla_i f(\alpha^*_i)\leftarrow \alpha^{*T} \bar{\mathbf{H}}_{ii}I_i-d^{*T}I_i$\\
$\alpha^{*old}_i \leftarrow \alpha^*_i$;\\
$\alpha^{*}_i \leftarrow min(max(\alpha^*_i-\frac{\nabla_i f^*(\alpha^*_i)}{\mathbf{\bar{H}}_{ii}},0), C_2)$;}
}
\caption{ Dual coordinate descent algorithm for nonlinear TSVQR.} \label{Alg:2}
\end{algorithm}

\subsection{Discussion}
TSVQR follows the spirit of $\varepsilon$-TSVR (\citeauthor{shao2013varepsilon}, \citeyear{shao2013varepsilon}), and generates two nonparallel functions to determine the $\varepsilon$-insensitive lower and upper bounds for capturing the asymmetric information in data points. Moreover, TSVQR brings the spirit of QR (\citeauthor{dempster1977simulation}, \citeyear{dempster1977simulation}) into TSVR, and uses the quantile parameter to compute a family of regression curves corresponding to various percentage points of the distributions. As the quantile parameter increases, the regression curves move up through the data to reflect the heterogeneous information in data points. TSVQR thus effectively captures the heterogeneous and asymmetric information in data points simultaneously.

Concerning the computational complexity of TSVQR, we find that the major computational cost of TSVQR comes from the kernel matrix.
Note that $\mathbf{A}$ is an $l \times n$ matrix. TSVQR and URALTSVR only solve two smaller-sized QPPs in which DCDM is applied.
Thus, the computational complexity of Algorithm \ref{Alg:2} is estimated as $2O(l^2)$. Although SVQR only solves one single QPP,
the QPP in SVQR has two groups of constraints for all data. The computational complexity of SVQR, $\varepsilon$-SVQR, Online-SVQR and RQSVR is approximately $O(2l)^3$. Thus, TSVQR and URALTSVR work faster than SVQR, $\varepsilon$-SVQR, Online-SVQR, and RQSVR. Table \ref{merit} shows the important merit and demerit of SVQR, $\varepsilon$-SVQR, GPQR, Online-SVQR, RQSVR, URALTSVR, and TSVQR.
\begin{table}
\tiny
 \begin{center}
  \caption{The merit and demerit of SVQR, $\varepsilon$-SVQR, GPQR, Online-SVQR, RQSVR, URALTSVR, and TSVQR.}
  \label{merit}
  \begin{tabular}{@{}lll@{}}
   \hline
   Methods          & Merits                        & Demerits/ Limitations  \\
   \hline
     SVQR              & $\bullet$ Using the quantile parameter.       & $\bullet$ High computation time due to large QPP. \\
   (\citeauthor{li2007quantile}, \citeyear{li2007quantile})         &         & $\bullet$ Does not solve the asymmetry problem. \\
   \hline
   $\varepsilon$-SVQR  & $\bullet$ Using the quantile parameter.             & $\bullet$ High computation time due to large QPP.  \\
   (\citeauthor{anand2020new}, \citeyear{anand2020new}) & $\bullet$ Using $\varepsilon$-insensitive loss.  & $\bullet$ Does not solve the asymmetry problem. \\
   \hline
   GPQR                & $\bullet$ Using the quantile parameter.             & $\bullet$ Does not solve the asymmetry problem.\\
   (\citeauthor{ouhourane2022group}, \citeyear{ouhourane2022group})       & $\bullet$ Sparseness.                           & \\
   \hline
   Online-SVQR         & $\bullet$ Using the quantile parameter.             & $\bullet$ High computation time due to large QPP. \\
  (\citeauthor{ye2021online}, \citeyear{ye2021online})              & $\bullet$ Using $\varepsilon$-insensitive loss.  & $\bullet$ Does not solve the asymmetry problem. \\
                       & $\bullet$ Reflecting the dynamic information.   &                                                \\
   \hline
   RQSVR               & $\bullet$ Using pinball loss function.       & $\bullet$ High computation time due to large QPP. \\
  (\citeauthor{yang2019robust},\citeyear{yang2019robust})      & $\bullet$ Robustness.                         & $\bullet$ Does not solve the asymmetry problem. \\
   \hline
   URALTSVR            & $\bullet$ Using pinball loss function.       & $\bullet$ Lack of sparseness.    \\
   (\citeauthor{gupta2021robust}, \citeyear{gupta2021robust})          & $\bullet$ Less computation time by using the &               \\
                       &~~gradient based iterative approaches.        &               \\
                       & $\bullet$ Controlling the fitting error   &               \\
                       &~~inside the asymmetric tube.               &               \\
  \hline
  TSVQR                & $\bullet$ Using $\varepsilon$-insensitive loss. &  $\bullet$ Lack of sparseness. \\
                       & $\bullet$ Depicting heterogenous information.      &   \\
                       & $\bullet$ Reflecting asymmetric information.         & \\
                       & $\bullet$ Fast processing by solving two              & \\
                       &~~small QPPs.                                  & \\
  \hline
  \end{tabular}
 \end{center}
\end{table}
\section{Experiments}\label{Sec:4}
To test the efficiency of the proposed TSVQR, we compare with SVQR, $\varepsilon$-SVQR, Online-SVQR, GPQR, RQSVR, URALTSVR, and least squares large margin distribution machine-based regression (LS-LDMR) (\citeauthor{gupta2021least}, \citeyear{gupta2021least}) on several data sets,
including six artificial data sets and five benchmark data sets. All of these methods are implemented
in the MATLAB R2021a environment on a PC running the 64-bit Windows XP OS and 16 GB of RAM.

\subsection{Performance criteria}
Suppose $m$ is the number of testing samples,
$y^t_i$ is the $i$th test sample, $\hat{y}^t_i$ is the predicted value of $y^t_i$, and ${\bar{y}}^t=\frac{1}{m}\sum\limits_{i=1}^{m}y^t_{i}$ is the average value of $y^t_1,\cdots,y^t_m$.
For the nonlinear version of TSVQR, we employ a Gaussian kernel, and its kernel parameter $P$ is selected from the set $\{ 2^{-8}, \dots , 2^8 \}$. Parameters $C_1$ and $C_2$ are selected from the set $\{ 2^{-8}, \dots , 2^8 \}$.
The quantile parameter is chosen from the set $\{0.10, 0.25, 0.50, 0.75, 0.90 \}$.
The insensitive parameters $\varepsilon_1$ and $\varepsilon_2$ are chosen from $0.01$ to $0.1$ with a fixed step size of $0.01$. The optimal values of the parameters in our experiments are obtained by using the grid search method.
We use the generalized approximate cross validation (GACV) criterion (\citeauthor{yuan2006gacv}, \citeyear{yuan2006gacv}; \citeauthor{xu2015weighted}, \citeyear{xu2015weighted}) to select the optimal values of the parameters.
 The definition of GACV is as follows
\begin{eqnarray*}\label{GACV}
\begin{array}{ll}
GACV_{\tau}=\frac{1}{l-|I_{SV}|}\sum_{i=1}^{l}\rho_{\tau}{(y^t_i-\hat{y}^t_i)}
 \end{array}
\end{eqnarray*}
where $|I_{SV}|$ is the cardinality of the set $I_{SV}$, and $\rho_{\tau}(r)$ is the pinball loss function of Koenker and Bassett (\citeyear{koenker1978regression})
\begin{equation*}\label{pinball loss}
\rho_{\tau}{(r)}=
\begin{cases}
\tau r, & if~ r > 0,\\
-(1-\tau) r, & otherwise.
\end{cases}
\end{equation*}
To evaluate the performance of the proposed method, we use the following evaluation criteria,
namely the empirical quantile risk (Risk), the root mean square error (RMSE), the mean absolute error (MAE) and the mean absolute percentage error (MAPE).
These evaluation criteria are defined as follows
\begin{eqnarray*}
\begin{aligned}
&\operatorname{Risk_{\tau}}=\frac{1}{m} \sum_{i=1}^{m} \rho_{\tau}(y^t_i-\hat{y}^t_i), \\
\end{aligned}
\end{eqnarray*}
\begin{eqnarray*}
\begin{aligned}
&\operatorname{RMSE}=\sqrt{\frac{1}{m} \sum_{i=1}^{m}(y^t_i-\hat{y}^t_i)^{2}}, \\
\end{aligned}
\end{eqnarray*}
\begin{eqnarray*}
\begin{aligned}
&\operatorname{MAE}=\frac{1}{m} \sum_{i=1}^{m} |y^t_i-\hat{y}^t_i|, \\
\end{aligned}
\end{eqnarray*}
\begin{eqnarray*}
\begin{aligned}
&\operatorname{MAPE}=\frac{\sum_{i=1}^{m} |y^t_i-\hat{y}^t_i|  /y^t_{i}} {m}.
\end{aligned}
\end{eqnarray*}
\subsection{Artificial data sets}
We first construct six artificial data sets to test the performance of TSVQR whose definitions are given in Table \ref{Function}. The specifications of these artificial data sets are listed in Table \ref{dataset}. The training samples of Type $A_3$ and Type $B_3$ contain outliers.
According to the values of GACV, we select the best parameters of TSVQR, and then list them in Table \ref{par1}.
\begin{table}
 \footnotesize
 \begin{center}
  \caption{Function used for generating artificial data sets.}
  \label{Function}
  \begin{tabular}{@{}ccl@{}}
   \hline
   Function definition&Domain of definition&Noise type \\
   \hline
   &&Type $A_1$: $\xi_i \sim\chi^{2}(3)$\\
 $y = (1-x+2 x^{2}) e^{-0.5 x^{2}}+{\frac{1}{5}(1+0.2 x)}\xi$&$x \in [-4,4]$&Type $A_2$: $\xi_i \sim\chi^{2}(5)$ \\
   &&Type $A_3$: Laplacian Noise \\
   \hline
   &&Type $B_1$: $\xi_i \sim N(0.3,0.6^{2})$\\
   $y = 6sin(0.5\pi-x)+3(sin(0.5\pi-x)) {\xi}$&$x \in [-4,4]$&Type $B_2$: $\xi_i \sim N(0.5,0.8^{2})$\\
   &&Type $B_3$: Laplacian Noise \\
   \hline
  \end{tabular}
 \end{center}
\end{table}

\begin{table}
\footnotesize
\begin{center}
\caption{Descriptive statistics of data sets.}
\label{dataset}
\begin{tabular}{@{}ccc@{}}
\hline
Data set&Number of training samples&Number of testing samples  \\
\midrule
Type $A_1$ & 401 & 400 \\
Type $A_2$ & 401 & 400 \\
Type $A_3$ & 405 & 400 \\
Type $B_1$ & 801 & 161 \\
Type $B_2$ & 801 & 161  \\
Type $B_3$ & 805 & 400 \\
Engel &187&48 \\
Bone density&388&97 \\
US girls&3209&802  \\
Motorcycle&99&33  \\
Boston housing&405&101 \\
\hline
\end{tabular}
\end{center}
\end{table}

\begin{table}
	\footnotesize
	\centering
	\caption{The optimal parameters selected of artificial data sets by GACV.}
	\begin{tabular}{cccccc}
		\hline
		$\tau$&0.10&0.25&0.50&0.75&0.90\\
		\hline
		Type $A_1$&&&&&\\
		$\rm C_1$&$2^3$&$2^3$&$2^3$&$\rm 2^{-2}$& $2^3$\\
		$\rm C_2$&$2^3$&$2^3$&$2^3$&$\rm 2^{-2}$&$2^3$ \\
		P&$\rm 2^0$&$\rm 2^0$&$\rm 2^0$&$\rm 2^0$&$\rm 2^0$\\
		\hline
		Type $A_2$&&&&&\\
		$\rm C_1$&$2^3$&$2^3$&$2^3$&$\rm 2^1$& $2^3$\\
		$\rm C_2$&$2^3$&$2^3$&$2^3$&$\rm 2^1$&$2^3$ \\
		P&$\rm 2^0$&$\rm 2^0$&$\rm 2^0$&$\rm 2^0$&$\rm 2^0$\\
		\hline
		Type $A_3$&&&&&\\
		$\rm C_1$&$2^3$&$2^3$&$2^3$&$\rm 2^3$& $2^3$\\
		$\rm C_2$&$2^3$&$2^3$&$2^3$&$\rm 2^3$&$2^3$ \\
		P&$\rm 2^0$&$\rm 2^0$&$\rm 2^0$&$\rm 2^0$&$\rm 2^0$\\
		\hline
		Type $B_1$&&&&&\\
		$\rm C_1$&$\rm 2^1$&$\rm 2^2$&$\rm 2^1$&$\rm 2^3$&$\rm 2^3$ \\
		$\rm C_2$&$\rm 2^1$&$\rm 2^2$&$\rm 2^1$&$\rm 2^3$& $\rm 2^3$\\
		P&$\rm 2^1$&$\rm 2^1$&$\rm 2^1$&$\rm 2^1$&$\rm 2^1$\\
		\hline
		Type $B_2$&&&&&\\
		$\rm C_1$&$\rm 2^1$&$\rm 2^2$&$\rm 2^1$&$\rm 2^3$&$\rm 2^3$ \\
		$\rm C_2$&$\rm 2^1$&$\rm 2^2$&$\rm 2^1$&$\rm 2^3$& $\rm 2^3$\\
		P&$\rm 2^1$&$\rm 2^1$&$\rm 2^1$&$\rm 2^1$&$\rm 2^1$\\
		\hline
		Type $B_3$&&&&&\\
		$\rm C_1$&$\rm 2^3$&$\rm 2^3$&$\rm 2^2$&$\rm 2^3$&$\rm 2^3$ \\
		$\rm C_2$&$\rm 2^3$&$\rm 2^3$&$\rm 2^2$&$\rm 2^3$& $\rm 2^3$\\
		P&$\rm 2^1$&$\rm 2^1$&$\rm 2^1$&$\rm 2^1$&$\rm 2^1$\\
		\hline
	\end{tabular}
	\label{par1}
\end{table}

\begin{figure*}[h!]
\centering
\subfigure[SVQR]{\includegraphics[width=0.35\textheight]{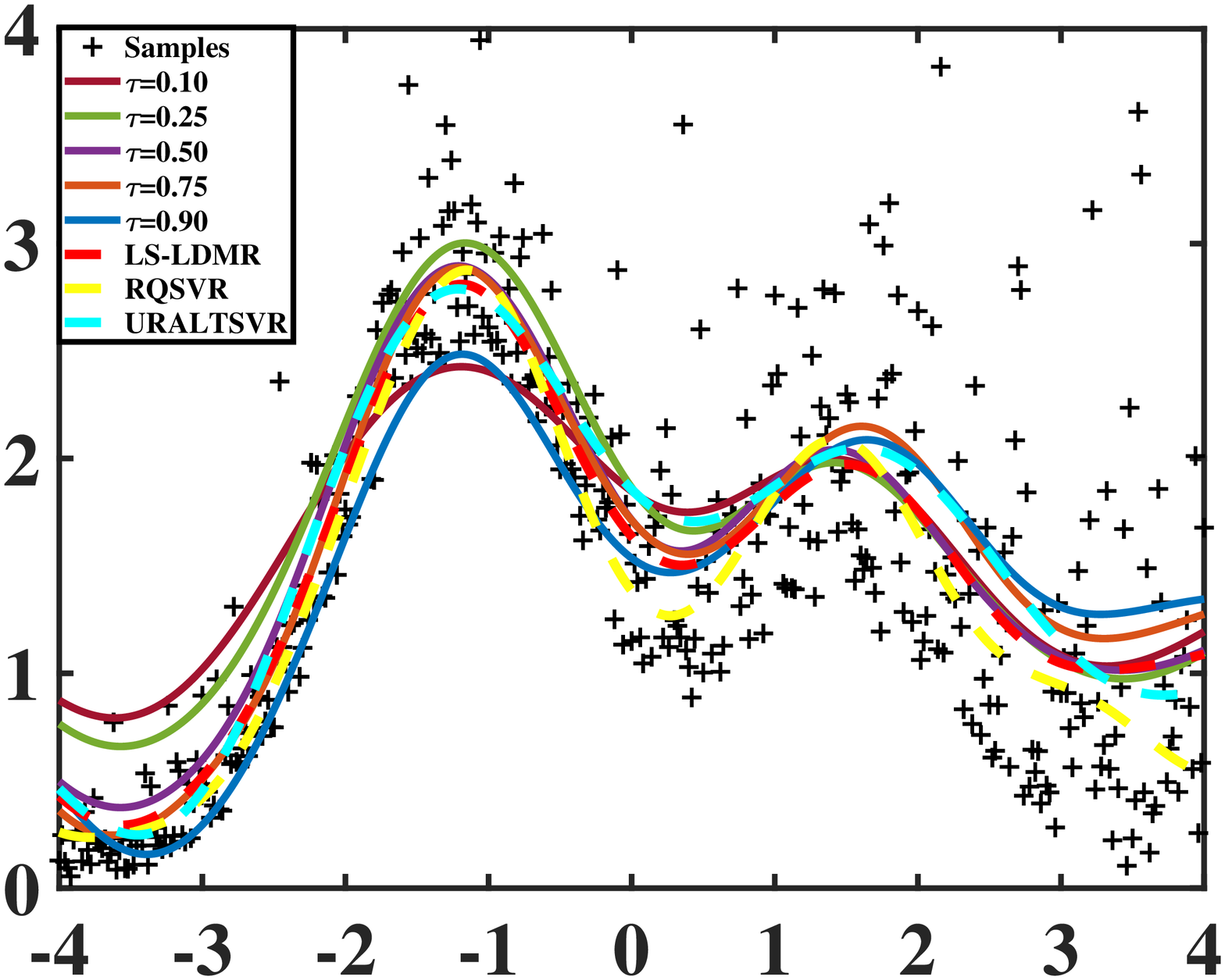}}
\subfigure[$\varepsilon$-SVQR]{\includegraphics[width=0.35\textheight]{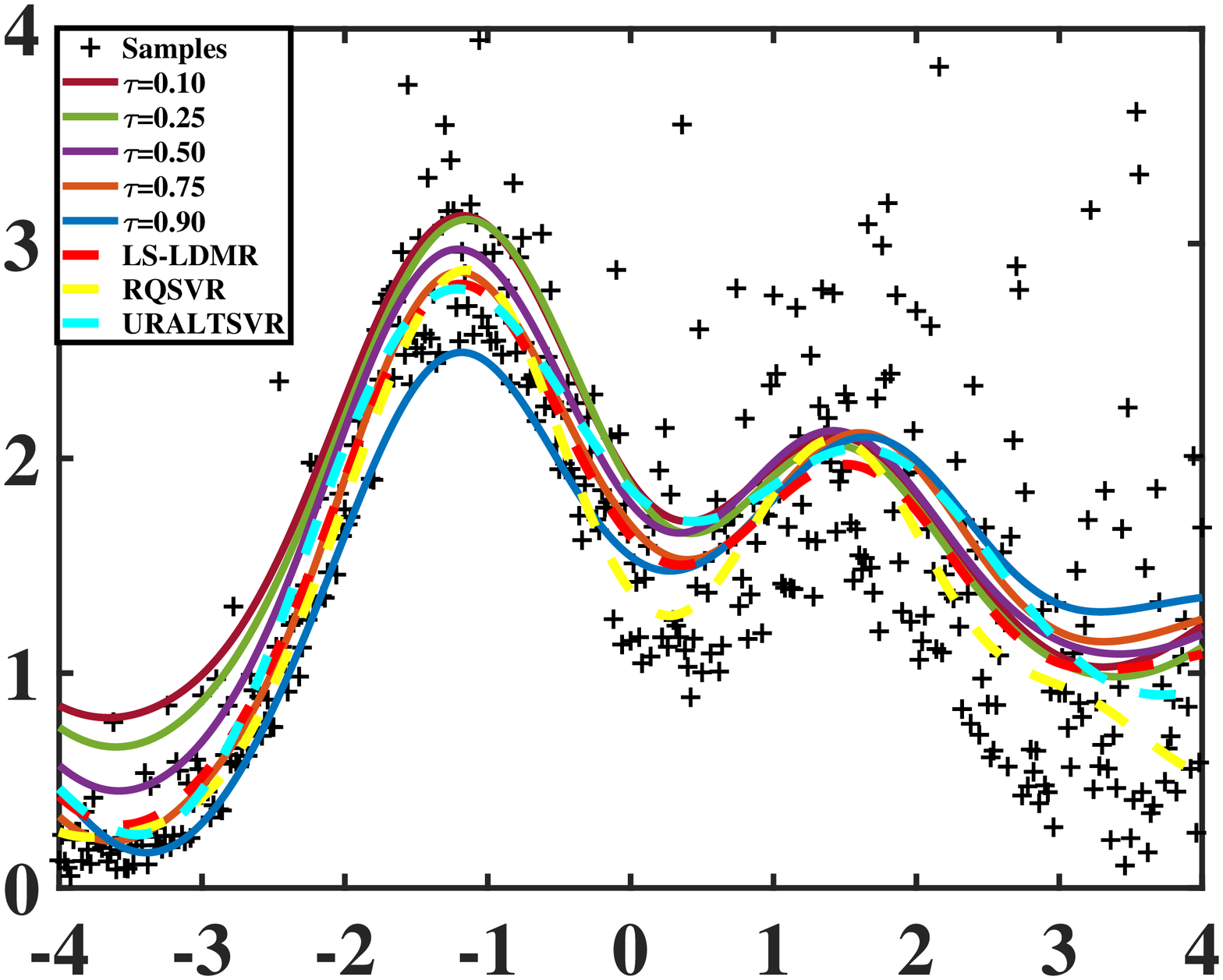}}
\subfigure[Online-SVQR]{\includegraphics[width=0.35\textheight]{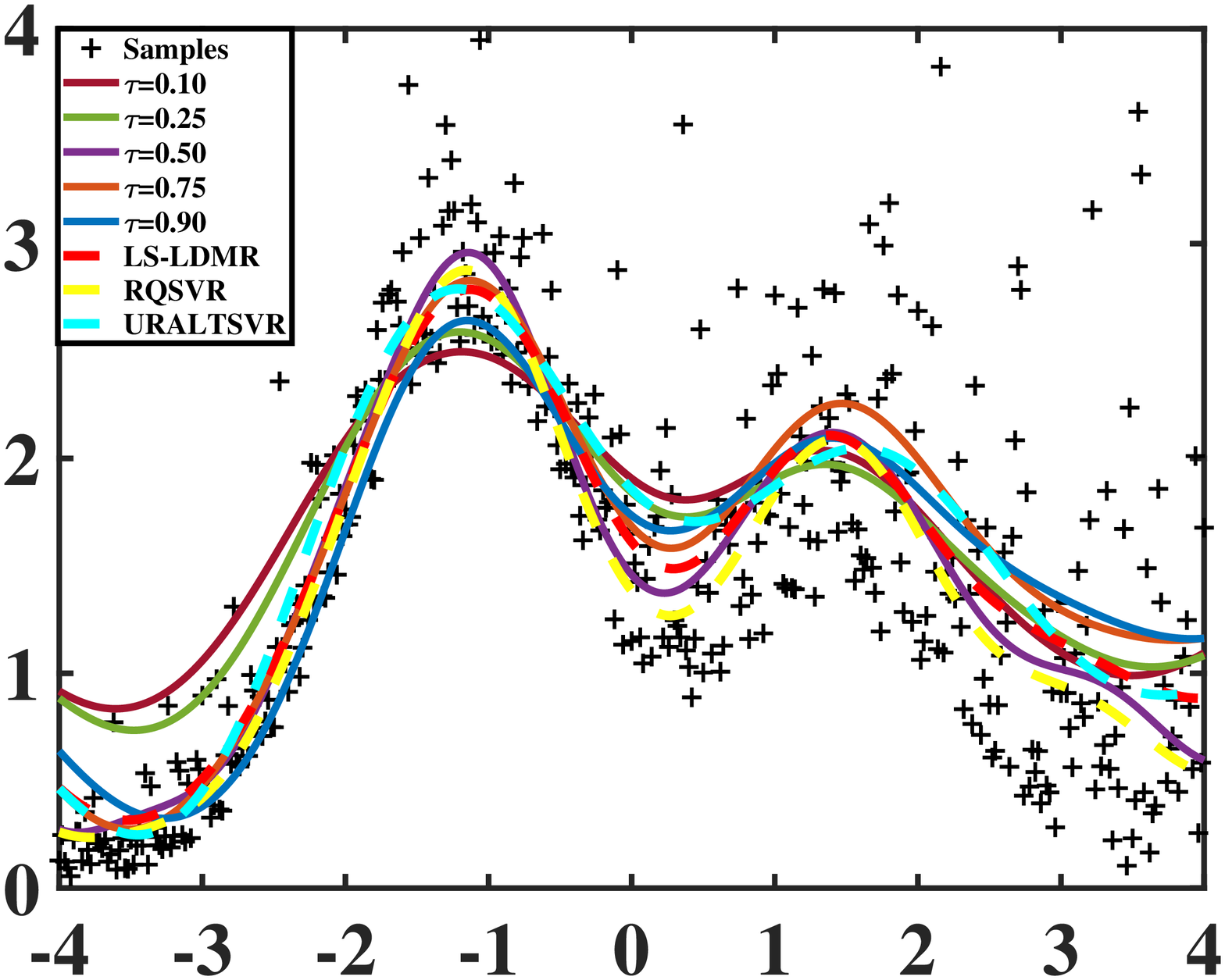}}
\subfigure[GPQR]{\includegraphics[width=0.35\textheight]{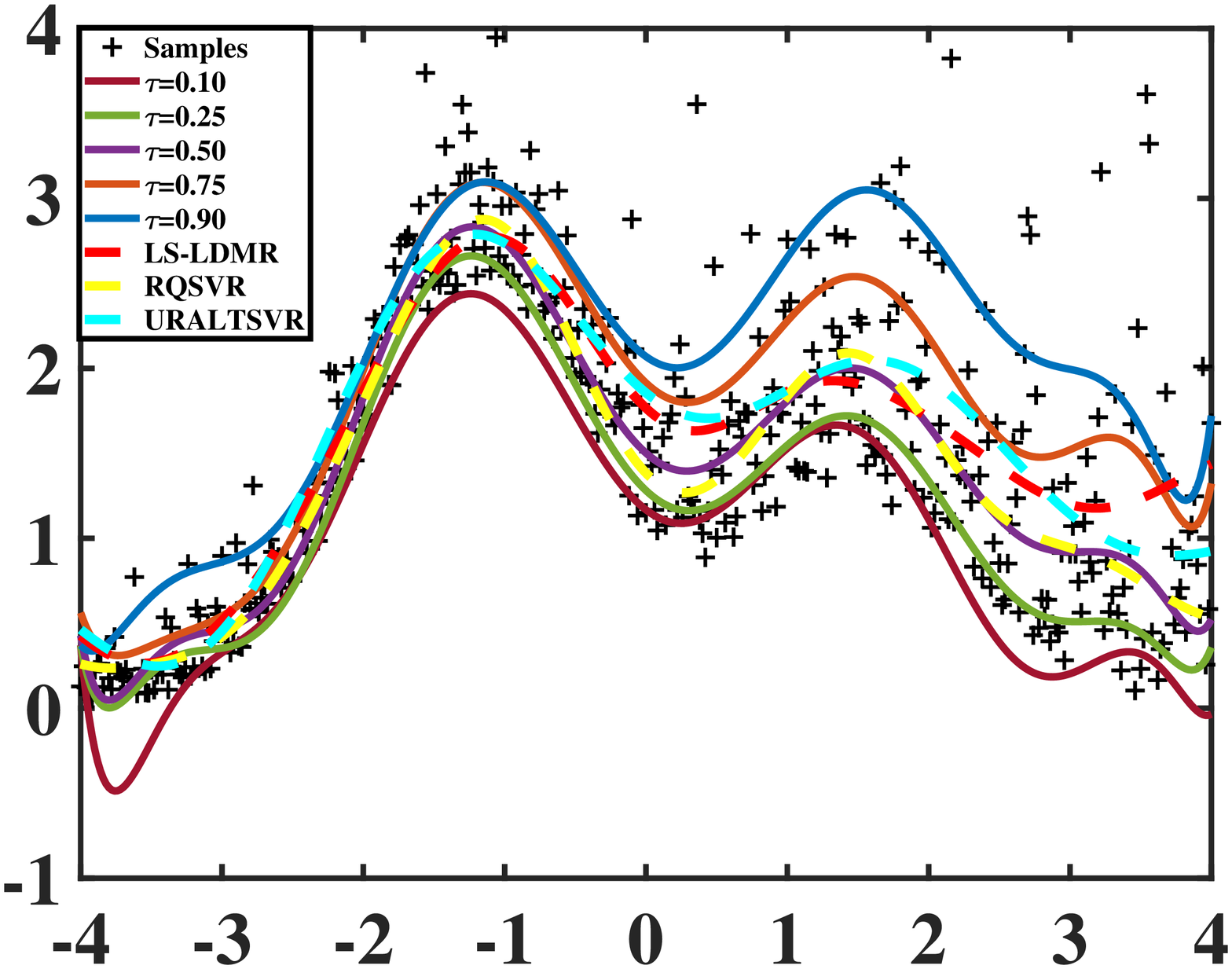}}
\subfigure[TSVQR]{\includegraphics[width=0.35\textheight]{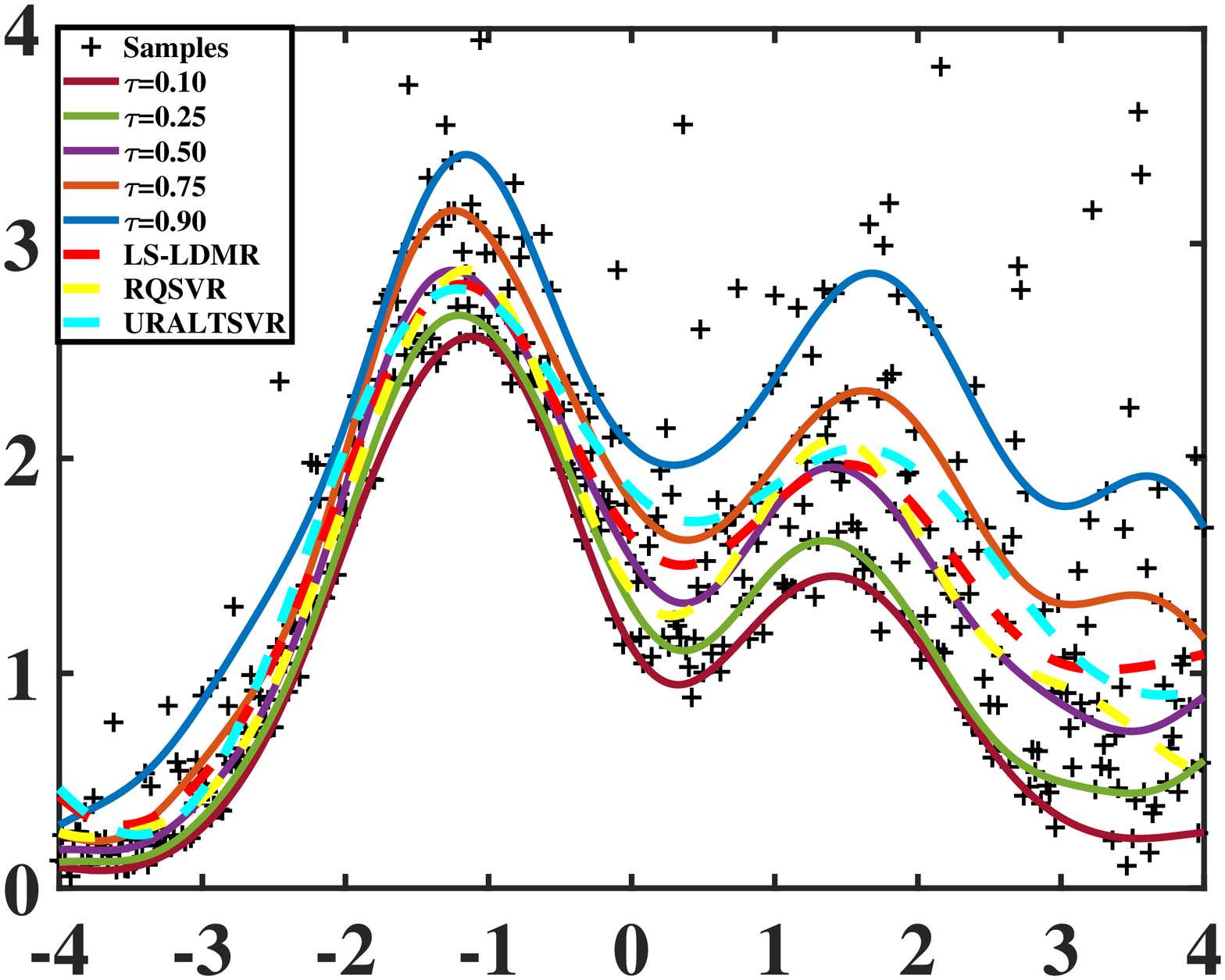}}
\caption{Regression results of SVQR, $\varepsilon$-SVQR, Online-SVQR, GPQR, TSVQR, LS-LDMR, RQSVR, and URALTSVR for Type $A_1$.}\label{fexp}
\end{figure*}

\begin{figure*}[h!]
\centering
\subfigure[SVQR]{\includegraphics[width=0.35\textheight]{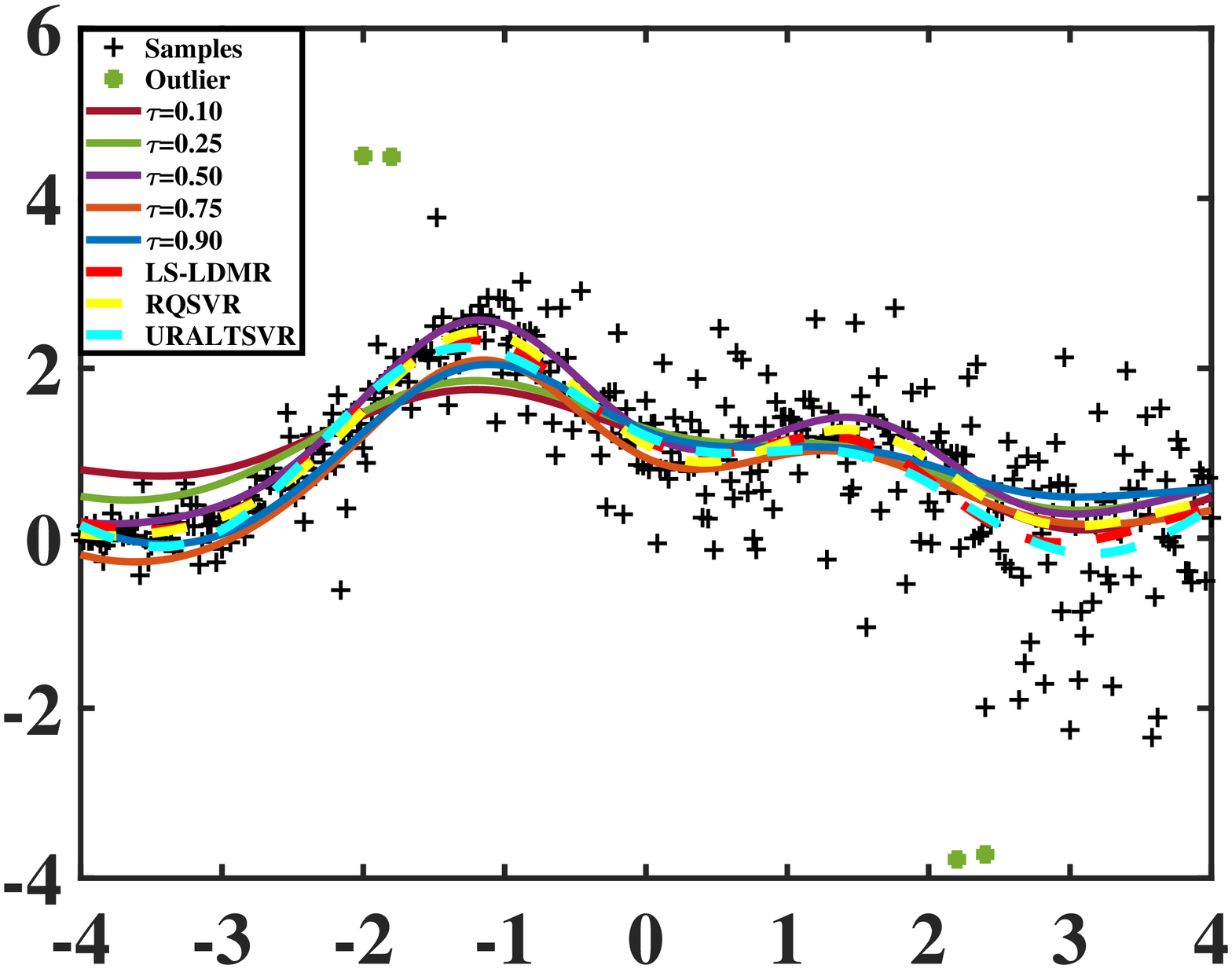}}
\subfigure[$\varepsilon$-SVQR]{\includegraphics[width=0.35\textheight]{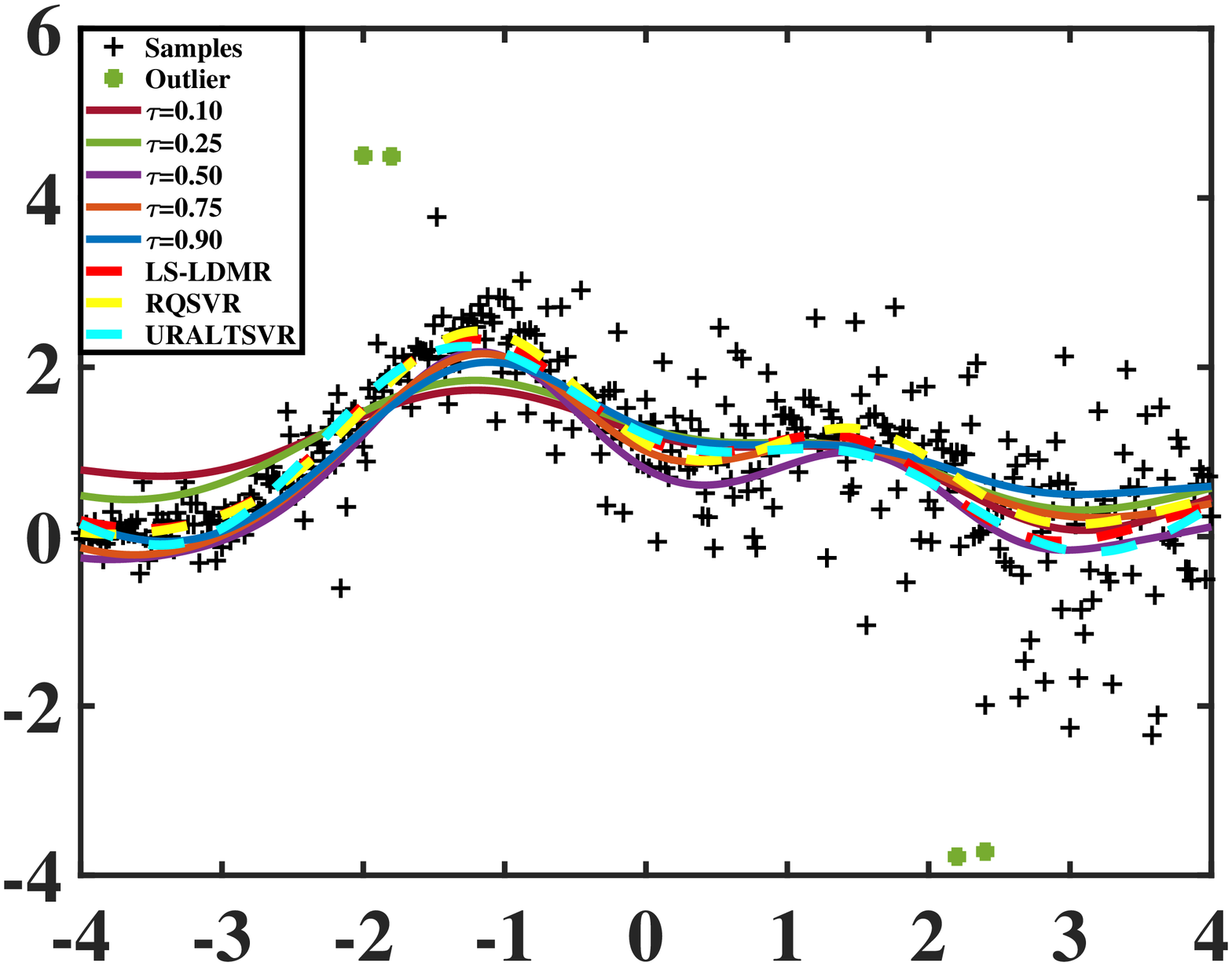}}
\subfigure[Online-SVQR]{\includegraphics[width=0.35\textheight]{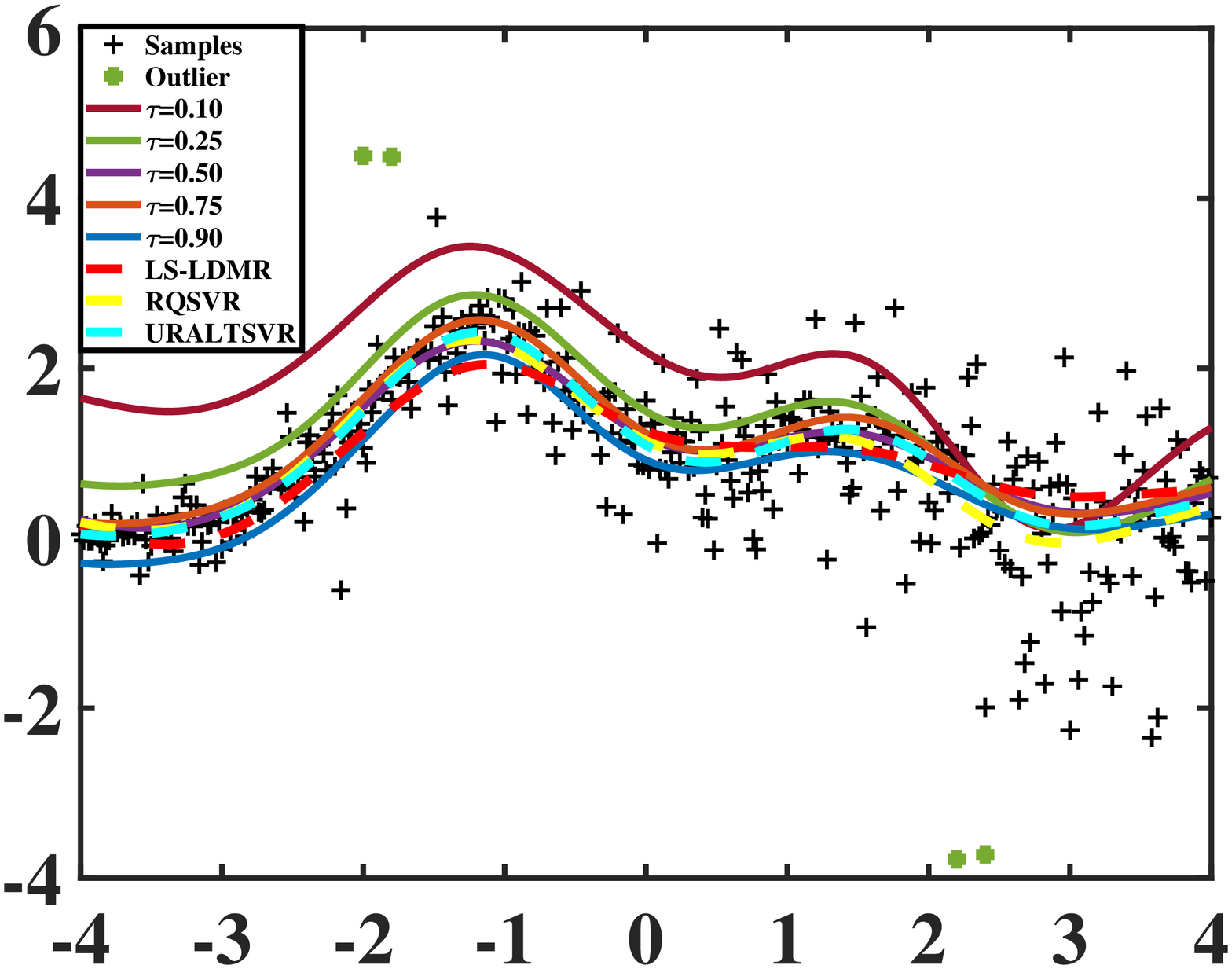}}
\subfigure[GPQR]{\includegraphics[width=0.35\textheight]{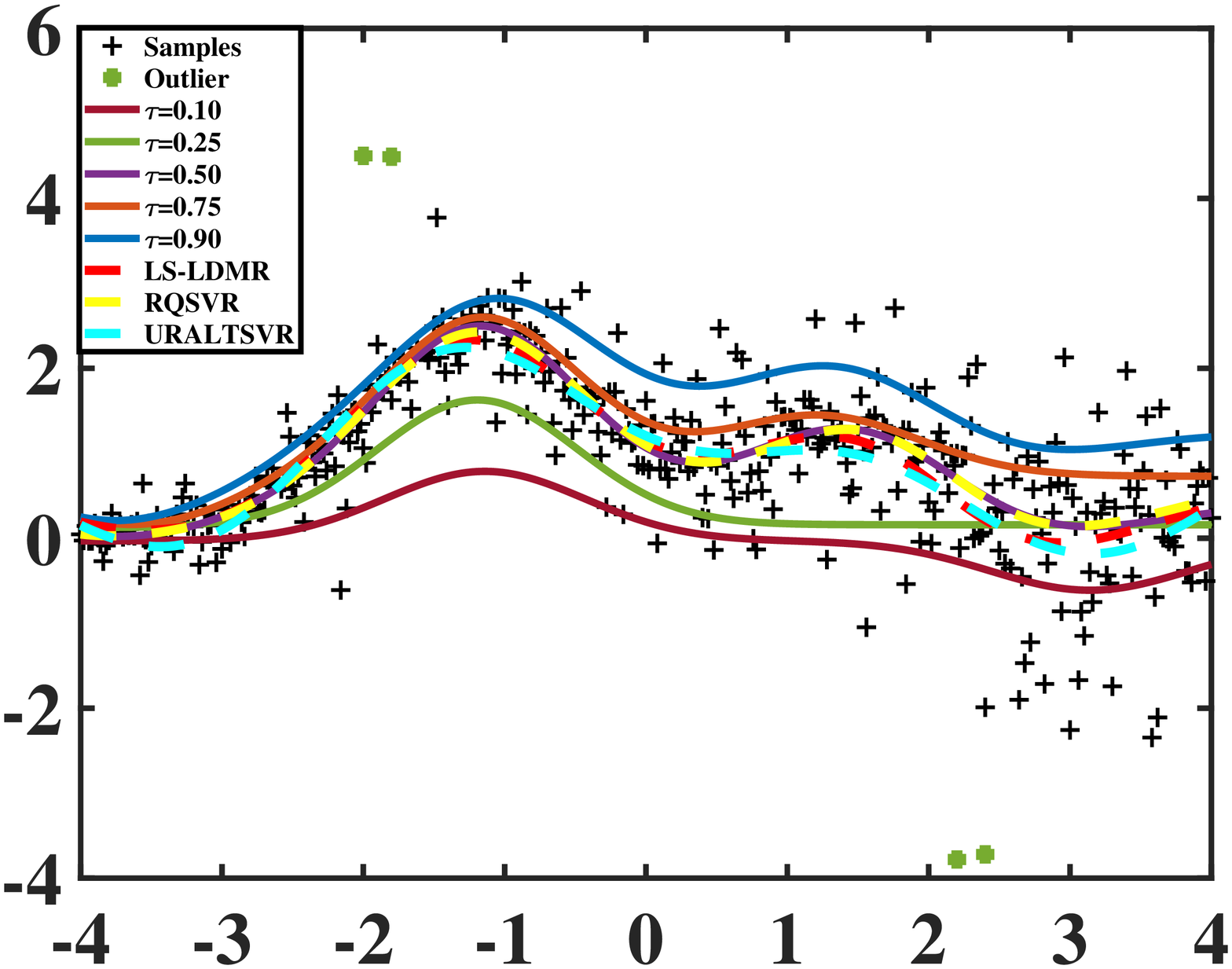}}
\subfigure[TSVQR]{\includegraphics[width=0.35\textheight]{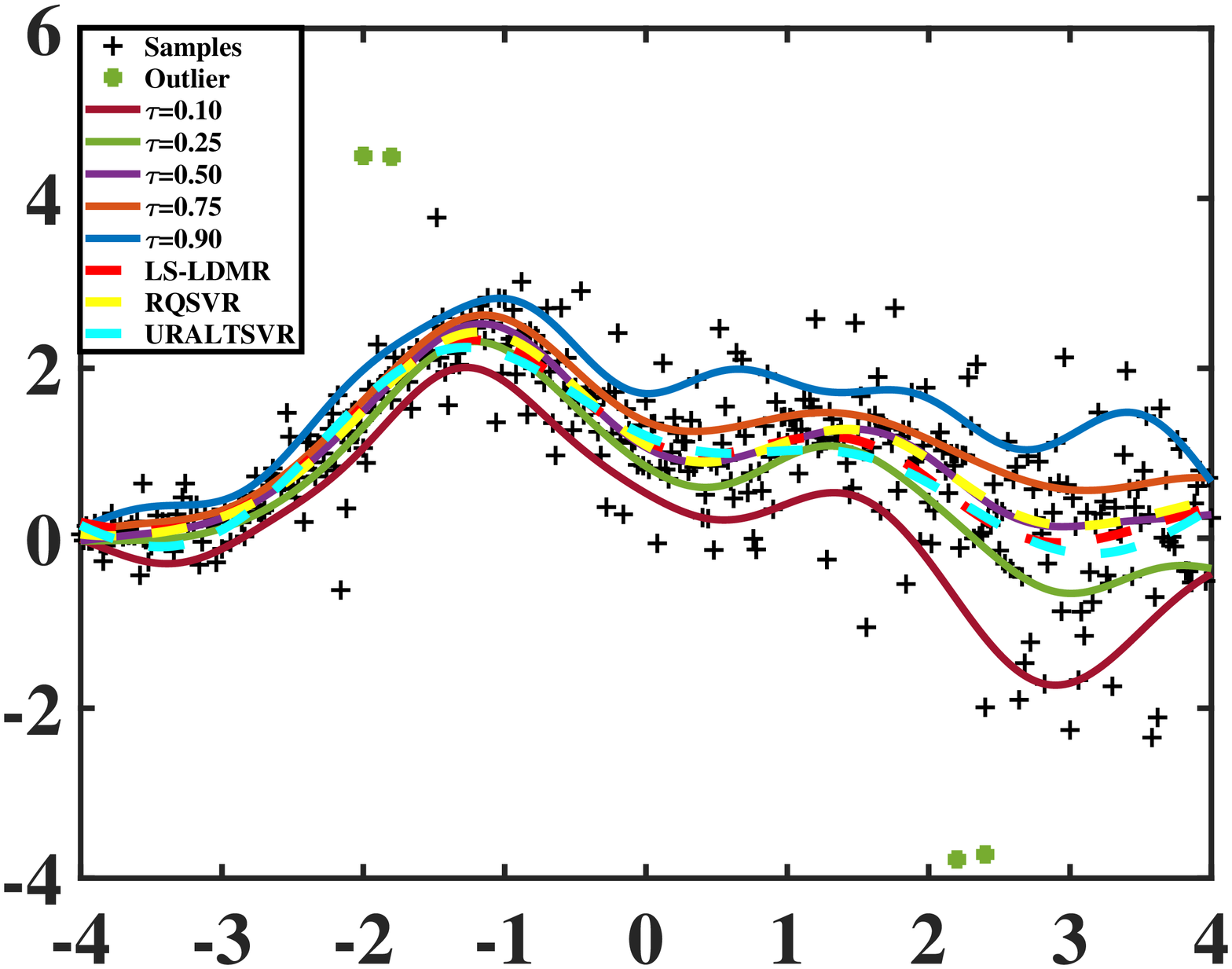}}
\caption{Regression results of SVQR, $\varepsilon$-SVQR, Online-SVQR, GPQR, TSVQR, LS-LDMR, RQSVR, and URALTSVR for Type $A_3$.}\label{fexp3}
\end{figure*}

\begin{figure*}[h!]
\centering
\subfigure[SVQR]{\includegraphics[width=0.35\textheight]{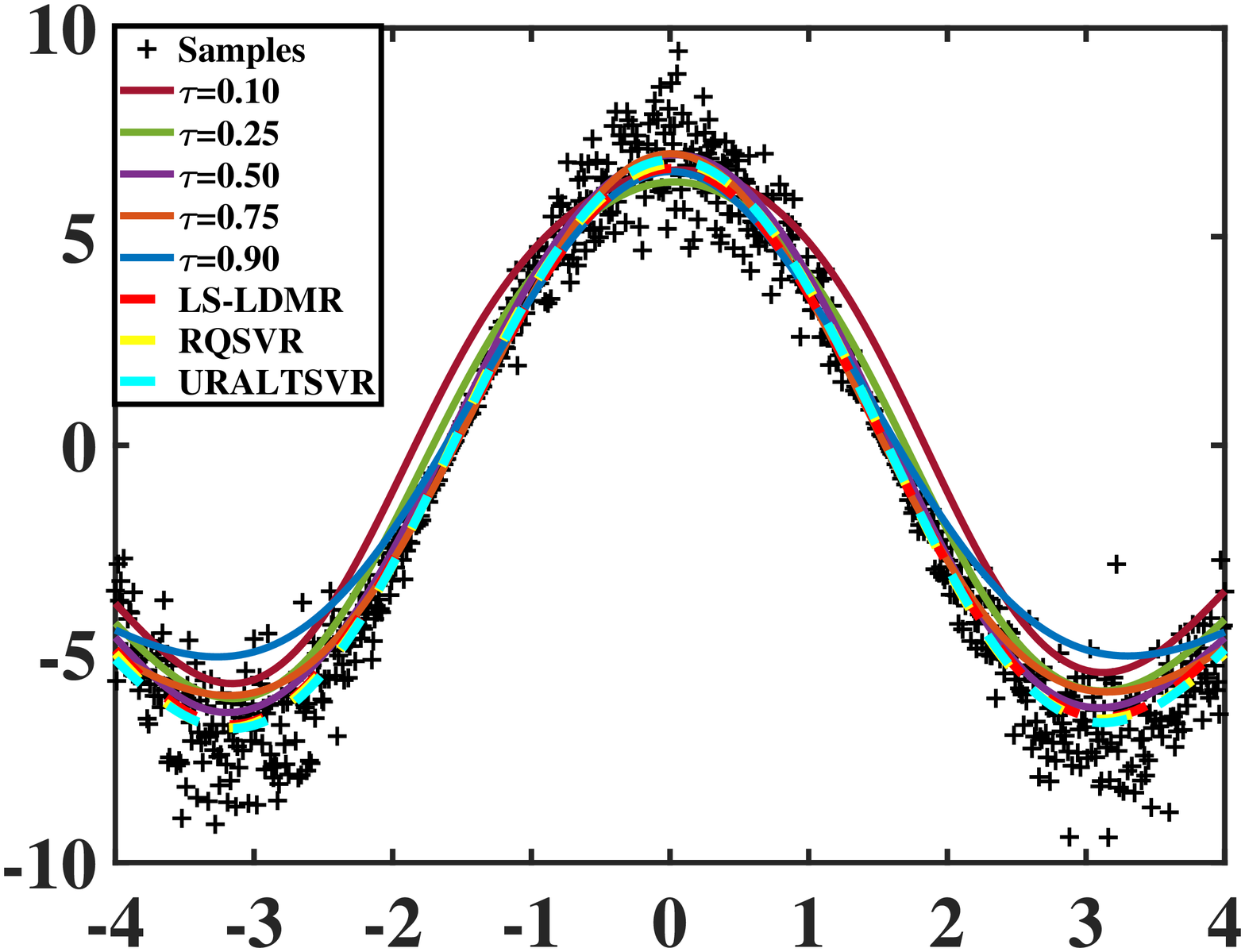}}
\subfigure[$\varepsilon$-SVQR]{\includegraphics[width=0.35\textheight]{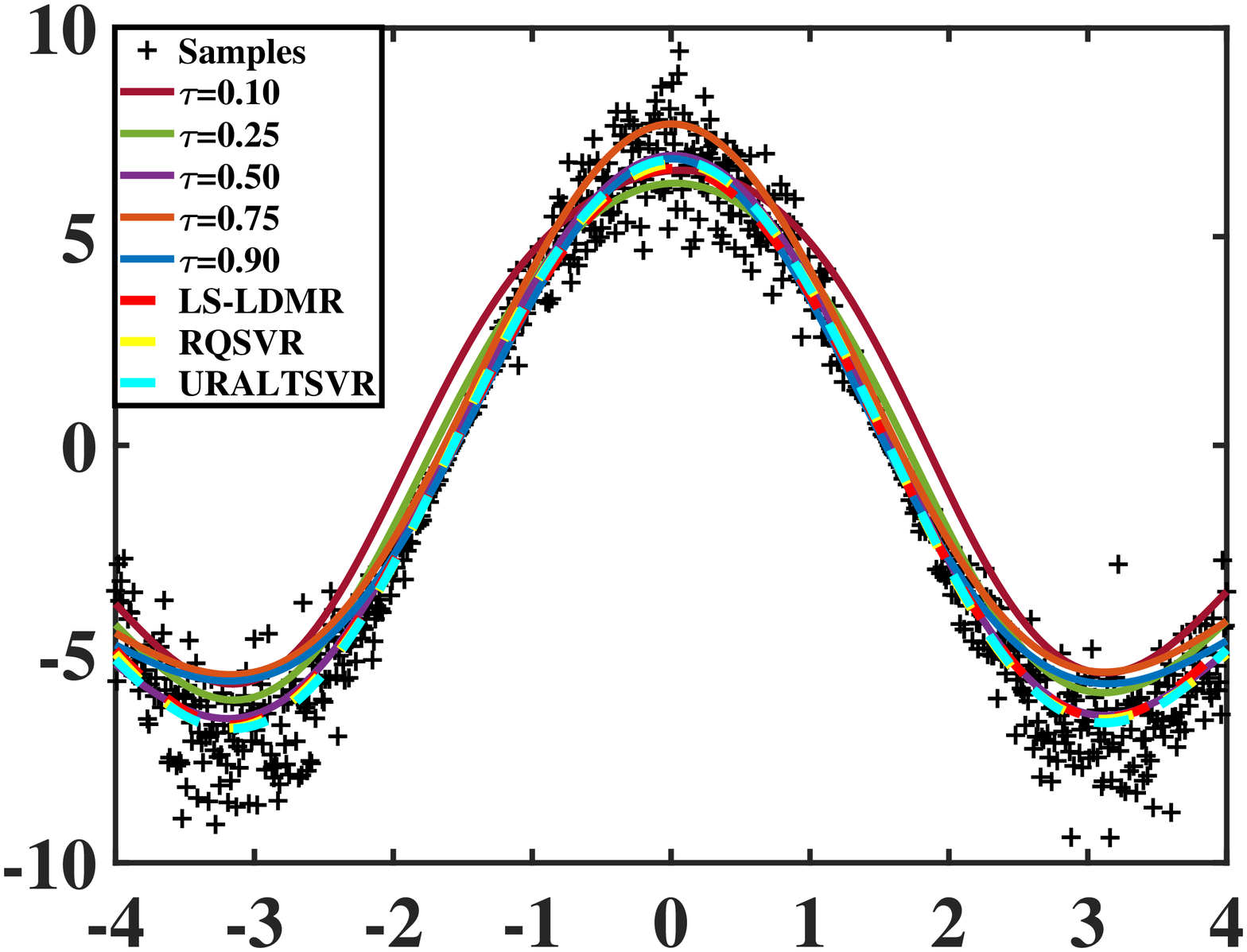}}
\subfigure[Online-SVQR]{\includegraphics[width=0.35\textheight]{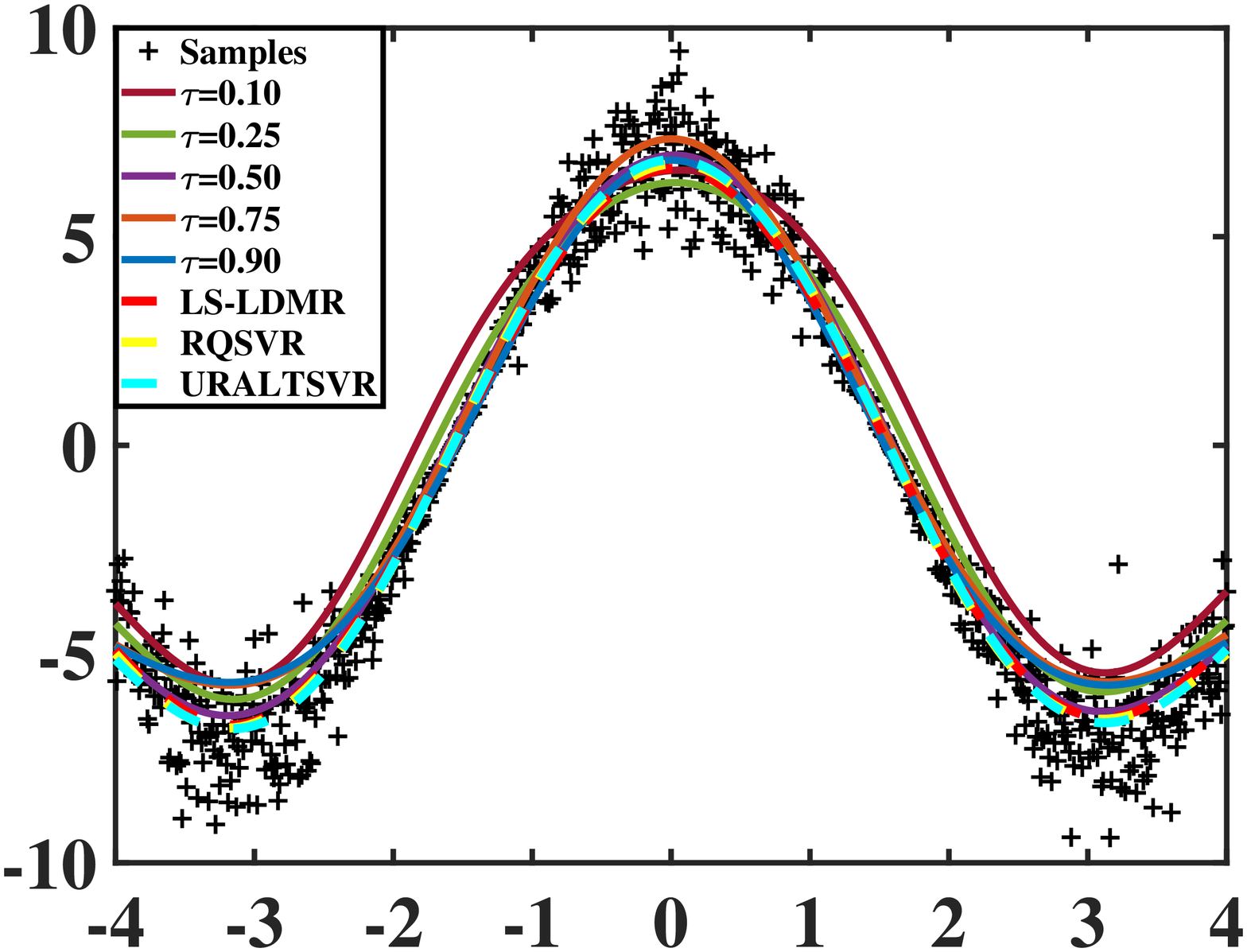}}
\subfigure[GPQR]{\includegraphics[width=0.35\textheight]{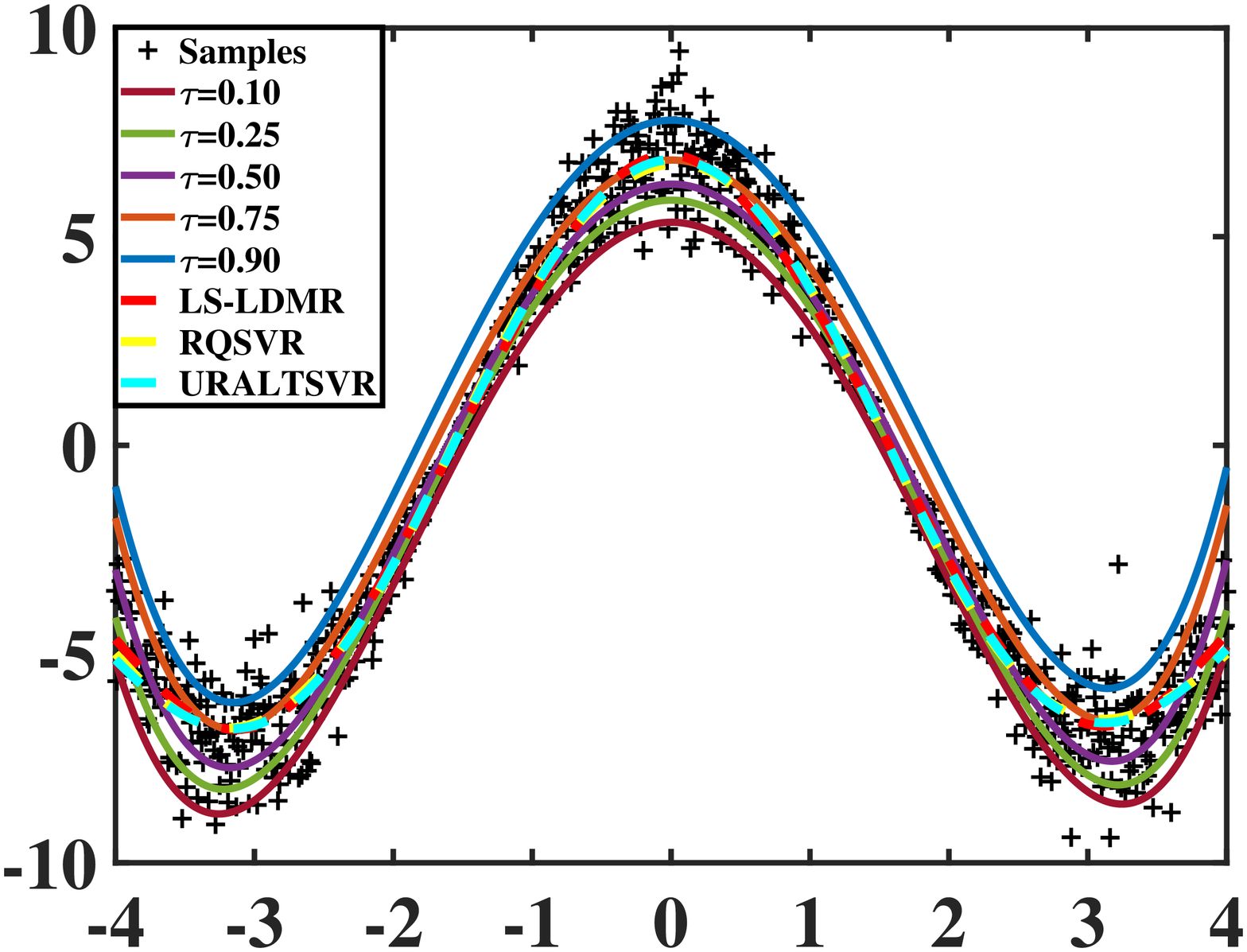}}
\subfigure[TSVQR]{\includegraphics[width=0.35\textheight]{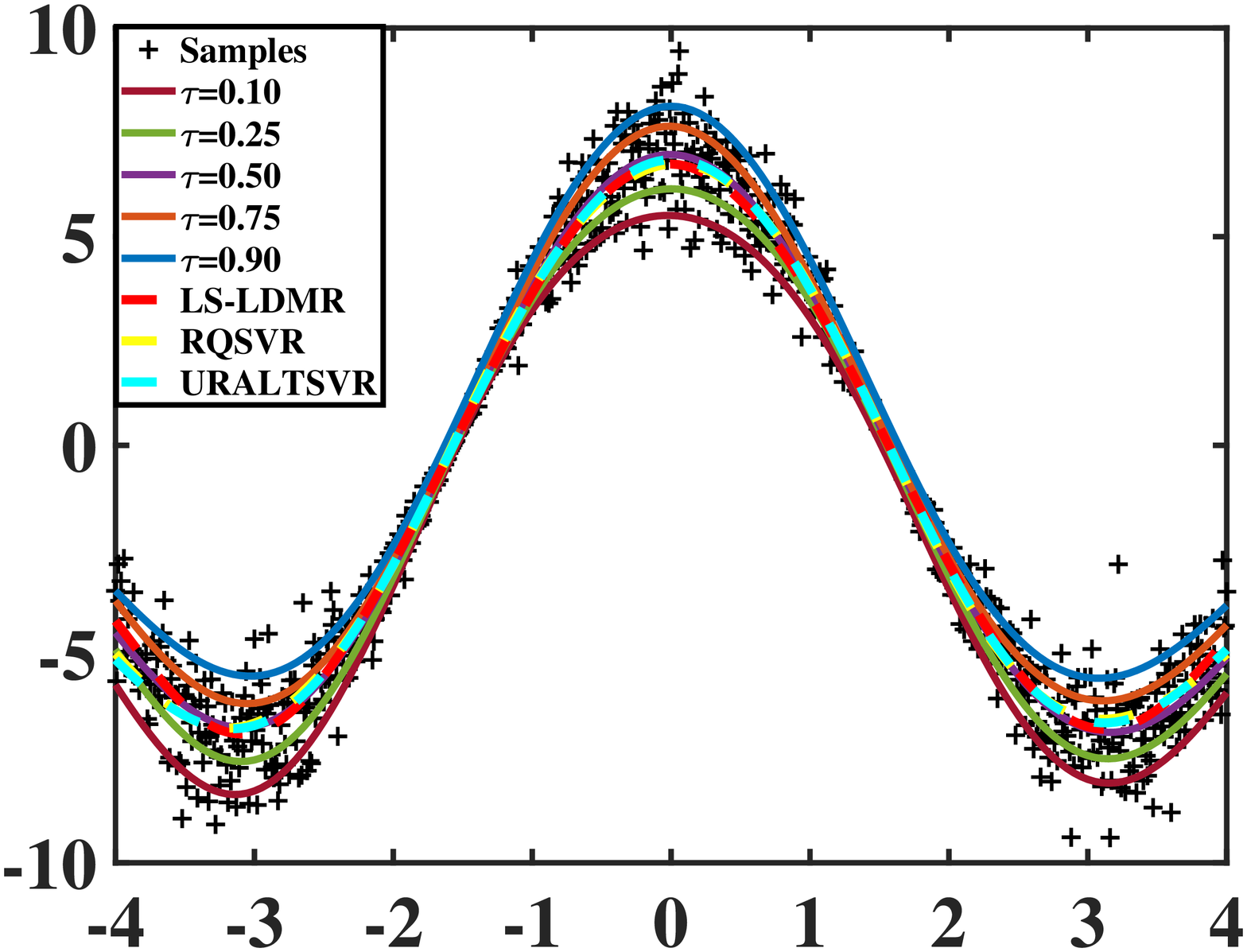}}
\caption{Regression results of SVQR, $\varepsilon$-SVQR, Online-SVQR, GPQR, TSVQR, LS-LDMR, RQSVR, and URALTSVR for Type $B_1$.}\label{fsin}
\end{figure*}

\begin{figure*}[h!]
\centering
\subfigure[SVQR]{\includegraphics[width=0.35\textheight]{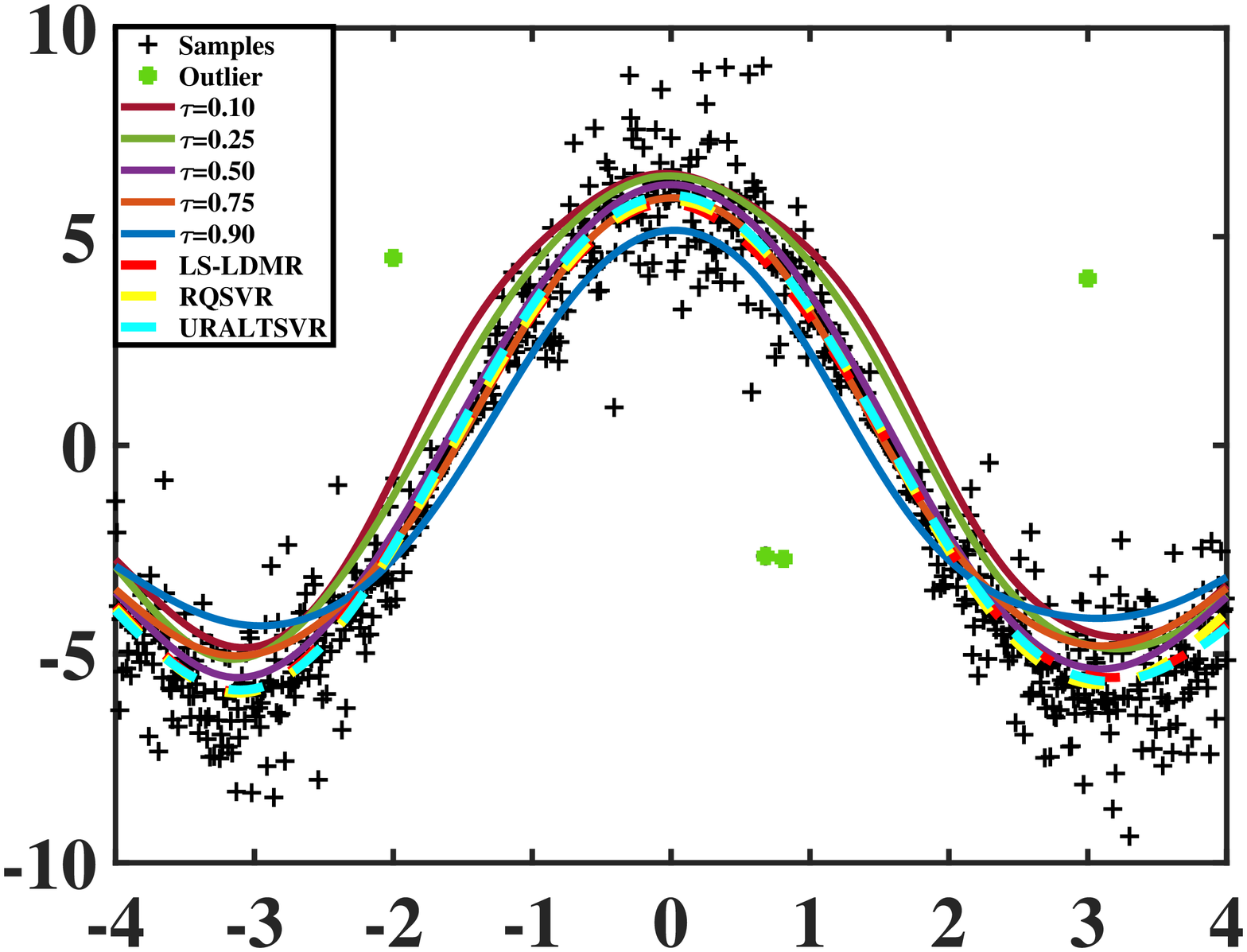}}
\subfigure[$\varepsilon$-SVQR]{\includegraphics[width=0.35\textheight]{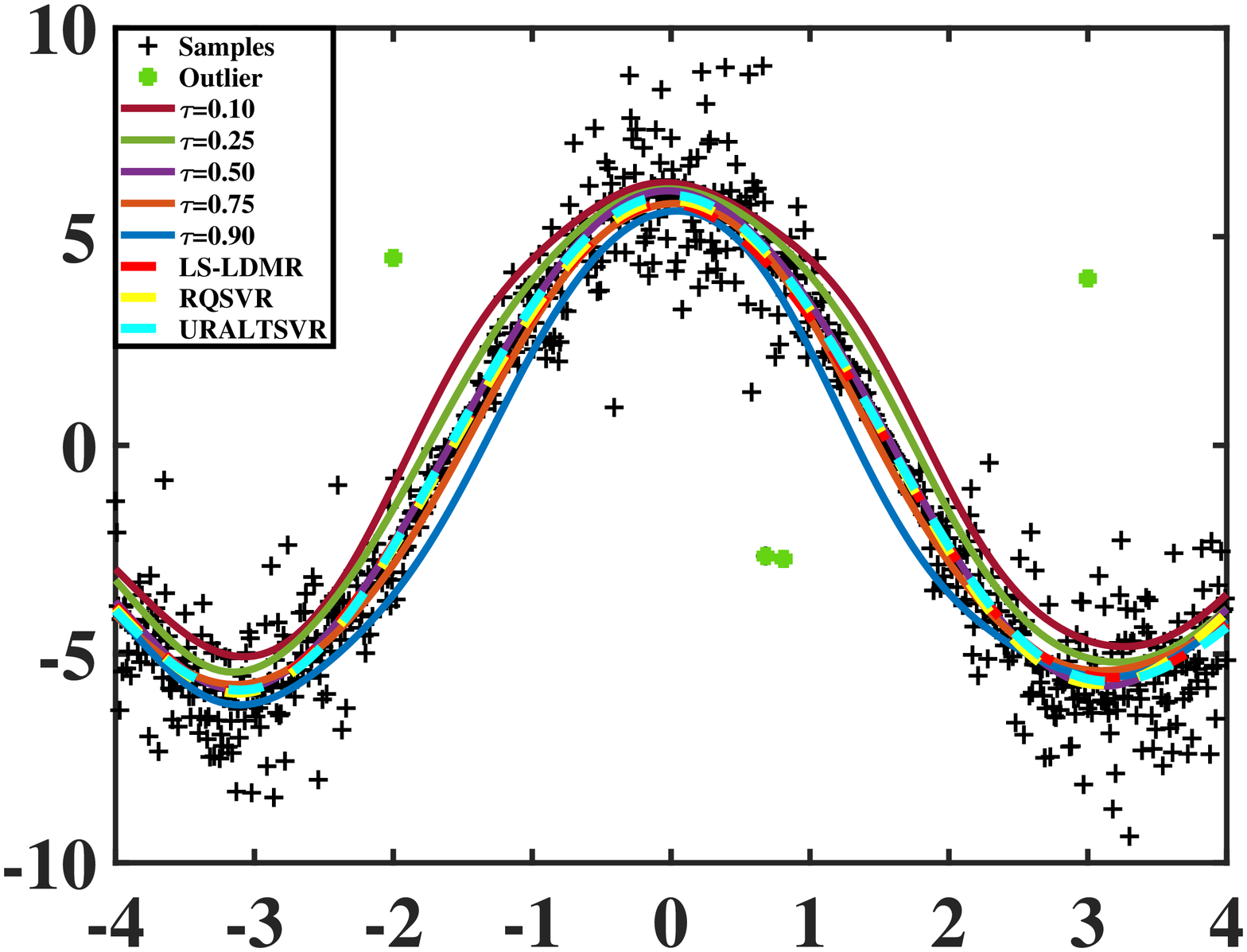}}
\subfigure[Online-SVQR]{\includegraphics[width=0.35\textheight]{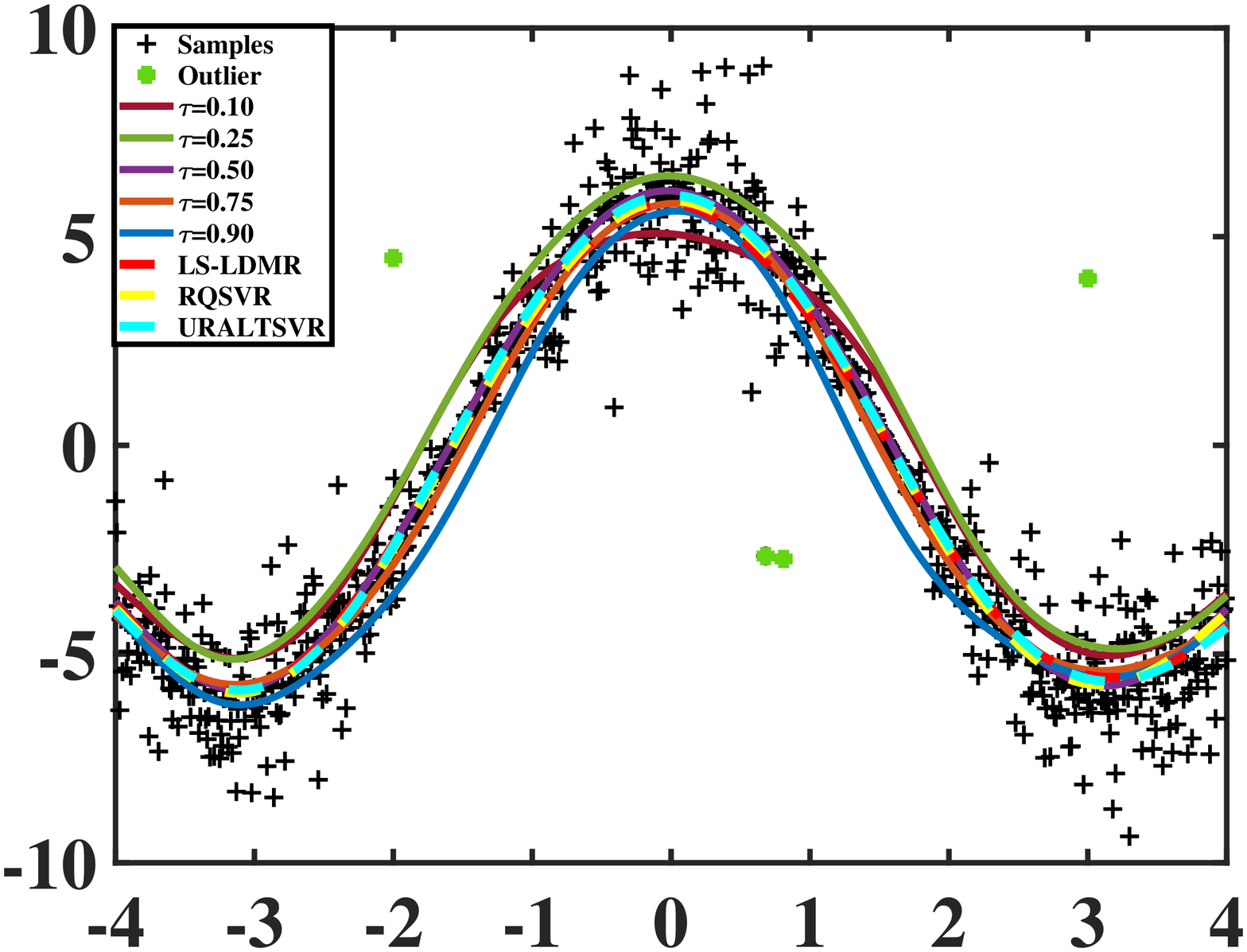}}
\subfigure[GPQR]{\includegraphics[width=0.35\textheight]{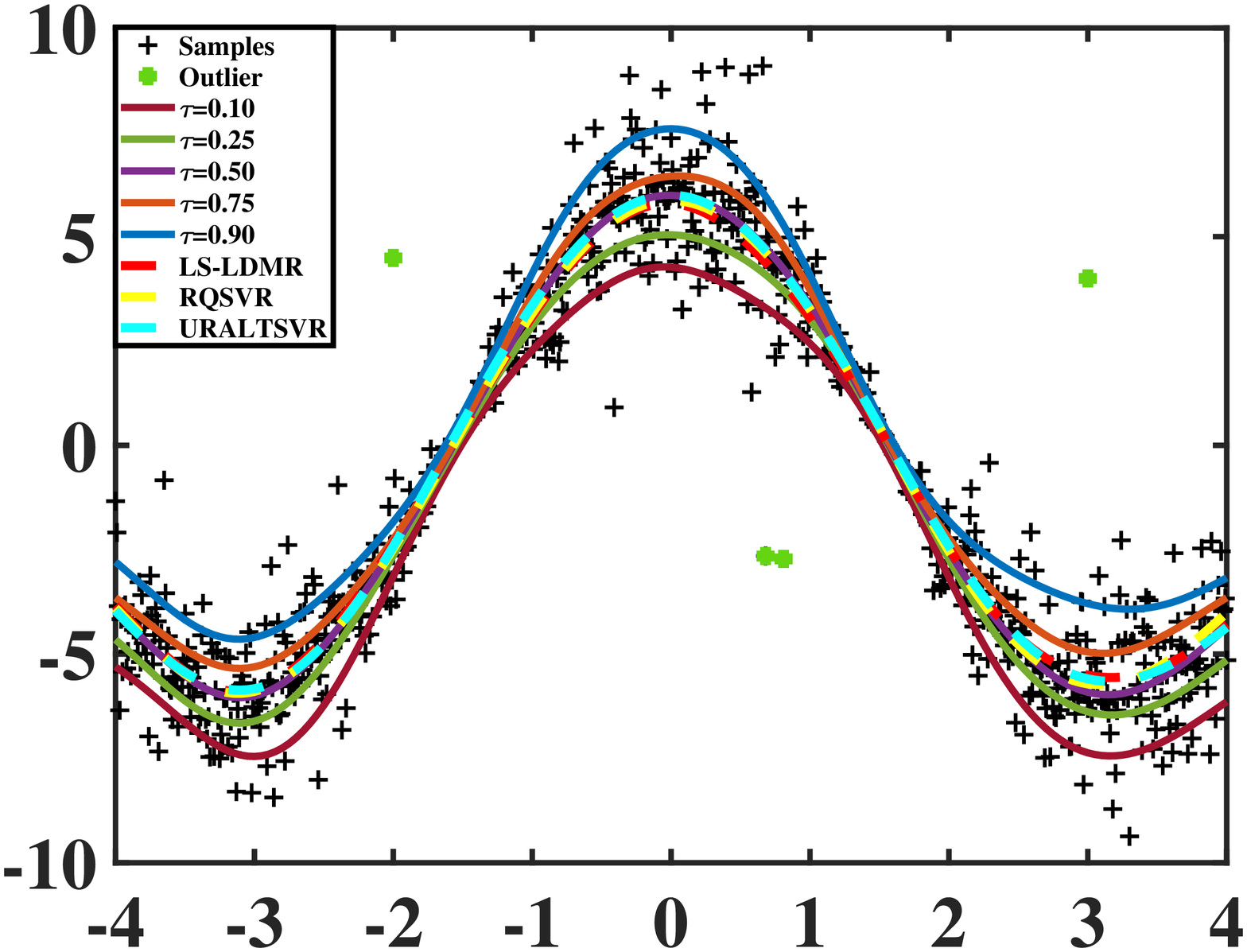}}
\subfigure[TSVQR]{\includegraphics[width=0.35\textheight]{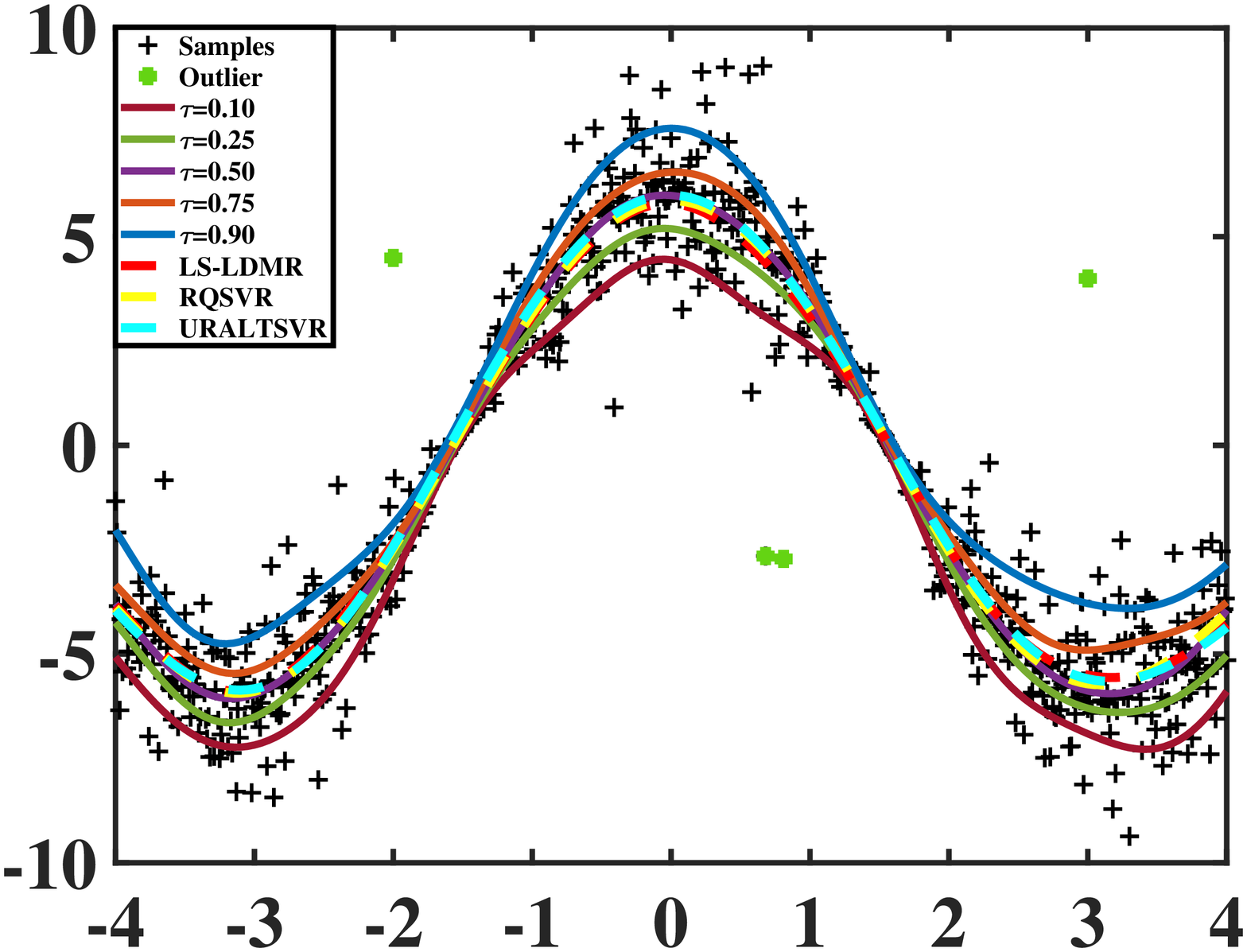}}
\caption{Regression results of SVQR, $\varepsilon$-SVQR, Online-SVQR, GPQR, TSVQR, LS-LDMR, RQSVR, and URALTSVR for Type $B_3$.}\label{fsin3}
\end{figure*}

Tables \ref{ind1-1}-\ref{ind1-6} show the regression results of SVQR, $\varepsilon$-SVQR, Online-SVQR, GPQR and TSVQR when $\tau=0.10, 0.25, 0.50, 0.75,$ and $0.90$,
 and their corresponding regression performance is depicted in Figs. \ref{fexp}-\ref{fsin3}. Table \ref{ind1-7} shows the regression results of LS-LDMR, RQSVR and URALTSVR. From these tables,
 we see that TSVQR achieves smaller Risk, RMSE, MAE, and MAPE than those of SVQR, $\epsilon$-SVQR, Online-SVQR and GPQR in most cases,
 which means that TSVQR fits the data set at every quantile location.
 From Figs. \ref{fexp}-\ref{fsin3}, we observe that there is a significant distributional heterogeneity at each data set.
 In Figs. \ref{fexp}-\ref{fsin3}, the dotted lines represent the final decision function of LS-LDMR, RQSVR, and URALTSVR, respectively.
 These dotted lines are located in the middle of the training data sets, which reflect the mean of the conditional distribution of the training data set. SVQR, $\epsilon$-SVQR, Online-SVQR, GPQR and TSVQR
 depict the distribution of the training data set at different quantile locations.
 Comparing with the regression results of SVQR, $\varepsilon$-SVQR, Online-SVQR, and GPQR in Figs. \ref{fexp}-\ref{fsin3},
 we find that TSVQR effectively explores the distribution information at each quantile level. TSVQR uses a pair of nonparallel bound functions to depict the distribution information up and down each quantile level.
 Therefore, TSVQR can effectively capture the heterogeneous and asymmetric information over the whole training data set.
 Moreover, for the CPU time in Tables \ref{ind1-1}-\ref{ind1-6}, the training speed of TSVQR and URALTSVR is much faster than that of SVQR, $\varepsilon$-SVQR, Online-SVQR, and GPQR since TSVQR and URALTSVR only solve two small QPPs in the learning process.

\begin{table}
	\footnotesize
	\centering
	\caption{Comparison results of SVQR, $\varepsilon$-SVQR, Online-SVQR, GPQR, and TSVQR for artificial data set Type $A_1$.}
	\begin{tabular}{c c c c c c}
		\hline
		Indices&SVQR&$\epsilon$-SVQR& Online-SVQR &GPQR& TSVQR \\
		\hline
		$\tau=0.10$&&&&& \\
		Risk&0.5695&0.5890&0.5732& 0.0956&$\bold{0.0325}$ \\
		RMSE&0.6808&0.7020&0.6914& 0.3847 &$\bold{0.0669}$\\
		MAE&0.6344&0.6551&0.6447& 0.2899&$\bold{0.0404}$ \\
		MAPE&0.4721&0.4780&0.4761& 0.3875&$\bold{0.0789}$\\
		CPU Time&0.3056&0.1822&0.2439 & 0.1671&$\bold{0.0220}$ \\
		$\tau=0.25$&&&&& \\
		Risk&0.3698&0.3719&0.3708 & 0.1734&$\bold{0.0352}$ \\
		RMSE&0.5334&0.5361&0.5347 & 0.3609&$\bold{0.0648}$\\
		MAE&0.4941&0.4969&0.4955& 0.2681&$\bold{0.0517}$  \\
		MAPE&0.3847&0.3858&0.3853& 0.3610&$\bold{0.0760}$   \\
		CPU Time&0.1748&0.1781&0.1765& 0.1488&$\bold{0.0271}$  \\
		$\tau=0.50$&&&&&\\
		Risk&0.0504&0.0957&0.0723& 0.2278&$\bold{0.0369}$\\
		RMSE&0.1190&0.2049&0.1587& 0.5054&$\bold{0.0937}$ \\
		MAE&0.1008&0.1914&0.1446& 0.4555&$\bold{0.0738}$ \\
		MAPE&0.1035&0.1713&0.1392 & 0.4345&$\bold{0.0786}$\\
		CPU Time&0.2045&0.1845&0.1945& 0.2292&$\bold{0.0216}$\\
		$\tau=0.75$&&&&&\\
		Risk&0.0335&0.0328&0.0321& 0.1970&$\bold{0.0302}$\\
		RMSE&$\bold{0.1041}$&0.1298&0.1154 & 0.8546&0.1202\\
		MAE&$\bold{0.0871}$&0.1084&0.0958& 0.7878&0.1017\\
		MAPE&$\bold{0.0732}$&0.0894&0.0782& 0.5374&0.0981\\
		CPU Time&0.1440&0.1726&0.1583& 0.1306&$\bold{0.0202}$\\
		$\tau=0.90$&&&&&\\
		Risk&0.1307&0.1195&0.1250& 0.1207&$\bold{0.0380}$\\
		RMSE&0.2040&$\bold{0.1947}$&0.1992 & 0.6067&0.4034\\
		MAE&0.1625&$\bold{0.1530}$&0.3644& 0.8514&0.1575 \\
		MAPE&0.1414&$\bold{0.1324}$&0.1367& 0.6151&0.2036\\
		CPU Time&0.1860&0.1845&0.1853 & 0.1346&$\bold{0.0331}$\\
		\hline
	\end{tabular}
	\label{ind1-1}
\end{table}

\begin{table*}[h!]
	\footnotesize
	\centering
	\caption{Comparison results of SVQR, $\varepsilon$-SVQR, Online-SVQR, GPQR, and TSVQR for artificial data set Type $A_2$.}
	\begin{tabular}{c c c c c c}
		\hline
		Indices&SVQR&$\epsilon$-SVQR& Online-SVQR &GPQR& TSVQR \\
		\hline
		$\tau=0.10$&&&&& \\
		Risk&0.8024&0.9554&0.9583& 0.2361&$\bold{0.2199}$ \\
		RMSE&0.9612&0.8015&0.8045& 0.4721&$\bold{0.2647}$\\
		MAE&1.0680&1.0616&1.0648& 0.3756&$\bold{0.2444}$ \\
		MAPE&0.5715&0.5703&0.5709& 0.4053&$\bold{0.2865}$\\
		CPU Time&0.0811&0.0848&0.0829 & 0.1217&$\bold{0.0205}$ \\
		$\tau=0.25$&&&&& \\
		Risk&0.6374&0.6523&0.6448 & 0.3513&$\bold{0.2622}$ \\
		RMSE&0.8973&0.9153&0.9063 &0.6262&$\bold{0.3968}$\\
		MAE&0.8499&0.8697&0.8598& 0.4801&$\bold{0.3496}$  \\
		MAPE&0.4889&0.4938&0.4913& 0.4219&$\bold{0.2949}$   \\
		CPU Time&0.0751&0.0693&0.0722& 0.1260&$\bold{0.0154}$  \\
		$\tau=0.50$&&&&&\\
		Risk&0.2403&0.3697&0.3050&0.4319&$\bold{0.2149}$\\
		RMSE&0.5317&0.7728&0.6522&0.8850&$\bold{0.4906}$ \\
		MAE&0.4805&0.7394&0.6099&0.8638&$\bold{0.4298}$ \\
		MAPE&0.3104&0.4080&0.3592& 0.5395&$\bold{0.2796}$\\
		CPU Time&0.0721&0.0548&0.0634& 0.1422&$\bold{0.0163}$\\
		$\tau=0.75$&&&&&\\
		Risk&0.0969&0.3011&0.1990& 0.3173&$\bold{0.0576}$\\
		RMSE&0.4733&0.4878&0.4805 &0.5694&$\bold{0.3985}$\\
		MAE&0.3877&0.4044&0.3961& 0.5118&$\bold{0.3305}$\\
		MAPE&$\bold{0.2195}$&0.2290&0.2242& 0.6189&0.3194\\
		CPU Time&0.0800&0.0726&0.0763& 0.1220&$\bold{0.0198}$\\
		$\tau=0.90$&&&&&\\
		Risk&0.1111&0.6079&0.3595&0.1843&$\bold{0.1055}$\\
		RMSE&$\bold{0.4383}$&0.4393&0.4388&0.9443&0.5821\\
		MAE&$\bold{0.3804}$&0.3803&0.3803& 0.8426&0.5549 \\
		MAPE&0.2245&$\bold{0.2228}$&0.2237&0.6846&0.4016\\
		CPU Time&0.0674&0.0622&0.0648& 0.1335&$\bold{0.0154}$\\
		\hline
	\end{tabular}
	\label{ind1-2}
\end{table*}

\begin{table*}[h!]
	\footnotesize
	\centering
	\caption{Comparison results of SVQR, $\varepsilon$-SVQR, Online-SVQR, GPQR, and TSVQR for artificial data set Type $A_3$.}
	\begin{tabular}{c c c c c c}
		\hline
		Indices&SVQR&$\epsilon$-SVQR& Online-SVQR &GPQR& TSVQR \\
		\hline
		$\tau=0.10$&&&&& \\
		Risk&0.1500&0.1441&0.3169& 0.0902&$\bold{0.0844}$ \\
		RMSE&0.4063&0.4104&0.5835& 1.0130 &$\bold{0.0995}$\\
		MAE&0.3407&0.3456&0.4132& 0.9015&$\bold{0.1712}$ \\
		MAPE&$\bold{0.4772}$&0.5431&0.6207& 1.9459&2.6641\\
		CPU Time&0.1721&0.1311&0.2671& 0.4606&$\bold{0.0263}$ \\
		$\tau=0.25$&&&&& \\
		Risk&0.1187&0.1168&0.2175 & 0.1346&$\bold{0.1051}$ \\
		RMSE&0.3665&0.3700&0.5724 & 0.6122&$\bold{0.3602}$\\
		MAE&$\bold{0.3082}$&0.3101&0.5097& 0.4985&0.5005  \\
		MAPE&$\bold{0.3598}$&0.3686&0.3648& 1.8338&2.4293   \\
		CPU Time&0.1813&0.1376&0.1949& 0.1896&$\bold{0.0263}$  \\
		$\tau=0.50$&&&&&\\
		Risk&0.1217&0.3411&0.2341& 0.3205&$\bold{0.0193}$\\
		RMSE&0.3102&0.7080&0.5221& $\bold{0.0677}$&0.4103 \\
		MAE&0.2542&0.6821&0.4681& $\bold{0.0410}$&0.3927 \\
		MAPE&0.3837&3.2275&1.2341 & 0.1877&$\bold{0.1672}$\\
		CPU Time&0.1308&0.1465&0.1865& 0.2447&$\bold{0.0280}$\\
		$\tau=0.75$&&&&&\\
		Risk&0.5925&0.5457&0.4691& 0.3687&$\bold{0.2682}$\\
		RMSE&0.8146&0.7543&0.8045 & 0.3334&$\bold{0.2038}$\\
		MAE&0.7900&0.7276&0.7588& $\bold{0.2746}$&0.3576\\
		MAPE&2.4777&2.4599&2.2729& 0.3498&$\bold{0.2951}$\\
		CPU Time&0.1454&0.1136&0.1713& 0.2076&$\bold{0.0280}$\\
		$\tau=0.90$&&&&&\\
		Risk&0.6778&0.6657&0.6219&$\bold{ 0.0613}$&0.1498\\
		RMSE&0.7997&0.7870&0.7989 & 0.6774&$\bold{0.2490}$\\
		MAE&0.7531&0.7396&0.7411& 0.6621&$\bold{0.1981}$ \\
		MAPE&2.8441&3.3715&1.3019& 0.6127&$\bold{0.1510}$\\
		CPU Time&0.1544&0.1647&0.1927 & 0.2779&$\bold{0.0288}$\\
		\hline
	\end{tabular}
	\label{ind1-3}
\end{table*}

\begin{table*}[h!]
	\footnotesize
	\centering
	\caption{Comparison results of SVQR, $\varepsilon$-SVQR, Online-SVQR, GPQR, and TSVQR for artificial data set Type $B_1$.}
	\begin{tabular}{c c c c c c}
		\hline
		Indices &SVQR&$\varepsilon$-SVQR&Online-SVQR &GPQR&TSVQR \\
		\hline
		$\tau=0.10$&&&&&\\
		Risk&1.8635&1.8564&1.8600&$\bold{ 0.1095}$&0.1169 \\
		RMSE&2.2745&2.2673&2.2709& 0.7385&$\bold{0.4595}$\\
		MAE&2.0706&2.0627&2.0666 & 0.9052&$\bold{0.4163}$\\
		MAPE&1.1695&1.1337&1.1507& 0.2619& $\bold{0.1465}$\\
		CPU Time&0.5707&0.5485&0.5597& 0.5360&$\bold{0.1058}$\\
		$\tau=0.25$&&&&& \\
		Risk&0.9243&0.9026&0.9135& 0.3901&$\bold{0.2748}$ \\
		RMSE&1.4754&1.4459&1.4606& 0.9114&$\bold{0.5362}$\\
		MAE&1.2645&1.2392&1.2518& 0.6984&$\bold{0.4677}$\\
		MAPE&0.8868&0.9090&0.7760 & 0.5653&$\bold{0.1467}$\\
		CPU Time&0.4838&0.5072&0.4966& 0.5487&$\bold{0.1023}$\\
		$\tau=0.50$&&&&&\\
		Risk&0.2994&0.2508&0.2715& 0.2932&$\bold{0.2356}$\\
		RMSE&0.7004&0.5860&0.6387& 0.6416&$\bold{0.5362}$\\
		MAE&0.5987&0.5015&0.5430& 0.5864&$\bold{0.4711}$\\
		MAPE&0.2081&0.1421&0.2454& 0.4292&$\bold{0.1000}$\\
		CPU Time&0.4423&0.4479&0.4451& 0.5635&$\bold{0.1141}$\\
		$\tau=0.75$&&&&&\\
		Risk&0.3458&0.2240&0.2716& 0.2514&$\bold{0.2174}$\\
		RMSE&0.7250&0.7488&0.6806& 0.6952&$\bold{0.5583}$\\
		MAE&0.6131&0.6489&0.6044& 0.4890&$\bold{0.4882}$\\
		MAPE&0.7131&0.2335&0.1618& 0.3114&$\bold{0.1112}$\\
		CPU Time&0.6067&0.5626&0.5849& 0.5001&$\bold{0.0957}$\\
		$\tau=0.90$&&&&&\\
		Risk&0.5434&0.5102&0.5265& $\bold{0.1527}$&0.1892\\
		RMSE&0.9540&0.9248&0.9390& 0.6802&$\bold{0.5999}$\\
		MAE&0.7220&0.7079&0.7143&0.6785&$\bold{0.5392}$ \\
		MAPE&0.2483&0.2826 &0.2497& 0.6192&$\bold{0.1883}$\\
		CPU Time&0.5612&0.5520&0.5566& 0.4627&$\bold{0.0878}$\\
		\hline
	\end{tabular}
	\label{ind1-4}
\end{table*}

\begin{table*}[h!]
	\footnotesize
	\centering
	\caption{Comparison results of SVQR, $\varepsilon$-SVQR, Online-SVQR, GPQR, and TSVQR for artificial data set Type $B_2$.}
	\begin{tabular}{c c c c c c}
		\hline
		Indices &SVQR&$\varepsilon$-SVQR&Online-SVQR &GPQR&TSVQR \\
		\hline
		$\tau=0.10$&&&&&\\
		Risk& 0.8239& 0.9657& 0.8948&$\bold{ 0.2348}$& 0.5496\\
		RMSE& 1.0961& 0.9510& 1.0235& 0.5847&$\bold{ 0.3121}$\\
		MAE& 0.8429& 0.8039& 0.8234 & 0.3485&$\bold{ 0.2223}$\\
		MAPE& 0.9661& 0.5475& 0.7568& 0.7149& $\bold{ 0.4299}$\\
		CPU Time& 0.4793& 0.3701& 0.4247& 0.6505&$\bold{ 0.0651}$\\
		$\tau=0.25$&&&&& \\
		Risk& 0.9851& 0.8427& 0.9139& 0.3192&$\bold{ 0.1688}$ \\
		RMSE& 0.8509& 0.8712& 0.8610& 0.6930&$\bold{ 0.3874}$\\
		MAE& 0.8991& 1.0362& 0.9676& 0.7768&$\bold{ 0.5366}$\\
		MAPE& 0.9127& 0.8616& 0.8871 & 0.5741&$\bold{ 0.2337}$\\
		CPU Time& 0.3526& 0.4066& 0.3796& 0.5200&$\bold{ 0.0687}$\\
		$\tau=0.50$&&&&&\\
		Risk& 0.3633& 0.3420& 0.3526&$\bold{ 0.2432}$&0.3184\\
		RMSE& 0.8180& 0.7866& 0.8023& 0.5587&$\bold{ 0.5172}$\\
		MAE& 0.7266& 0.6840& 0.7053& 0.4865&$\bold{ 0.2369}$\\
		MAPE& 0.2834& 0.3691& 0.3262& $\bold{0.1039}$& 0.1371\\
		CPU Time& 0.3232& 0.3053& 0.3143& 0.5921&$\bold{ 0.0970}$\\
		$\tau=0.75$&&&&&\\
		Risk& 0.4423& 0.4905& 0.4664& 0.2415&$\bold{ 0.1817}$\\
		RMSE& 0.9532& 0.8955& 0.9243& 0.9447&$\bold{ 0.6772}$\\
		MAE& 0.8279& 0.7721& 0.8000& 0.9661&$\bold{ 0.7074}$\\
		MAPE& 0.2829& 0.3863& 0.3346& 0.4356&$\bold{ 0.2396}$\\
		CPU Time& 0.3400& 0.3707& 0.3554& 0.5307&$\bold{ 0.0797}$\\
		$\tau=0.90$&&&&&\\
		Risk& 0.7038& 0.7005& 0.7021& 0.2050&$\bold{ 0.1359}$\\
		RMSE& 0.9762& 0.7047& 0.8404& 1.0640&$\bold{0.5817}$\\
		MAE& 0.7466& 0.5750& 0.6608& 2.0498&$\bold{ 0.3588}$ \\
		MAPE& 0.4351& 0.4347& 0.4349& 0.6808&$\bold{ 0.1042}$\\
		CPU Time& 0.3820& 0.3660& 0.3741& 0.5149&$\bold{ 0.0713}$\\
		\hline
	\end{tabular}
	\label{ind1-5}
\end{table*}

\begin{table*}[h!]
	\footnotesize
	\centering
	\caption{Comparison results of SVQR, $\varepsilon$-SVQR, Online-SVQR, GPQR, and TSVQR for artificial data set Type $B_3$.}
	\begin{tabular}{c c c c c c}
		\hline
		Indices &SVQR&$\varepsilon$-SVQR&Online-SVQR &GPQR&TSVQR \\
		\hline
		$\tau=0.10$&&&&&\\
		Risk&2.1839& 1.9878& 1.0231&0.5974& $\bold{0.5162}$\\
		RMSE& 2.7531& 2.5632& 2.1413&$\bold{1.1216}$&1.1991\\
		MAE& 2.4266& 2.2093& 1.8937& 1.0744&$\bold{1.0247}$\\
		MAPE& 1.4777& 1.7382& 1.1591& 0.6817& $\bold{0.3095}$\\
		CPU Time& 0.4376& 0.4182& 0.6279& 1.4200&$\bold{0.0837}$\\
		$\tau=0.25$&&&&& \\
		Risk& 1.3821& 1.1999& 1.2291& 0.7580&$\bold{ 0.6428}$ \\
		RMSE&2.1900& 1.9404& 2.1253&$\bold{ 0.7710}$& 1.2376\\
		MAE& 2.8604& 1.6517& 2.2601& $\bold{ 0.7347}$&1.1106\\
		MAPE& 1.9190& 1.9529& 1.7376 & 0.6171&$\bold{ 0.2654}$\\
		CPU Time& 0.4353& 0.4441& 0.4809& 0.5822&$\bold{ 0.0803}$\\
		$\tau=0.50$&&&&&\\
		Risk& 0.6098& 2.2063& 0.8193&$\bold{0.5116}$&0.5139\\
		RMSE&1.3695& 1.0971& 1.8129& $\bold{0.5373}$& 1.1227\\
		MAE& 1.2197& 0.3278& 0.9142& $\bold{0.5233}$&1.0278\\
		MAPE& 0.4109& 1.2063& 0.8633& 0.5076& $\bold{0.2451}$\\
		CPU Time& 0.3170& 0.3722& 0.5542& 0.5798&$\bold{0.0701}$\\
		$\tau=0.75$&&&&&\\
		Risk& 1.0659& 0.6193& 0.8127& 0.6632&$\bold{0.4638}$\\
		RMSE& 1.4655&$\bold{ 0.1287}$& 2.1234& 0.7881&1.1820\\
		MAE& 1.3193& 1.0923& 2.0051& $\bold{0.7567}$&1.0691\\
		MAPE& 0.5185& 0.4985& 0.5087& 0.5163&$\bold{0.2782}$\\
		CPU Time& 0.3931& 0.4879& 0.7409& 0.5970&$\bold{0.0734}$\\
		$\tau=0.90$&&&&&\\
		Risk& 1.0652&1.0827& 1.2711& 0.5589&$\bold{0.3324}$\\
		RMSE& 2.0803& 1.6964& 1.8160& 1.1437&$\bold{1.0690}$\\
		MAE& 1.9389&1.3087& $\bold{1.0717}$& 1.0895&1.1969 \\
		MAPE& 1.3047&1.4091&1.7522& 0.6974&$\bold{0.4182}$\\
		CPU Time& 0.3599& 0.3892& 0.7461& 0.6487&$\bold{0.0734}$\\
		\hline
	\end{tabular}
	\label{ind1-6}
\end{table*}

\begin{table*}[h!]
	\footnotesize
	\begin{center}
		\caption{Evaluation indices of artificial data sets for LS-LDMR, RQSVR and URALTSVR.}
		\begin{tabular}{@{}cccccc@{}}
			\hline
			Data set&Method&RMSE&MAE&MAPE&CPU Time  \\
			\hline
			Type $A_1$&LS-LDMR&$\bold{0.1989}$&$\bold{0.1779}$&$\bold{0.1698}$&$\bold{0.0105}$\\
			&RQSVR&0.5117&0.4694&0.4447&0.3569\\
			&URALTSVR&0.7020&0.6451&1.5015&0.0232  \\
			\hline
			Type $A_2$&LS-LDMR&$\bold{0.6205}$&$\bold{0.5676}$&$\bold{0.3470}$&$\bold{0.0023}$\\
			&RQSVR&0.9370&0.8502&0.5468&0.1075 \\
			&URALTSVR&1.1316&1.0343&0.5848&0.0212 \\
			\hline
			Type $A_3$&LS-LDMR&0.2603&0.2056&0.2811&$\bold{0.0220}$\\
			&RQSVR&$\bold{0.0875}$&$\bold{0.0539}$&$\bold{0.1826}$&0.1051 \\
			&URALTSVR&0.2096&0.1832&0.3077&0.0253 \\
			\hline
			Type $B_1$&LS-LDMR&0.6332&$\bold{0.5698}$&0.1343&$\bold{0.0249}$\\
			&RQSVR&$\bold{0.6240}$&0.5791&0.1287&0.4437\\
			&URALTSVR&0.6490&0.5956&$\bold{0.1274}$&0.1352 \\
			\hline
			Type $B_2$&LS-LDMR&0.7257&0.6527&0.1469&$\bold{0.0113}$\\
			&RQSVR&$\bold{0.4769}$&$\bold{0.4114}$&$\bold{0.0940}$&0.4419\\
			&URALTSVR&0.5532&0.5028&0.1218&0.0781  \\
			\hline
			Type $B_3$&LS-LDMR&4.3206&3.9747&0.2171&$\bold{0.0074}$\\
			&RQSVR&$\bold{0.0477}$&$\bold{0.0368}$&$\bold{0.0111}$&0.2999\\
			&URALTSVR&0.1471&0.1088&0.0546&0.0829  \\
			\hline
		\end{tabular}
       \label{ind1-7}
	\end{center}
\end{table*}

\subsection{Benchmark data sets}
In this subsection, we consider five benchmark data sets to demonstrate the efficacy of TSVQR.
The benchmark data sets include ``Engel" (\citeauthor{koenker1982robust},\citeyear{koenker1982robust}),
``Bone density" (\citeauthor{bachrach1999bone},\citeyear{bachrach1999bone}), ``US girls"(\citeauthor{cole1988fitting}, \citeyear{cole1988fitting}),
``Motorcycle" (\citeauthor{hardle1990applied},\citeyear{hardle1990applied}),
and ``Boston housing" (\citeauthor{harrison1978hedonic},\citeyear{harrison1978hedonic}).
The ``Engel" data set illustrates the relationship between food expenditure
and household income increases.
The ``Bone density" data set presents the relationship between age and bone mineral density.
 The ``US girls" data set demonstrates the relationship between age and weight for US girls.
The ``Boston housing" data set and ``Motorcycle" data set are very popular in the UC Irvine (UCI) Machine Learning Repository.
These five real-world data sets are very popular in QR studies. The specifications of these standardized real-world data sets are listed in Table \ref{dataset}. We adopt a meta-analysis to quantify the degree of heterogeneity of these five real-world data sets. Typically, in meta-analysis, heterogeneity is assessed with the $I^2$ index (\citeauthor{higgins2002quantifying},
\citeyear{higgins2002quantifying}). All the heterogeneity tests of these real-world data sets are implemented in Stata 15.1. $I^2$ values of $25\%$, $50\%$, and $75\%$ are interpreted as representing low, moderate, and high levels of heterogeneity, respectively. ``Engel", ``Bone density", ``US girls", ``Motorcycle" and ``Boston housing" $I^2$ values are $0\%$, $0\%$, $77.4\%$, $98.7\%$, and $100\%$, respectively. These values indicate that ``US girls", ``Motorcycle" and ``Boston housing" reflect great heterogeneity. The $I^2$ values of `Engel" and `Bone density" are $0$, indicating that there is no heterogeneity.

\begin{figure*}[h!]
\centering
\subfigure[SVQR]{\includegraphics[width=0.35\textheight]{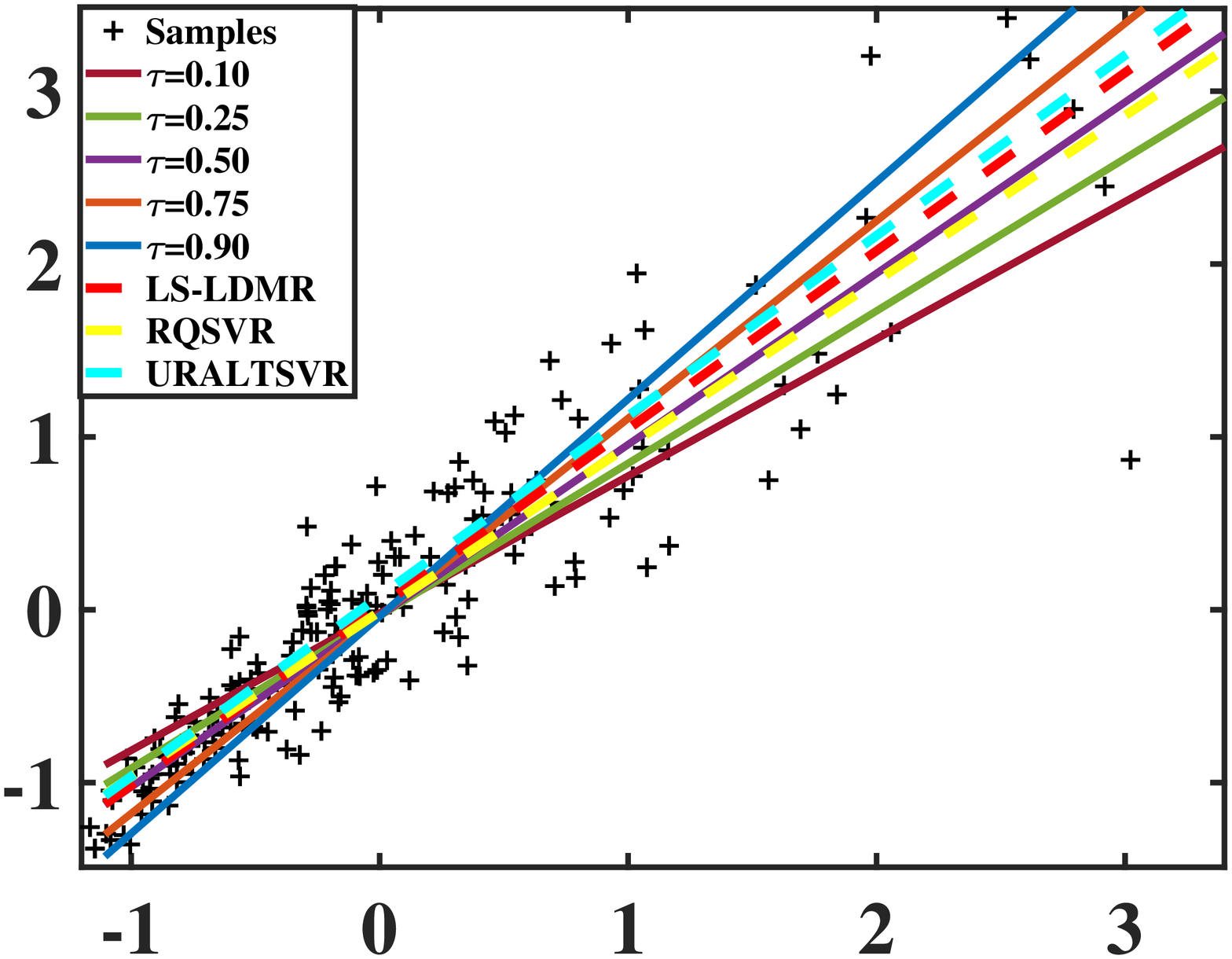}}
\subfigure[$\varepsilon$-SVQR]{\includegraphics[width=0.35\textheight]{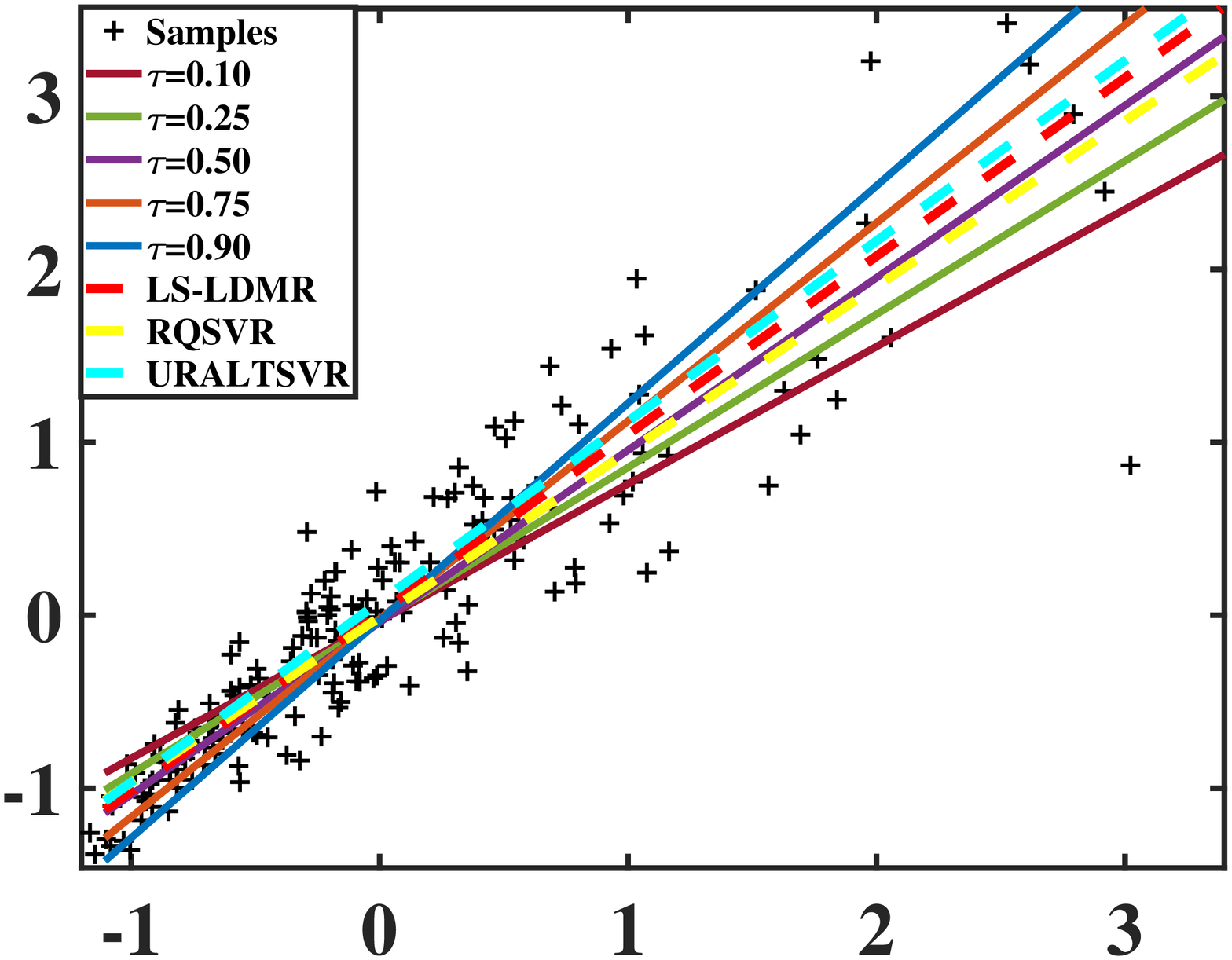}}
\subfigure[Online-SVQR]{\includegraphics[width=0.35\textheight]{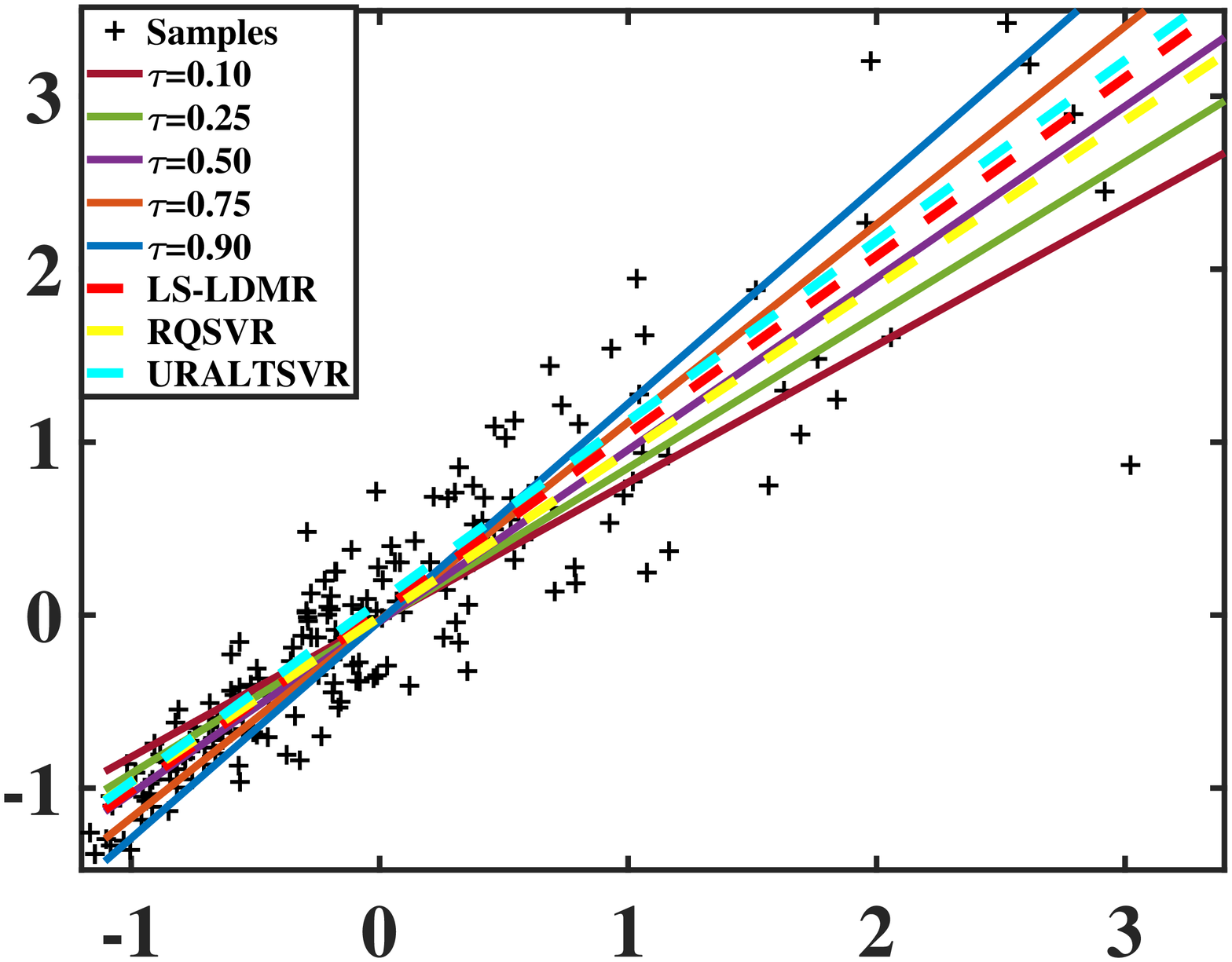}}
\subfigure[GPQR]{\includegraphics[width=0.35\textheight]{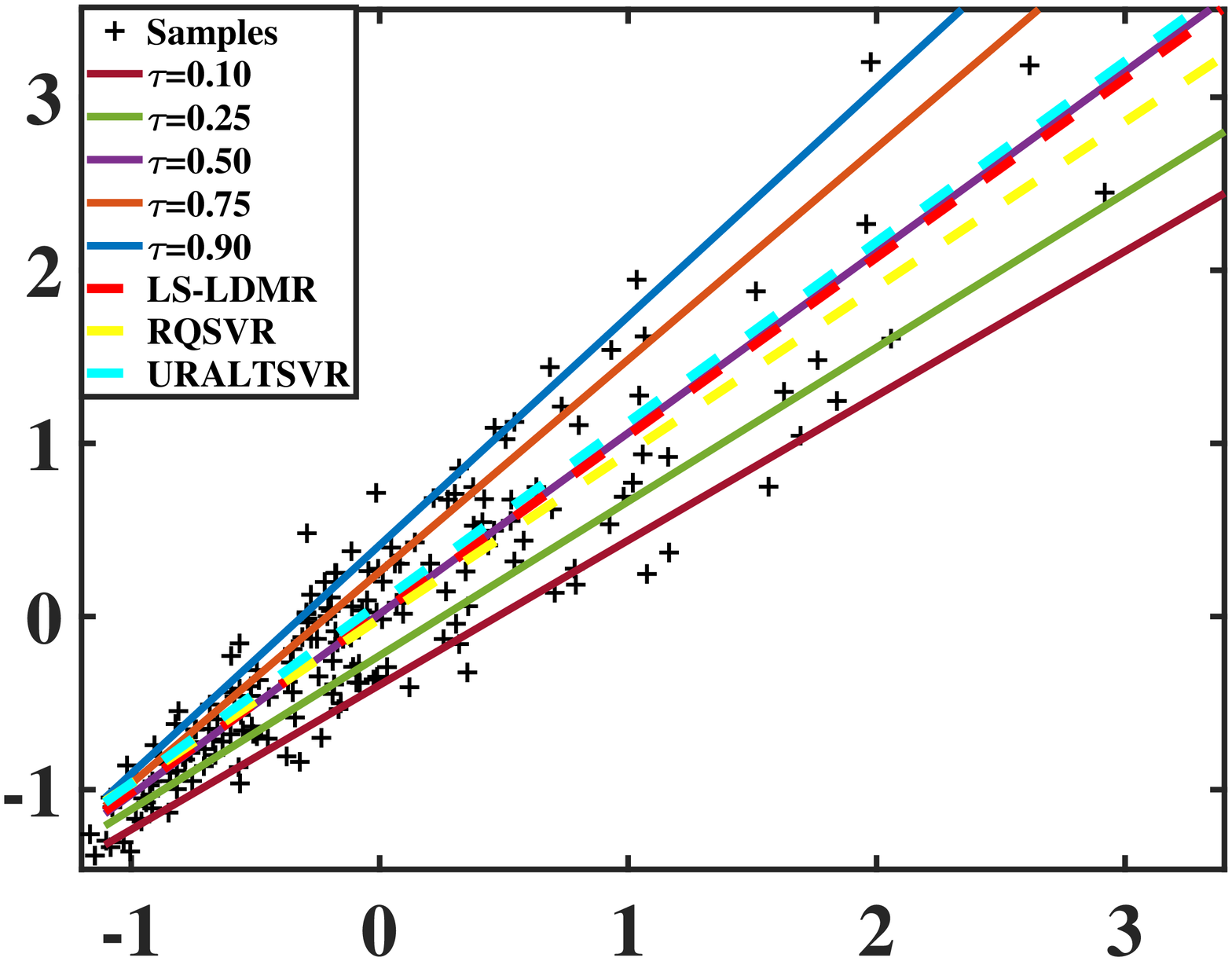}}
\subfigure[TSVQR]{\includegraphics[width=0.35\textheight]{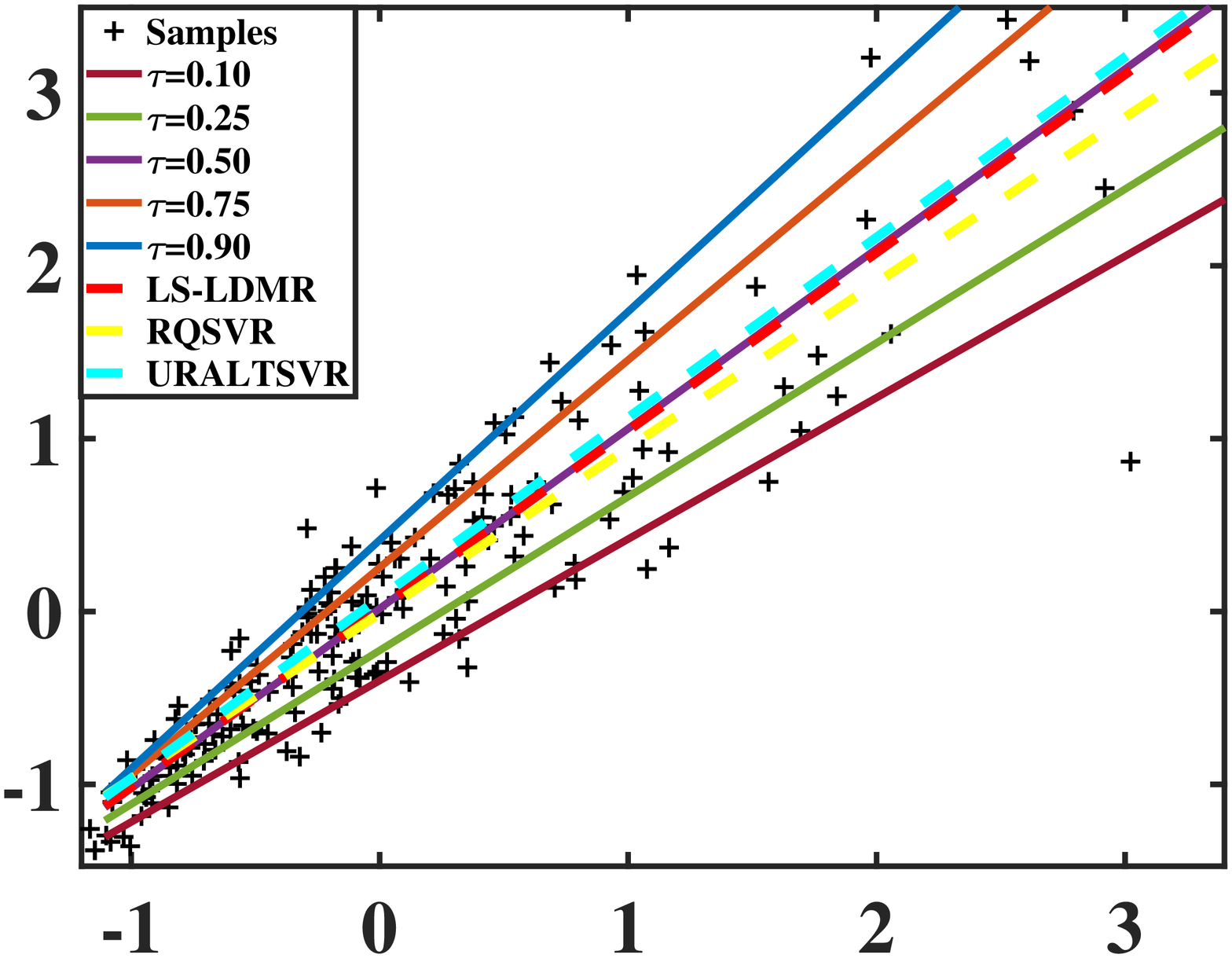}}
\caption{Regression results of SVQR, $\varepsilon$-SVQR, Online-SVQR, GPQR, TSVQR, LS-LDMR, RQSVR, and URALTSVR for Engel data set.}\label{fengel}
\end{figure*}

\begin{figure*}[h!]
\centering
\subfigure[SVQR]{\includegraphics[width=0.35\textheight]{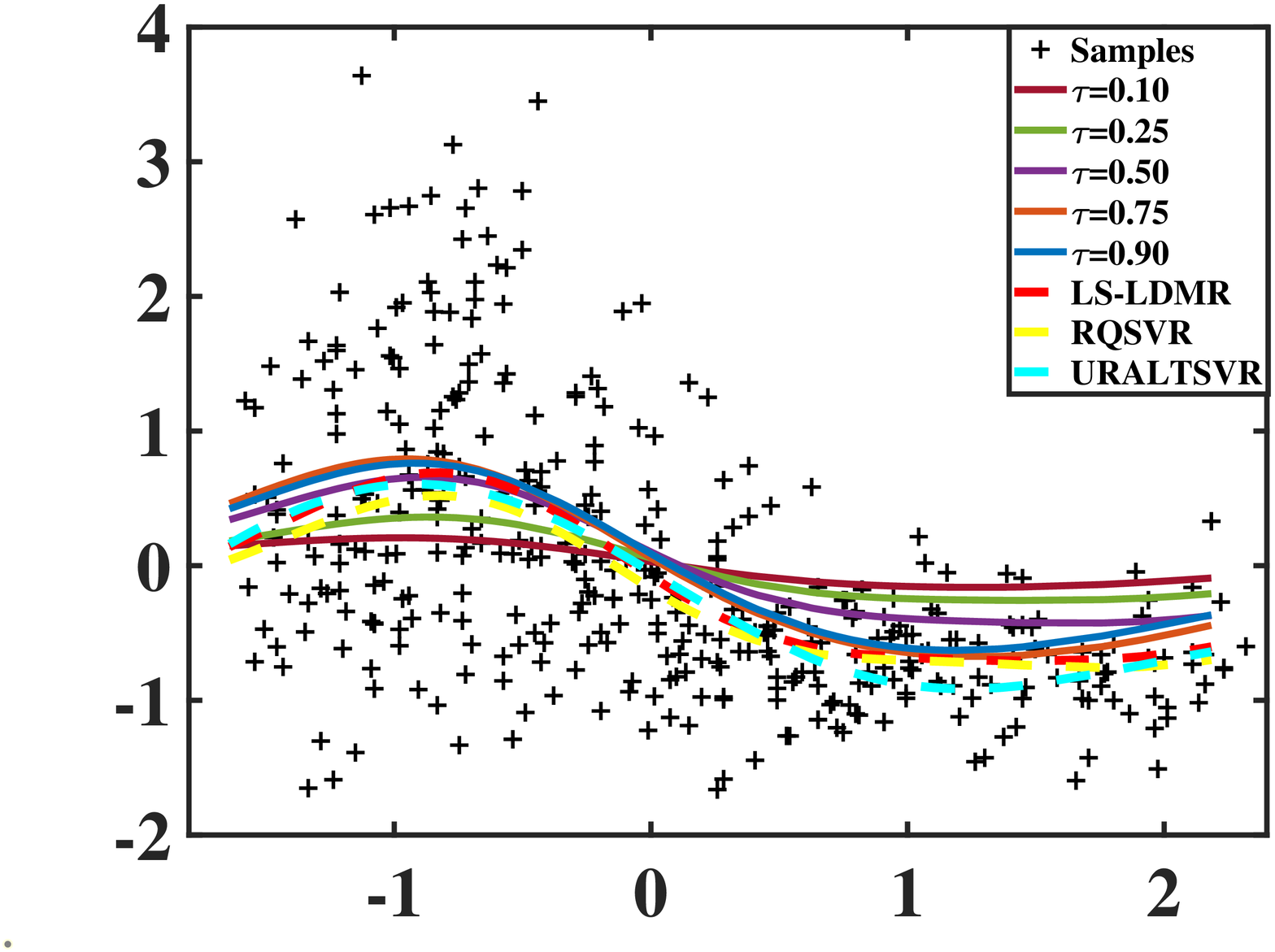}}
\subfigure[$\varepsilon$-SVQR]{\includegraphics[width=0.35\textheight]{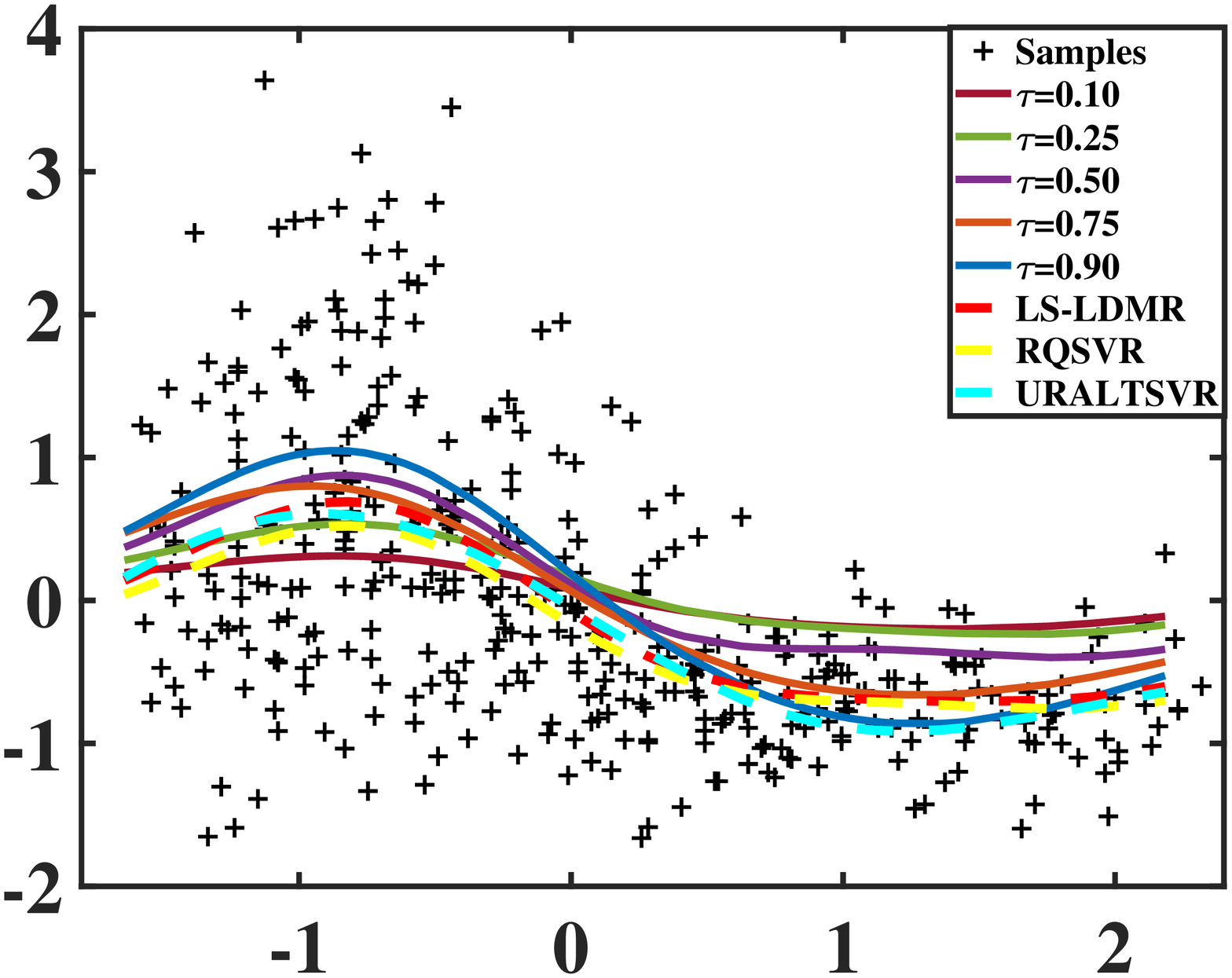}}
\subfigure[Online-SVQR]{\includegraphics[width=0.35\textheight]{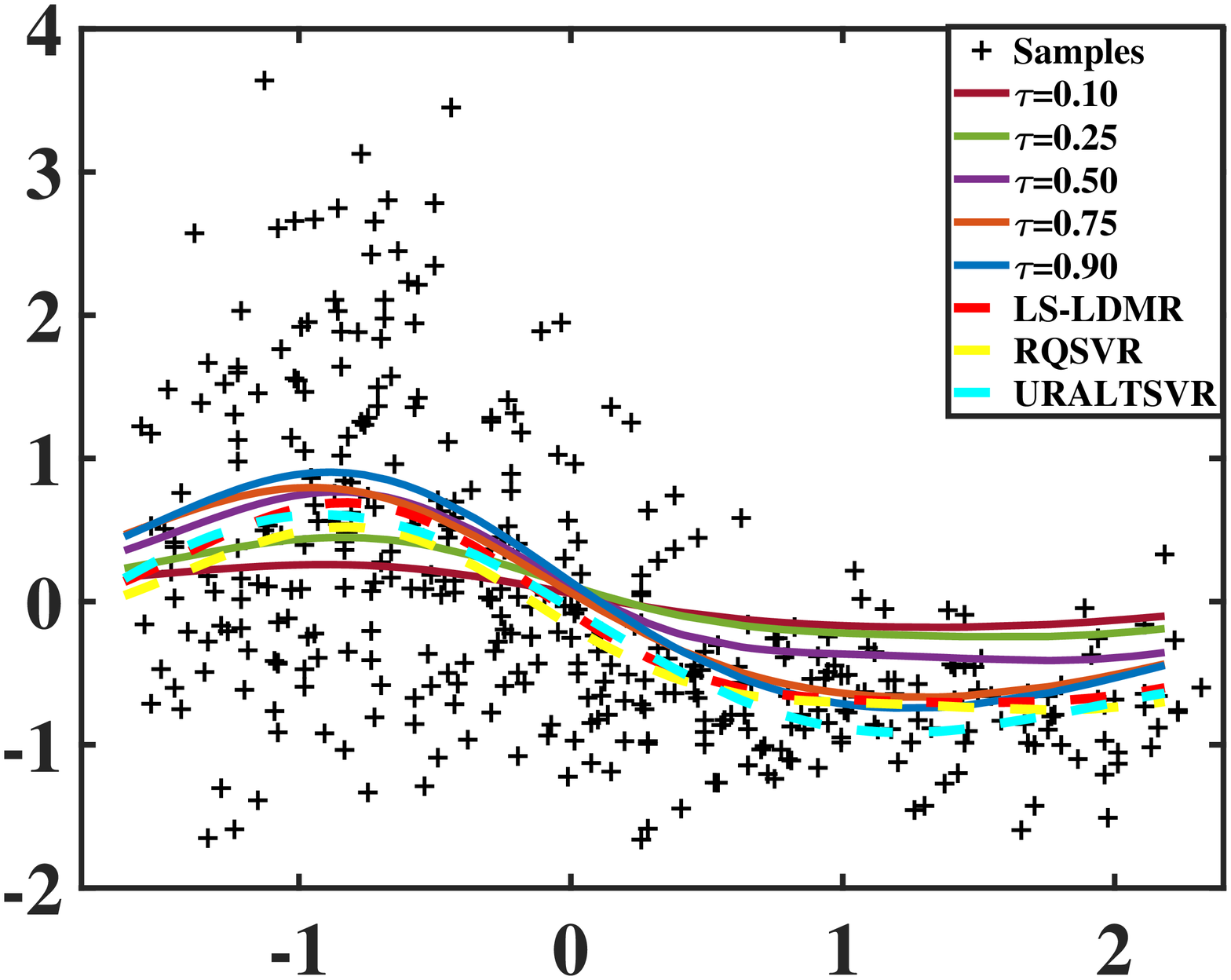}}
\subfigure[GPQR]{\includegraphics[width=0.35\textheight]{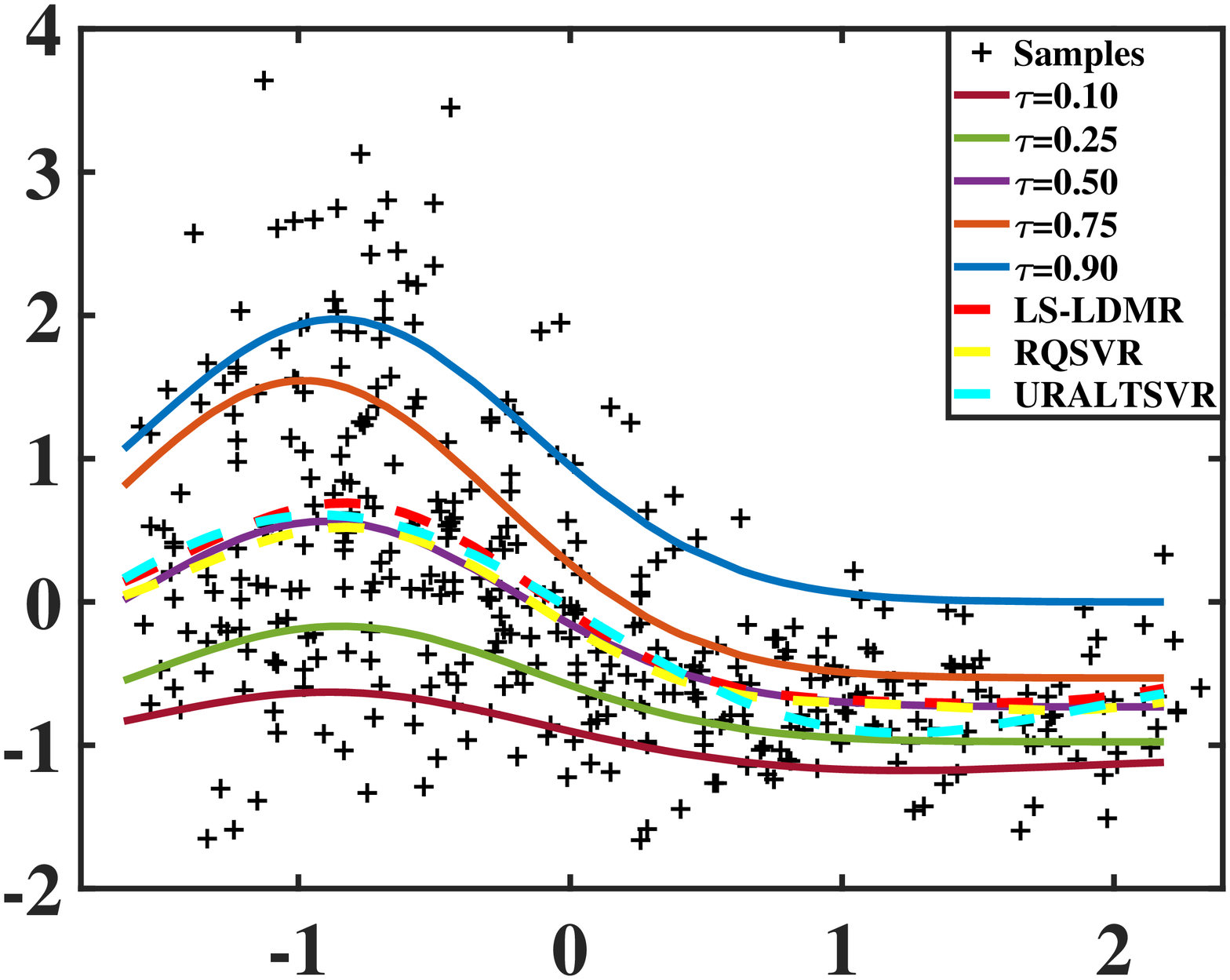}}
\subfigure[TSVQR]{\includegraphics[width=0.35\textheight]{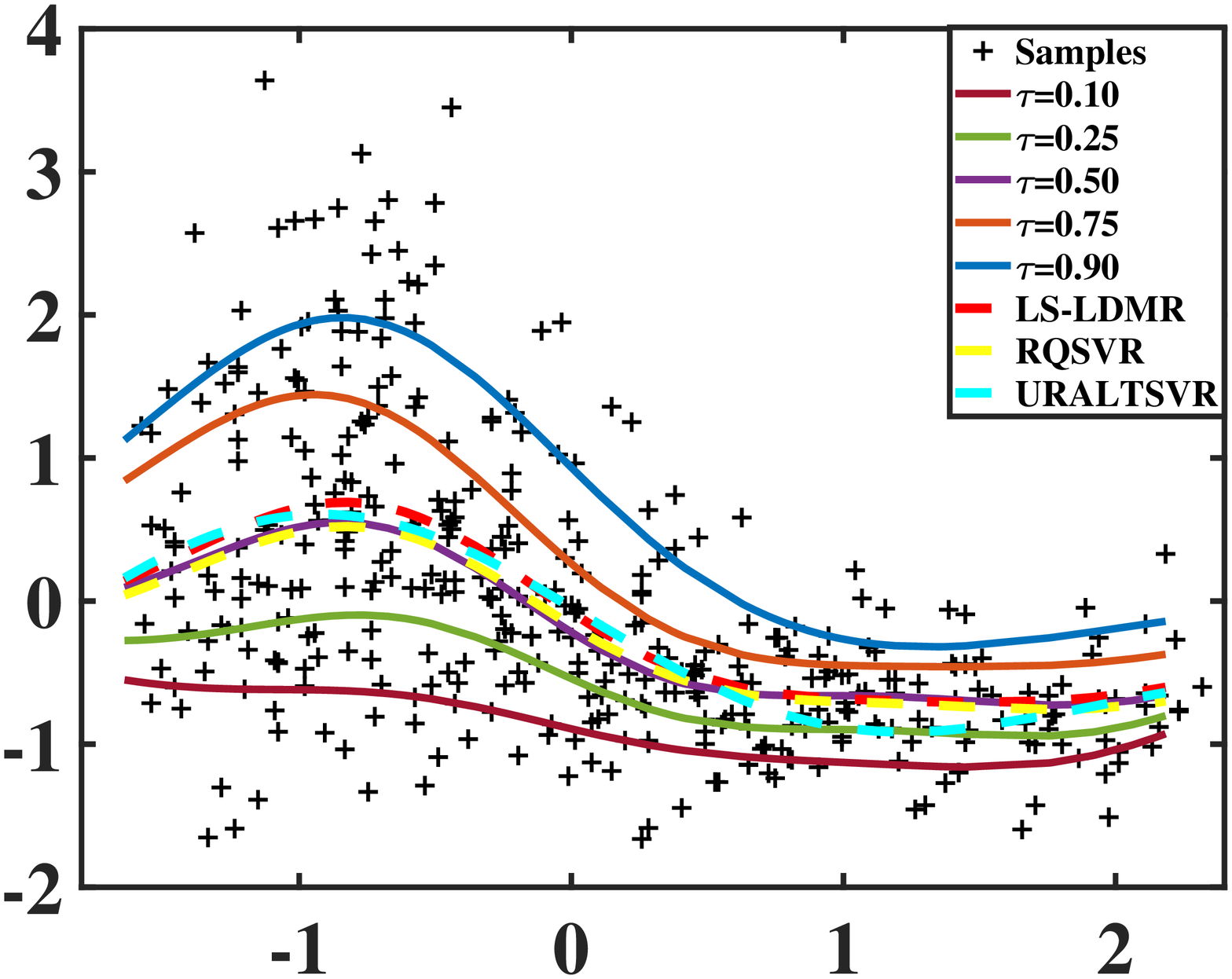}}
\caption{Regression results of SVQR, $\varepsilon$-SVQR, Online-SVQR, GPQR, TSVQR, LS-LDMR, RQSVR, and URALTSVR for Bone density data set.}\label{fbonedata}
\end{figure*}

\begin{figure*}[h!]
\centering
\subfigure[SVQR]{\includegraphics[width=0.35\textheight]{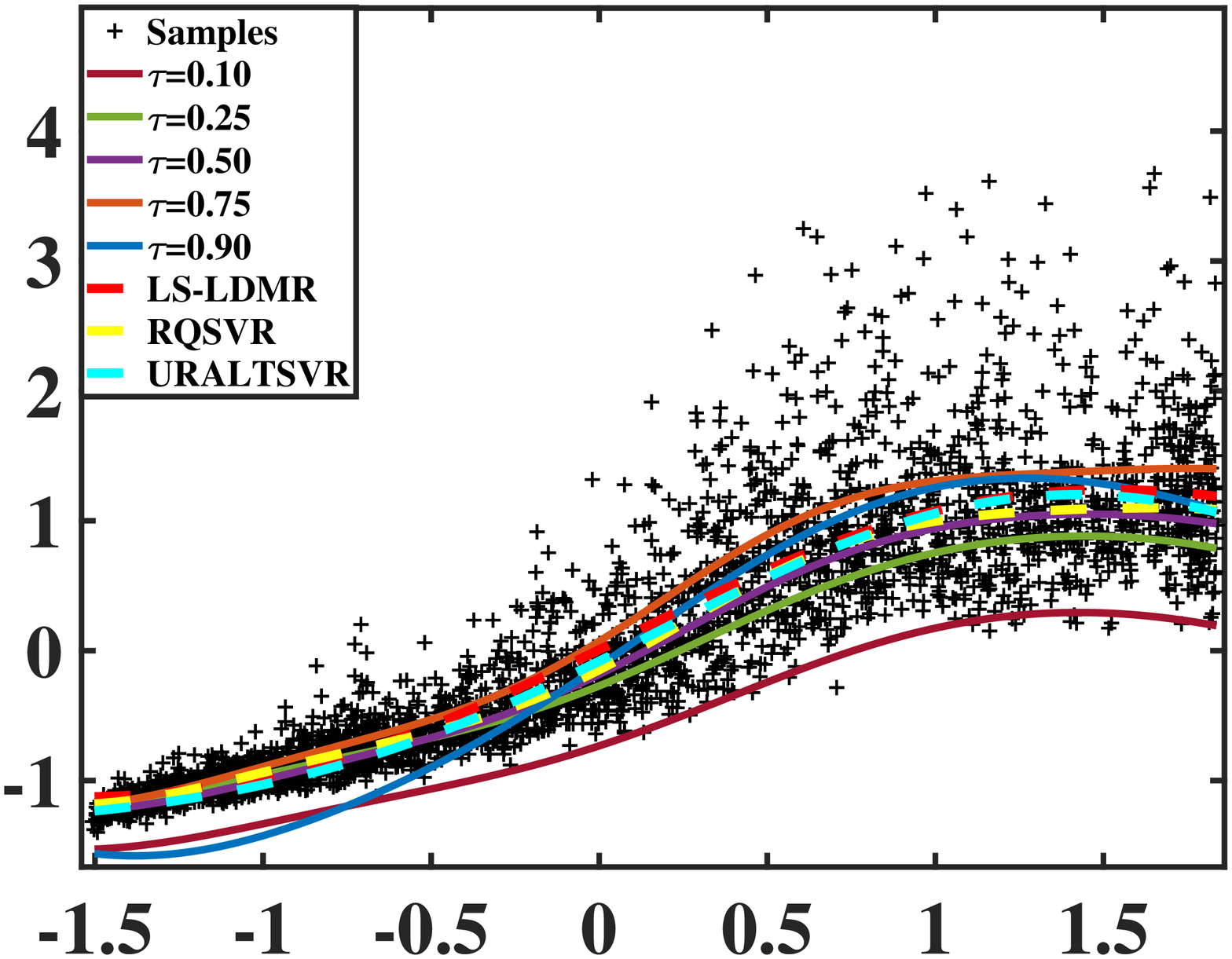}}
\subfigure[$\varepsilon$-SVQR]{\includegraphics[width=0.35\textheight]{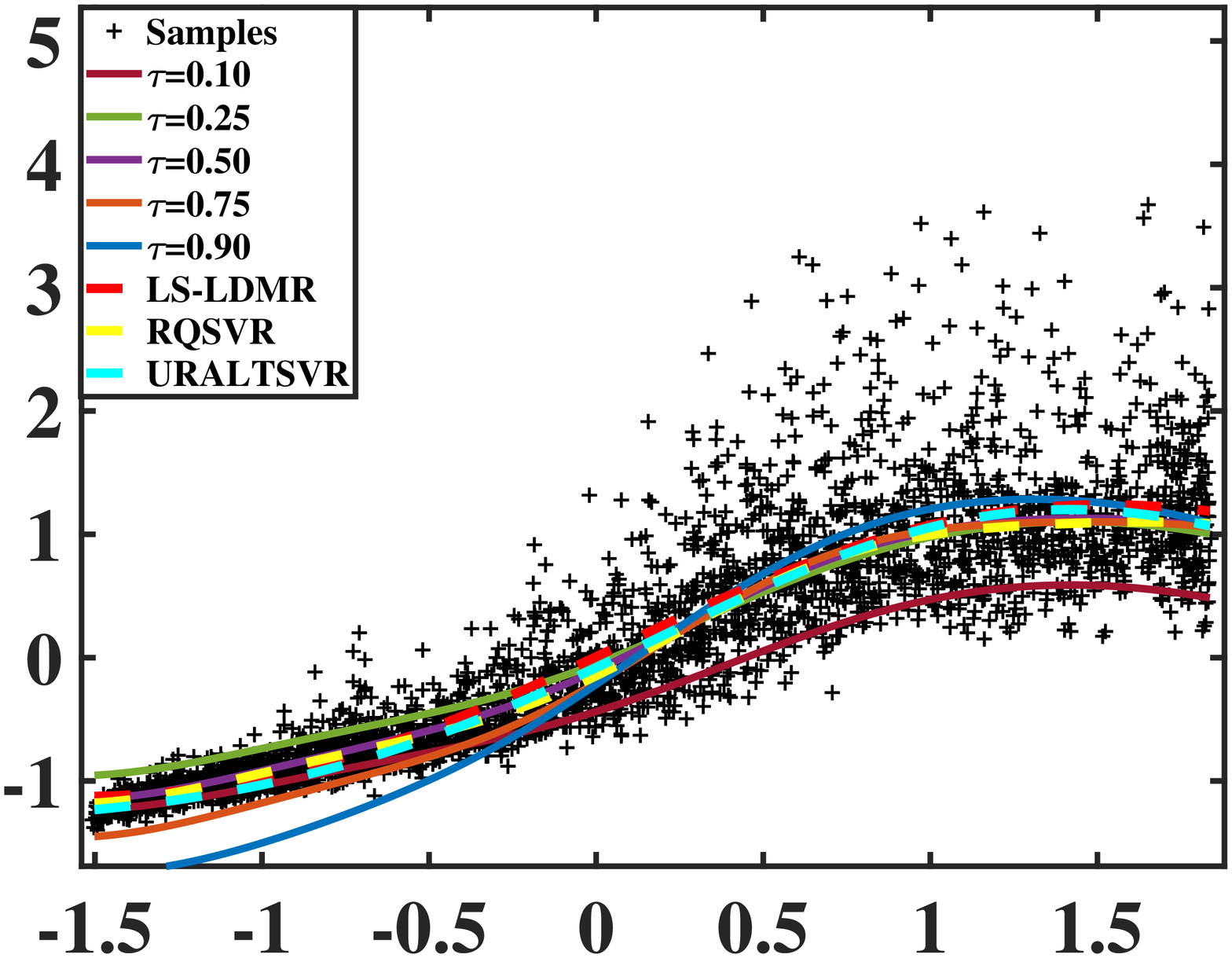}}
\subfigure[Online-SVQR]{\includegraphics[width=0.35\textheight]{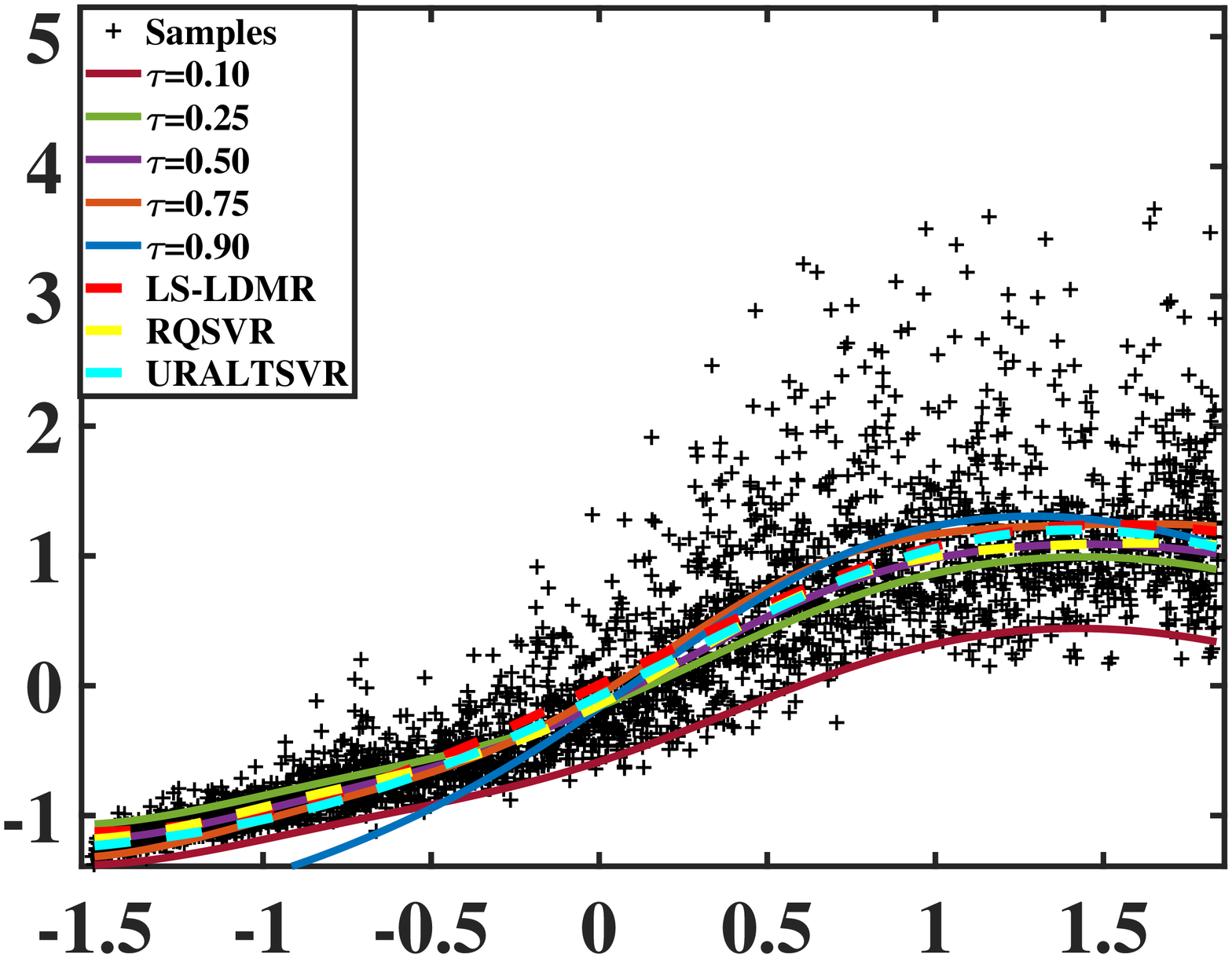}}
\subfigure[GPQR]{\includegraphics[width=0.35\textheight]{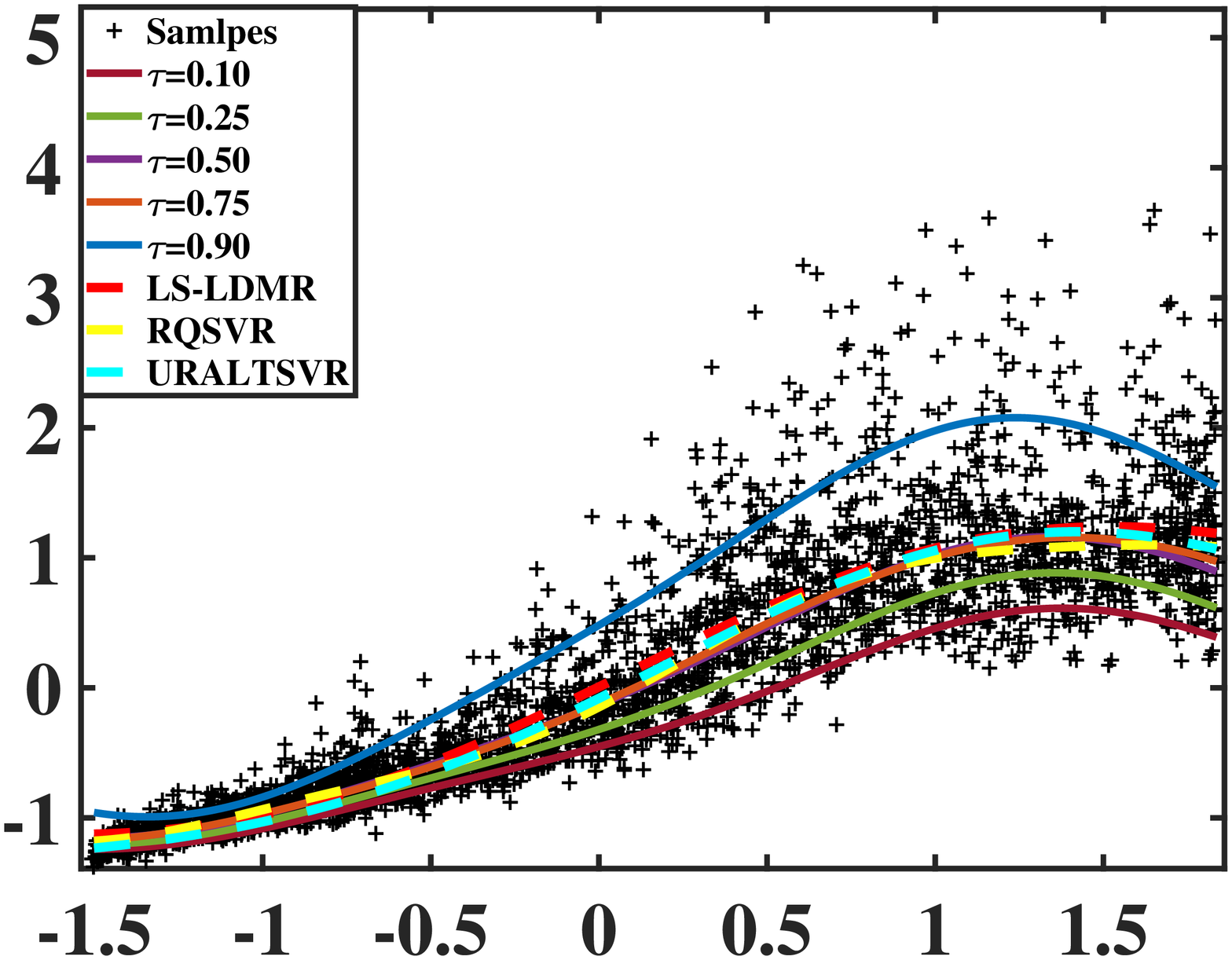}}
\subfigure[TSVQR]{\includegraphics[width=0.35\textheight]{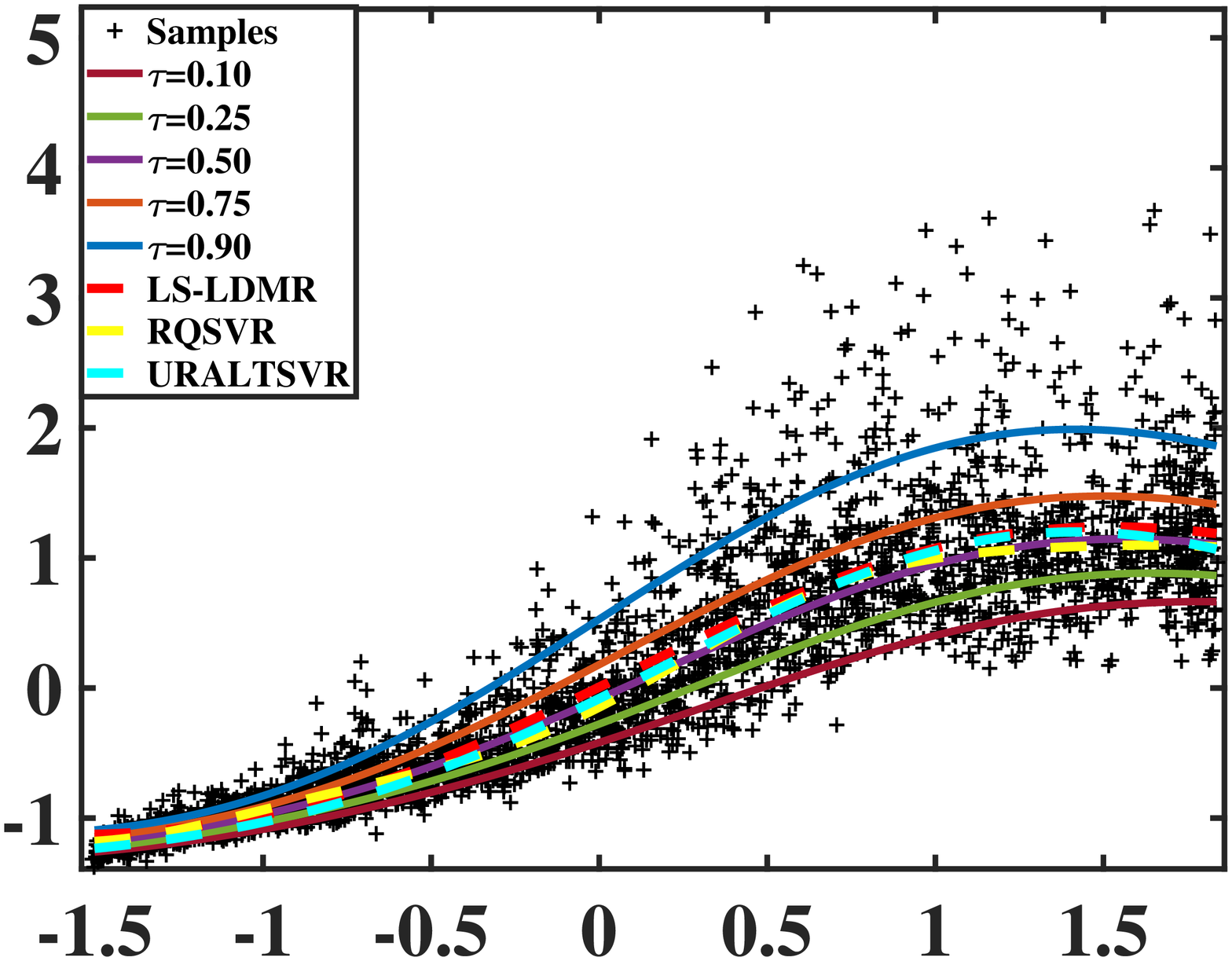}}
\caption{Regression results of SVQR, $\varepsilon$-SVQR, Online-SVQR, GPQR, TSVQR, LS-LDMR, RQSVR, and URALTSVR for US girls data set.}\label{fusgirls}
\end{figure*}

\begin{figure*}[h!]
\centering
\subfigure[SVQR]{\includegraphics[width=0.35\textheight]{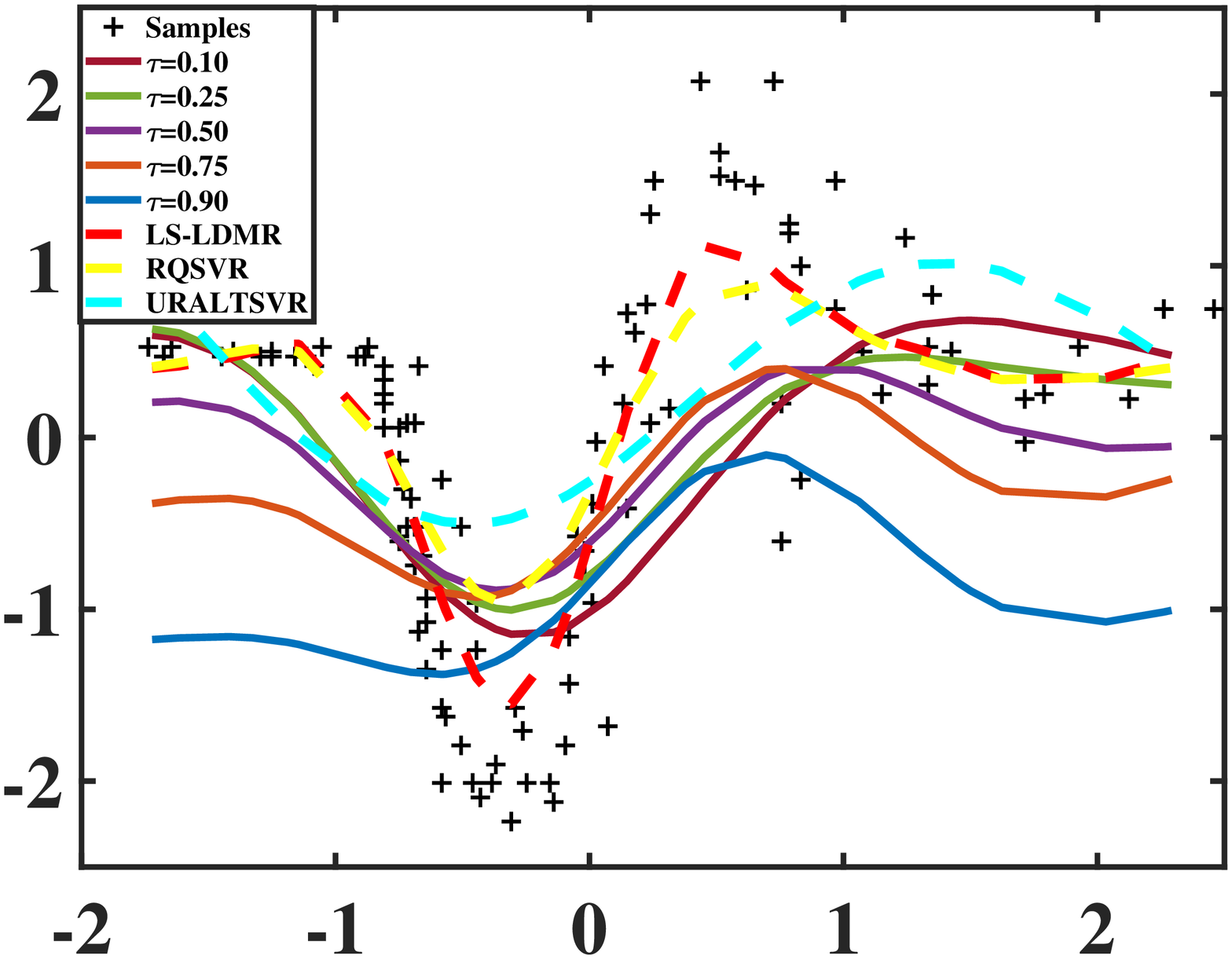}}
\subfigure[$\varepsilon$-SVQR]{\includegraphics[width=0.35\textheight]{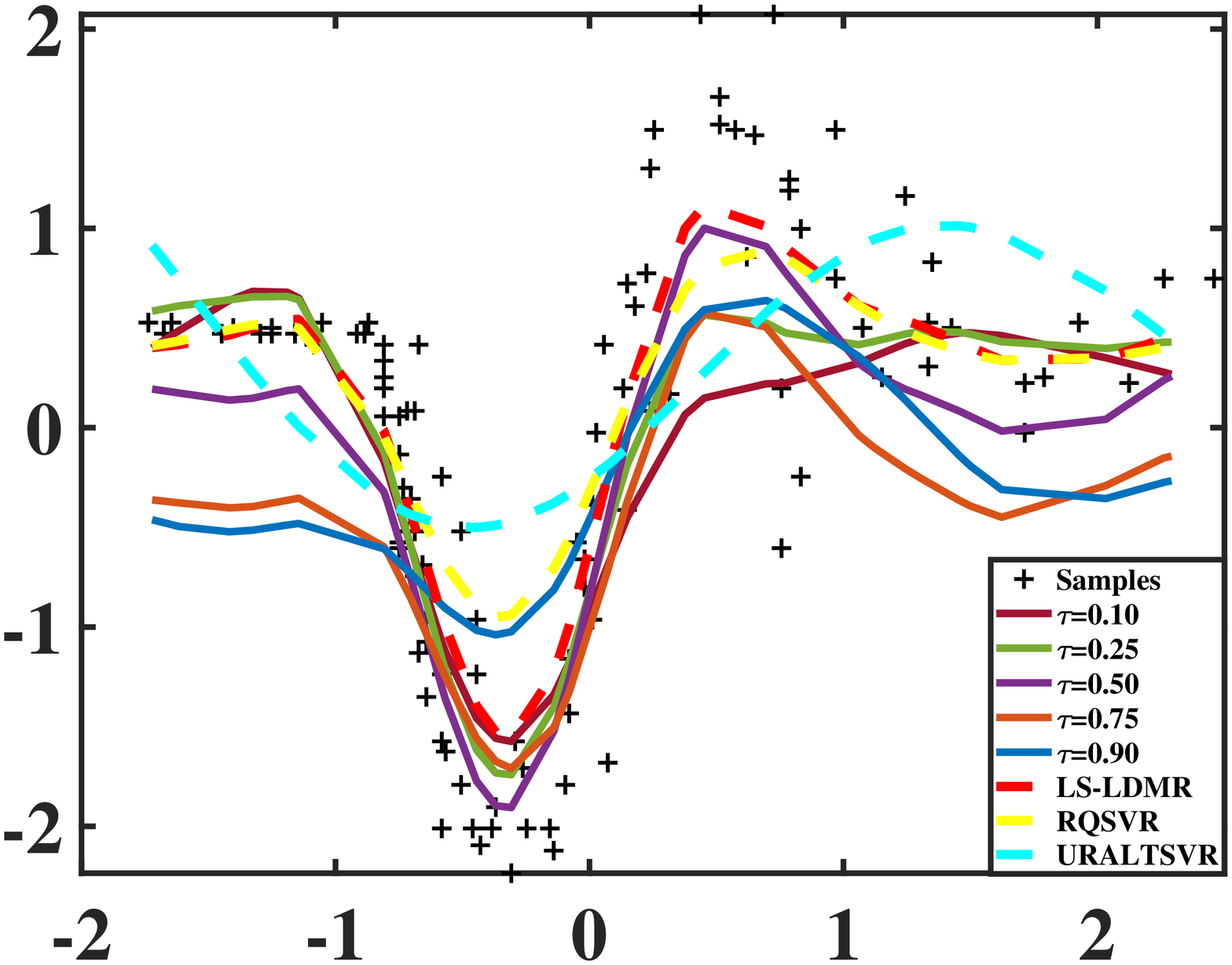}}
\subfigure[Online-SVQR]{\includegraphics[width=0.35\textheight]{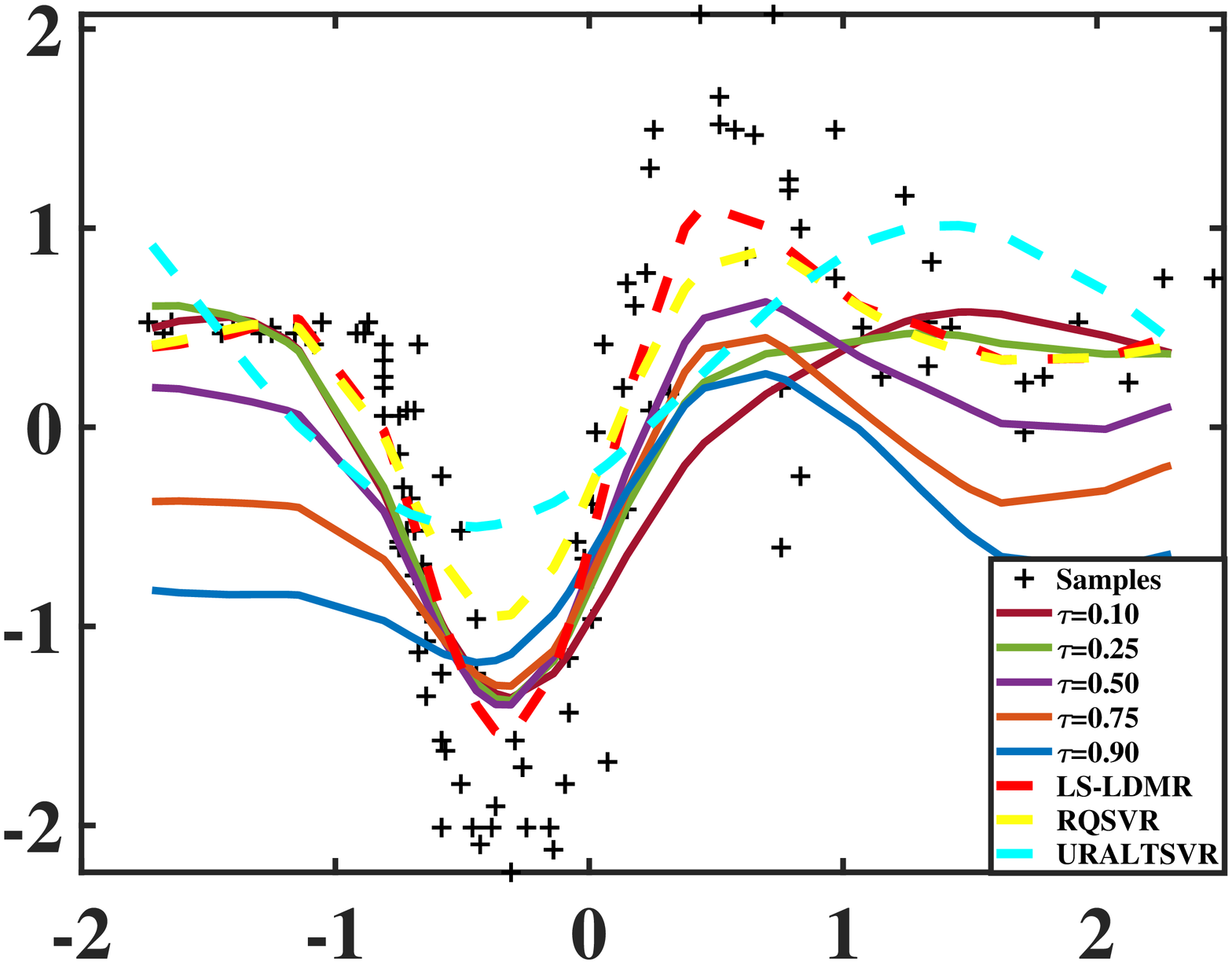}}
\subfigure[GPQR]{\includegraphics[width=0.35\textheight]{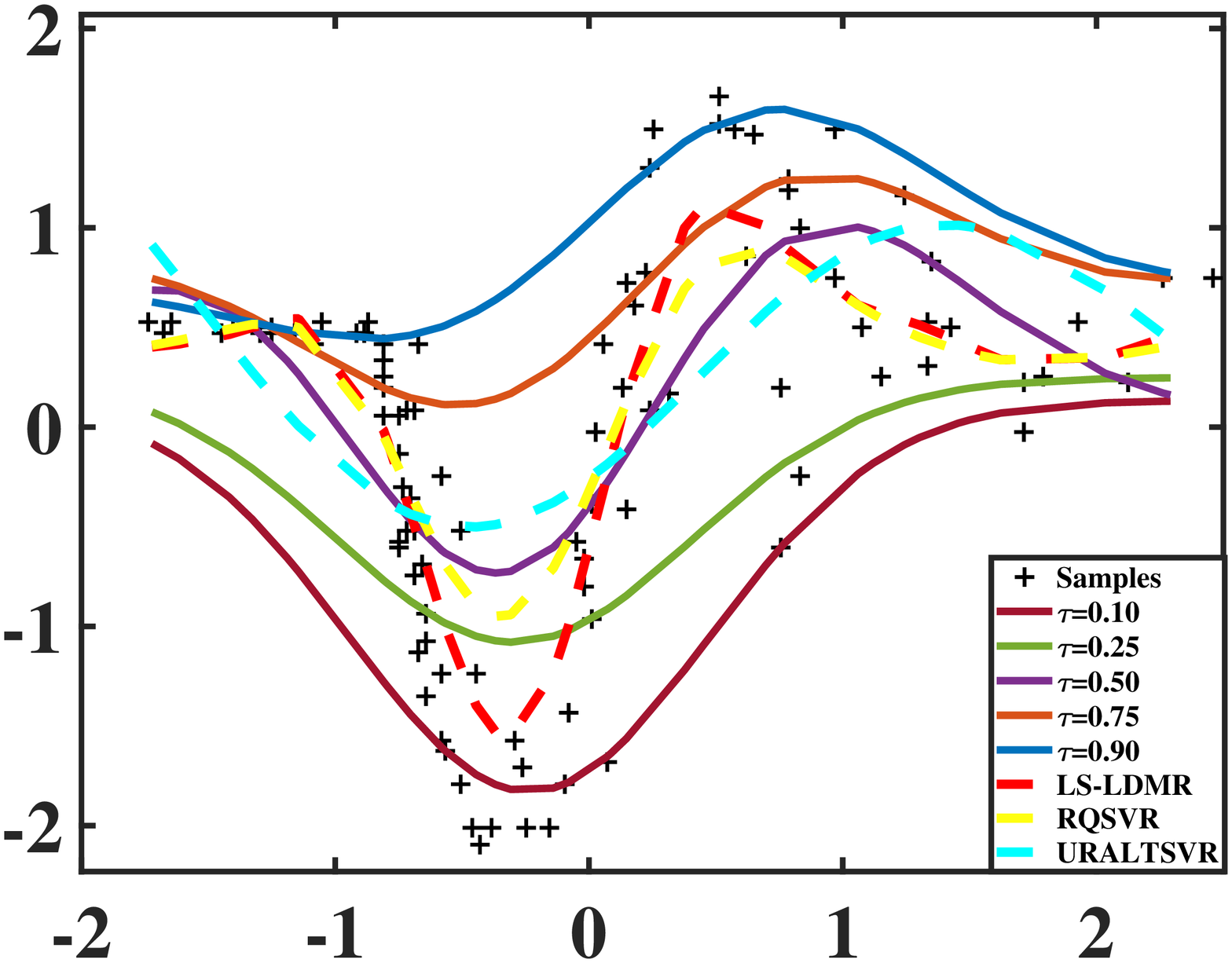}}
\subfigure[TSVQR]{\includegraphics[width=0.35\textheight]{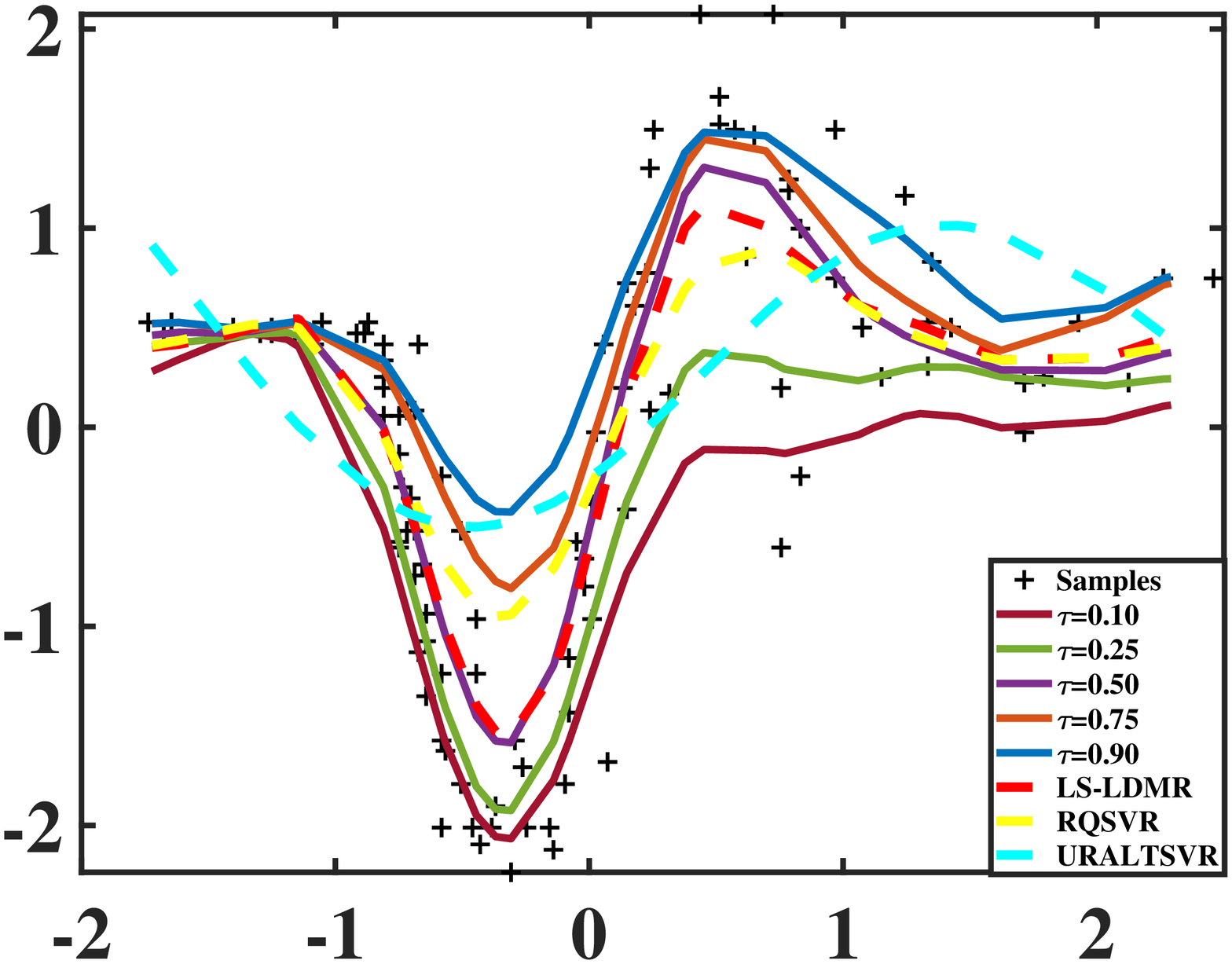}}
\caption{Regression results of SVQR, $\varepsilon$-SVQR, Online-SVQR, GPQR, TSVQR, LS-LDMR, RQSVR, and URALTSVR for Motorcycle data set.}\label{fmotor}
\end{figure*}

The optimal parameters of TSVQR for benchmark data sets are shown in Table \ref{par2}. Tables \ref{engel}-\ref{boston} list the regression results of real-world data sets for SVQR, $\varepsilon$-SVQR, Online-SVQR, GPQR and TSVQR, when $\tau=0.10$, 0.25, 0.50, 0.75, and $0.90$, and their corresponding regression performance is depicted in Figs. \ref{fengel}-\ref{fmotor}. Table \ref{benchmark} shows the regression results of benchmark data sets for LS-LDMR, RQSVR and URALTSVR.
From Fig. \ref{fengel}, we observe the tendency of the dispersion of food expenditure
to increase along with increases in the level of household income, indicating the heterogeneous statistical features in the ``Engel" data set.
From Fig. \ref{fengel}(e),
we find that different income groups have different marginal propensities for food expenditure.
Combining the regression results in Table~\ref{engel}, we see that TSVQR and GPQR yield a smaller Risk than SVQR, $\varepsilon$-SVQR, and Online-SVQR in most cases. This means that the statistical information at different quantile levels is well presented by TSVQR since the empirical quantile risk (Risk) uses the quantile parameter to measure the information capturing ability at different quantile levels. For the CPU time in Table \ref{engel}, we find that the training speed of TSVQR is faster than those of SVQR, $\varepsilon$-SVQR, Online-SVQR, and GPQR
as TSVQR only solves two smaller-sized QPPs without any equality constraint in the learning process and the dual coordinate descent algorithm is adopted to solve the QPPs.
Figs. \ref{fbonedata}-\ref{fmotor} demonstrate that TSVQR depicts more complete pictures of training data set than SVQR, $\varepsilon$-SVQR, and Online-SVQR.
The results in Tables \ref{bonedata}-\ref{benchmark} also confirm the effectiveness of TSVQR to demonstrate the heterogeneous statistical information in the data sets. Moreover, it can be seen that the differences between TSVQR and the other seven models with respect to RMSE and MAE are not significant. The main reason is that neither definition adopts the quantile parameter $\tau$. Therefore, it is worth exploring more reasonable criteria to evaluate quantile regression results in our future work. Regarding the training speed, we find that the training speed of TSVQR, LS-LDMR, and URALTSVR is also the fastest in all cases.
\begin{table}[h!]
	\footnotesize
	\centering
	\caption{The optimal parameters of Benchmark data sets selected by GACV.}
	\begin{tabular}{cccccc}
		\hline
		$\tau$&0.10&0.25&0.50&0.75&0.90  \\
		\hline
		Engel&&&&&\\
		$\rm C_1$&$\rm 2^{-1}$&$\rm 2^{-1}$&$\rm 2^{-1}$&$\rm 2^{-1}$&$\rm 2^{-1}$ \\
		$\rm C_2$&$\rm 2^{-1}$&$\rm 2^{-1}$&$\rm 2^{-1}$&$\rm 2^{-1}$&$\rm 2^{-1}$ \\
		P&$2^0$&$2^0$&$2^0$&$2^0$&$2^0$\\
		\hline
		Bone density&&&&&\\
		$\rm C_1$&$\rm 2^{-2}$&$\rm 2^{-2}$&$\rm 2^{-2}$&$\rm 2^{-2}$&$\rm 2^{-2}$ \\
		$\rm C_2$&$\rm 2^{-2}$&$\rm 2^{-2}$&$\rm 2^{-2}$&$\rm 2^{-2}$&$\rm 2^{-2}$ \\
		P&$2^0$&$2^0$&$2^0$&$2^0$&$2^0$\\
		\hline
		US girls&&&&&\\
		$\rm C_1$&$\rm 2^{-2}$&$\rm 2^{-2}$&$\rm 2^{-2}$&$\rm 2^{-2}$&$\rm 2^{-2}$ \\
		$\rm C_2$&$\rm 2^{-2}$&$\rm 2^{-2}$&$\rm 2^{-2}$&$\rm 2^{-2}$&$\rm 2^{-2}$ \\
		P&$2^1$&$2^1$&$2^1$&$2^1$&$2^1$\\
		\hline
		Motorcycle&&&&&\\
		$\rm C_1$&$\rm 2^{-1}$&$\rm 2^{-1}$&$\rm 2^{-1}$&$\rm 2^{-1}$&$\rm 2^{-1}$ \\
		$\rm C_2$&$\rm 2^{-1}$&$\rm 2^{-1}$&$\rm 2^{-1}$&$\rm 2^{-1}$&$\rm 2^{-1}$ \\
		P&$\rm 2^{-1}$&$\rm 2^{-1}$&$\rm 2^{-1}$&$\rm 2^{-1}$&$\rm 2^{-1}$\\
		\hline
		Boston housing&&&&&\\
		$\rm C_1$&$\rm 2^{-1}$&$\rm 2^{-1}$&$\rm 2^{-1}$&$\rm 2^{-1}$&$\rm 2^{-1}$ \\
		$\rm C_2$&$\rm 2^{-1}$&$\rm 2^{-1}$&$\rm 2^{-1}$&$\rm 2^{-1}$&$\rm 2^{-1}$ \\
		P&$2^0$&$2^0$&$2^0$&$2^0$&$2^0$\\
		\hline
	\end{tabular}
	\label{par2}
\end{table}

\begin{table*}[h!]
	\footnotesize
	\begin{center}
		\caption{Evaluation indices of Engel data set.}
		\label{engel}
		\begin{tabular}{@{}cccccc@{}}
			\hline
			$\tau$&Method&Risk&RMSE&MAE&CPU Time  \\
			\hline
			0.10&SVQR&0.1888&$\bold{0.4983}$&0.3809&0.2224\\
			&$\varepsilon$-SVQR&0.1925&$\bold{0.4983}$&$\bold{0.3786}$&0.0520\\
			&Online-SVQR&0.4258&1.2055&0.7492&0.1373  \\
			&GPQR&$\bold{0.1028}$&0.6508&0.5032&0.0518\\
			&TSVQR&0.1265&0.7867&0.7216&$\bold{0.0106}$\\
			\hline
			0.25&SVQR&0.2220&0.6240&0.4882&0.0904\\
			&$\varepsilon$-SVQR&0.2229&0.6236&0.4874&0.0411\\
			&Online-SVQR&0.4525&1.3422&0.8351&0.0658  \\
			&GPQR&$\bold{0.1518}$&$\bold{0.5885}$&$\bold{0.3991}$&0.0496\\
			&TSVQR&0.2323&0.7608&0.6659&$\bold{0.0091}$\\
			\hline
			0.50&SVQR&0.3066&0.7722&0.6133&0.0264\\
			&$\varepsilon$-SVQR&$\bold{0.3028}$&$\bold{0.7646}$&$\bold{0.6056}$&0.0559\\
			&Online-SVQR&0.4695&1.5084&0.9390&0.0412  \\
			&GPQR&0.5266&0.8781&0.8332&0.0909\\
			&TSVQR&0.3150&0.7887&0.6300&$\bold{0.0056}$\\
			\hline
			0.75&SVQR&0.4049&0.9126&0.7237&0.0335\\
			&$\varepsilon$-SVQR&0.4029&0.9102&0.7211&0.0450\\
			&Online-SVQR&0.4898&1.7388&1.0769&0.0392  \\
			&GPQR&0.4118&0.9419&0.7027&0.0312\\
			&TSVQR&$\bold{0.3393}$&$\bold{0.9015}$&$\bold{0.6852}$&$\bold{0.0082}$\\
			\hline
			0.90&SVQR&0.4669&0.9929&0.7866&0.0265\\
			&$\varepsilon$-SVQR&0.4618&0.9875&0.7810&0.0784\\
			&Online-SVQR&0.5073&1.9096&1.1837&0.0525  \\
			&GPQR&0.3586&$\bold{0.7243}$&0.7417&0.0294\\
			&TSVQR&$\bold{0.2943}$&1.0193&$\bold{0.7382}$&$\bold{0.0065}$\\
			\hline
		\end{tabular}
	\end{center}
\end{table*}

\begin{table*}[h!]
	\footnotesize
	\begin{center}
		\caption{Evaluation indices of Bone density data set.}
		\label{bonedata}
		\begin{tabular}{@{}cccccc@{}}
			\hline
			$\tau$&Method&Risk&RMSE&MAE&CPU Time  \\
			\hline
			0.10&SVQR&0.0899&$\bold{0.1490}$&$\bold{0.1369}$&0.1681\\
			&$\varepsilon$-SVQR&0.1797&0.5471&0.5027&0.1332\\
			&Online-SVQR&0.1152&0.1816&0.1682&0.1506  \\
			&GPQR&0.1204&0.8365	&0.9482	&	0.0963\\
			&TSVQR&$\bold{0.0829}$&0.8549&0.8293&$\bold{0.0097}$\\
			\hline
			0.25&SVQR&0.1385&$\bold{0.2510}$&$\bold{0.2317}$&0.1280\\
			&$\varepsilon$-SVQR&0.1504&0.2641&0.2438&0.1401\\
			&Online-SVQR&0.1724&0.2964&0.2718&0.1341  \\
			&GPQR&0.2091&0.9018&0.6947&0.0945\\
			&TSVQR&$\bold{0.1170}$&0.5639&0.4680&$\bold{0.0179}$\\
			\hline
			0.50&SVQR&$\bold{0.2039}$&$\bold{0.4448}$&$\bold{0.4078}$&0.2035\\
			&$\varepsilon$-SVQR&0.2399&0.5416&0.4797&0.1620\\
			&Online-SVQR&0.2219&0.4920&0.4437&0.1828  \\
			&GPQR&0.3116&0.8114&0.6231&0.1066\\
			&TSVQR&0.2138&0.4763&0.4276&$\bold{0.0134}$\\
			\hline
			0.75&SVQR&0.2208&$\bold{0.5720}$&$\bold{0.5233}$&0.1281\\
			&$\varepsilon$-SVQR&0.2806&0.6786&0.5239&0.1210\\
			&Online-SVQR&$\bold{0.2199}$&0.5725&0.5236&0.1245  \\
			&GPQR&0.2719&0.7097&0.8073&0.0941\\
			&TSVQR&0.2593&0.9013&0.7808&$\bold{0.0243}$\\
			\hline
			0.90&SVQR&0.1817&$\bold{0.5445}$&$\bold{0.5003}$&0.1359\\
			&$\varepsilon$-SVQR&0.2640&0.6987&0.6345&0.1161\\
			&Online-SVQR&0.2126&0.6425&0.5924&0.1260  \\
			&GPQR&0.1630&0.7442&0.6457&0.0878\\
			&TSVQR&$\bold{0.1584}$&1.3002&1.1054&$\bold{0.0106}$\\
			\hline
		\end{tabular}
	\end{center}
\end{table*}

\begin{table*}[h!]
	\footnotesize
	\begin{center}
		\caption{Evaluation indices of US girls data set.}
		\label{usgirls}
		\begin{tabular}{@{}cccccc@{}}
			\hline
			$\tau$&Method&Risk&RMSE&MAE&CPU Time  \\
			\hline
			0.10&SVQR&0.1220&0.9269&0.7635&21.6099\\
			&$\varepsilon$-SVQR&0.1891&0.7500&0.6611&20.4738\\
			&Online-SVQR&0.1513&0.8300&0.7038&21.0418  \\
			&GPQR&$\bold{0.0446}$&$\bold{0.6439}$&$\bold{0.4128}$&6.7215\\
			&TSVQR&0.1896&0.7757&0.6762&$\bold{2.6566}$\\
			\hline
			0.25&SVQR&0.3126&0.7758&0.7141&15.9121\\
			&$\varepsilon$-SVQR&0.3573&0.7562&0.6931&14.6243\\
			&Online-SVQR&0.3337&0.7581&0.7011&15.2682  \\
			&GPQR&0.6087&$\bold{0.7303}$&$\bold{0.3037}$&6.2507\\
			&TSVQR&$\bold{0.3029}$&0.7958&0.7200&$\bold{2.2384}$\\
			\hline
			0.50&SVQR&0.3989&0.8591&0.7978&14.5569\\
			&$\varepsilon$-SVQR&0.3962&0.8539&0.7923&13.4608\\
			&Online-SVQR&0.3973&0.8556&0.7947&14.0089  \\
			&GPQR&$\bold{0.1261}$&$\bold{0.4454}$&$\bold{0.2522}$&6.4887\\
			&TSVQR&0.3948&0.8583&0.7895&$\bold{1.7579}$\\
			\hline
			0.75&SVQR&0.4224&0.9895&0.9126&15.8177\\
			&$\varepsilon$-SVQR&0.4966&0.9864&0.9187&16.7182\\
			&Online-SVQR&0.4576&0.9776&0.9120&16.2679  \\
			&GPQR&0.5256&$\bold{0.4424}$&$\bold{0.2512}$&6.4887\\
			&TSVQR&$\bold{0.4153}$&0.9959&0.9100&$\bold{2.0672}$\\
			\hline
			0.90&SVQR&0.5937&1.1482&1.0676&20.6675\\
			&$\varepsilon$-SVQR&0.6405&1.1899&1.1011&20.9365\\
			&Online-SVQR&0.6169&1.1685&1.0839&20.8021  \\
			&GPQR&$\bold{0.0834}$&$\bold{0.6751}$&$\bold{0.5229}$&6.1780\\
			&TSVQR&0.3694&1.2542&1.1096&$\bold{2.5321}$\\
			\hline
		\end{tabular}
	\end{center}
\end{table*}

\begin{table*}[h!]
	\footnotesize
	\begin{center}
		\caption{Evaluation indices of Motorcycle data set.}
		\label{motor}
		\begin{tabular}{@{}cccccc@{}}
			\hline
			$\tau$&Method&Risk&RMSE&MAE&CPU Time  \\
			\hline
			0.10&SVQR&0.3192&0.5218&0.4731&0.0406\\
			&$\varepsilon$-SVQR&0.3090&$\bold{0.5158}$&0.4654&0.0312\\
			&Online-SVQR&0.2619&0.7064&0.6194&0.0360  \\
			&GPQR&0.1241&0.8311&0.9398&0.0122\\
			&TSVQR&$\bold{0.1226}$&0.5645&$\bold{0.3836}$&$\bold{0.0116}$\\
			\hline
			0.25&SVQR&$\bold{0.1856}$&$\bold{0.4396}$&$\bold{0.3615}$&0.0303\\
			&$\varepsilon$-SVQR&0.3730&0.6070&0.5606&0.0525\\
			&Online-SVQR&0.2766&0.6789&0.5934&0.0414  \\
			&GPQR&0.2546&0.8739&0.7295&0.0112\\
			&TSVQR&0.2266&0.4884&0.4181&$\bold{0.0088}$\\
			\hline
			0.50&SVQR&$\bold{0.1560}$&$\bold{0.4173}$&$\bold{0.3119}$&0.0333\\
			&$\varepsilon$-SVQR&0.1680&0.5027&0.3360&0.0273\\
			&Online-SVQR&0.2375&0.6347&0.4751&0.0303  \\
			&GPQR&0.2632&0.6751&0.5263&0.0126\\
			&TSVQR&0.2595&0.6318&0.5191&$\bold{0.0114}$\\
			\hline
			0.75&SVQR&0.3320&0.5514&0.4834&0.0413\\
			&$\varepsilon$-SVQR&0.3287&$\bold{0.5486}$&$\bold{0.4796}$&0.0305\\
			&Online-SVQR&0.3737&0.6454&0.5315&0.0359  \\
			&GPQR&0.2003&0.7043&0.7043&0.0117\\
			&TSVQR&$\bold{0.1643}$&0.7127&0.6093&$\bold{0.0115}$\\
			\hline
			0.90&SVQR&0.5958&0.7962&0.7112&0.0287\\
			&$\varepsilon$-SVQR&0.2755&$\bold{0.4762}$&$\bold{0.4222}$&0.0341\\
			&Online-SVQR&0.5792&0.7512&0.6654&0.0314  \\
			&GPQR&0.0997&0.8170&0.9018&0.0188\\
			&TSVQR&$\bold{0.0784}$&0.8134&0.7245&$\bold{0.0130}$\\
			\hline
		\end{tabular}
	\end{center}
\end{table*}

\begin{table*}[h!]
	\footnotesize
	\begin{center}
		\caption{Evaluation indices of Boston housing data set.}
		\label{boston}
		\begin{tabular}{@{}cccccc@{}}
			\hline
			$\tau$&Method&Risk&RMSE&MAE&CPU Time  \\
			\hline
			0.10&SVQR&0.2112&$\bold{0.2431}$&0.2385&0.1543\\
			&$\varepsilon$-SVQR&0.2019&0.2557&0.2439&0.1343\\
			&Online-SVQR&0.2034&0.2437&0.2349&0.1442  \\
			&GPQR&0.1827&0.5746&0.4695&0.0610\\
			&TSVQR&$\bold{0.0183}$&0.3404&$\bold{0.1829}$&$\bold{0.0398}$\\
			\hline
			0.25&SVQR&0.1962&$\bold{0.2727}$&0.2678&0.1440\\
			&$\varepsilon$-SVQR&0.1955&0.2997&0.2840&0.1457 \\
			&Online-SVQR&0.1926&0.2789&0.2694&0.1449  \\
			&GPQR&0.3375&0.6474&0.5164&0.0491\\
			&TSVQR&$\bold{0.0429}$&0.3192&$\bold{0.1711}$&$\bold{0.0427}$ \\
			\hline
			0.50&SVQR&0.0855&$\bold{0.1745}$&0.1710 &0.1423\\
			&$\varepsilon$-SVQR&0.1053&0.2622&0.2106&0.1313\\
			&Online-SVQR&0.0838&0.1889&0.1676&0.1369  \\
			&GPQR&0.3633&0.8783&0.7267&0.0521\\
			&TSVQR&$\bold{0.0743}$&0.2851&$\bold{0.1486}$&$\bold{0.0346}$\\
			\hline
			0.75&SVQR&$\bold{0.0367}$&$\bold{0.1291}$&0.1260&0.1471\\
			&$\varepsilon$-SVQR&0.0862&0.2141&0.1388&0.1627\\
			&Online-SVQR&0.0501&0.1351&0.1096&0.1549  \\
			&GPQR&0.2233&0.8230&0.8635&0.0684\\
			&TSVQR&0.0641&0.1711&$\bold{0.0888}$&$\bold{0.0210}$\\
			\hline
			0.90&SVQR&$\bold{0.0134}$&0.1365&0.1331&0.1377\\
			&$\varepsilon$-SVQR&0.0319&0.1419&0.1299&0.1509\\
			&Online-SVQR&0.0172&0.1280&0.1207&0.1443  \\
			&GPQR&0.4833&0.9585&0.8493&0.1629\\
			&TSVQR&0.0245&$\bold{0.0704}$&$\bold{0.0392}$&$\bold{0.0301}$\\
			\hline
		\end{tabular}
	\end{center}
\end{table*}

\begin{table*}[h!]
	\footnotesize
	\begin{center}
		\caption{Evaluation indices of Benchmark data sets for LS-LDMR, RQSVR and URALTSVR.}
		\label{benchmark}
		\begin{tabular}{@{}cccccc@{}}
			\hline
			Data set&Method&RMSE&MAE&MAPE&CPU Time  \\
			\hline
			Engel&LS-LDMR&0.7071&0.5671&1.0211&0.0070\\
			&RQSVR&$\bold{0.5996}$&0.3412&$\bold{1.0096}$&0.0164\\
			&URALTSVR&0.6869&$\bold{0.3314}$&1.0428&$\bold{0.0023}$  \\
			\hline
			Bone density&LS-LDMR&$\bold{0.5221}$&$\bold{0.4748}$&$\bold{1.3127}$&0.0318\\
			&RQSVR&0.8174&0.6249&2.6513&0.0953 \\
			&URALTSVR&0.8185&0.6547&3.0161&$\bold{0.0147}$ \\
			\hline
			US girls&LS-LDMR&0.9000&0.8321&0.9710&$\bold{0.2411}$\\
			&RQSVR&0.4413&$\bold{0.2519}$&0.9336&11.5615\\
			&URALTSVR&$\bold{0.4386}$&0.2600&$\bold{0.8261}$&2.1039  \\
			\hline
			Motorcycle&LS-LDMR&$\bold{0.5839}$&$\bold{0.5099}$&$\bold{1.0010}$&0.0078\\
			&RQSVR&0.7563&0.5338&1.7136&$\bold{0.0072}$\\
			&URALTSVR&0.7743&0.5894&3.8043&0.0117  \\
			\hline
			Boston housing&LS-LDMR&$\bold{0.2891}$&$\bold{0.2519}$&$\bold{0.5622}$&$\bold{0.0026}$\\
			&RQSVR&0.9476&0.7607&5.2051&0.0938\\
			&URALTSVR&0.8971&0.6953&3.3802&0.0174 \\
			\hline
		\end{tabular}
	\end{center}
\end{table*}

\subsection{Statistical test}
To fairly evaluate the experimental results of SVQR, $\varepsilon$-SVQR, Online-SVQR, GPQR, and TSVQR on six artificial data sets and five benchmark data sets, we adopt the Friedman test (\citeauthor{demvsar2006statistical}, \citeyear{demvsar2006statistical}; \citeauthor{ye2020robust}, \citeyear{ye2020robust}).
Table \ref{Friedman test} lists the Friedman test results of SVQR, $\varepsilon$-SVQR, Online-SVQR, GPQR, and TSVQR. At the $10\%$ significance level, the critical values of $\chi^2 (4)$ and $F(4, 32)$ are $7.78$ and $2.14$, respectively. From Table \ref{Friedman test}, we find that
the $\chi^2$ and $F_F$ values of the Risk and the CPU time at different quantile levels are much greater than their corresponding critical values. Therefore, the null hypothesis is failed to acceptance level that means all five methods are not equivalent to each other.

We further use Nemenyi test (\citeauthor{demvsar2006statistical}, \citeyear{demvsar2006statistical}; \citeauthor{ye2020robust}, \citeyear{ye2020robust}) to compare SVQR, $\varepsilon$-SVQR, Online-SVQR, GPQR, and TSVQR in terms of the Risk and the CPU time. The critical value of Nemenyi test at the $10\%$ significance level is $1.8328$. Nemenyi test results for the Risk and the CPU time are list in Table \ref{Nemenyi test}. From Table \ref{Nemenyi test}, we find that the differences between TSVQR and other models with respect to the Risk are larger than $1.8328$ in most cases. Let us compare TSVQR with SVQR, $\varepsilon$-SVQR, and Online-SVQR. When $\tau=0.10$, the difference values are $2.0000$, $2.3636$, and $2.2727$, respectively, which are more than $1.8328$. It is evident that regression performance of TSVQR is superior to SVQR, $\varepsilon$-SVQR, and Online-SVQR. When $\tau=0.90$, we can get the same conclusion. For the running time, we find that the training speed of TSVQR is significantly faster than SVQR, Online-SVQR, and GPQR in all cases, since the difference between them is larger than the critical value.

We further adopt Friedman test to evaluate the training speed of LS-LDMR, URALTSVR, RQSVR, SVQR, $\varepsilon$-SVQR, Online-SVQR, GPQR, and TSVQR.
The $\chi^2$ value of CPU time is 57.88, which is larger than the critical value 14.067 at the $5\%$ significance level. It can be concluded that the null hypothesis is rejected and all eight approaches are concluded to be not equivalent. The Nemenyi test has been applied for pairwise comparative analysis on SVQR, $\varepsilon$-SVQR, Online-SVQR, GPQR, TSVQR, LS-LDMR, RQSVR and URALTSVR. Tabel \ref{Nemenyi test2} lists the results of the Nemenyi test. Comparing the training speed of LS-LDMR with SVQR, $\varepsilon$-SVQR, Online-SVQR, GPQR, and RQSVR, we find the difference values are 5.4545, 4.2727, 5.2727, 4.9090, and 3.4545, respectively, which are more than the critical value 2.69 at the $5\%$ significance level, indicating that the training speed of LS-LDMR is superior to SVQR, $\varepsilon$-SVQR, Online-SVQR, GPQR, and RQSVR. Checking the difference of LS-LDMR with URALTSVR and TSVQR, we find that their difference values are 1.0909 and 1.3636, respectively, which are less than the critical value 2.69. Therefore, there is no significant difference between the training speed of LS-LDMR, URALTSVR and TSVQR.
\begin{table}
	\footnotesize
	\begin{center}
		\caption{Friedman test of SVQR, $\varepsilon$-SVQR, Online-SVQR, GPQR, and TSVQR.}
		\label{Friedman test}
		\begin{tabular}{@{}clcc@{}}
			\hline
			$\tau$&Evaluation criteria&Value of $\chi^2$ &Value of $F_F$ \\
			\hline
			0.10&Risk& $\bold{24.1454}$  & $\bold{12.1612}$  \\
			&CPU Time& $\bold{27.5636}$  & $\bold{16.7699} $\\
			\hline
			0.25&Risk& $\bold{17.2363} $ & $ \bold{6.4402} $\\
			&CPU Time& $\bold{31.2363}$  & $\bold{24.4729} $\\
			\hline
			0.50&Risk& $\bold{12.8363}$   & $\bold{4.1190} $\\
			&CPU Time& $\bold{26.8727}$  & $\bold{15.6900}$\\
			\hline
			0.75&Risk& $\bold{13.3818}$  & $\bold{4.3705}$ \\
			&CPU Time& $\bold{23.9273}$  & $\bold{11.9203}$\\
			\hline
			0.90&Risk& $\bold{21.7454}$  & $\bold{9.7712}$ \\
			&CPU Time& $\bold{22.6182} $ & $\bold{10.5782}$ \\
			\hline
		\end{tabular}
	\end{center}
\end{table}

\begin{table}
	\footnotesize
	\begin{center}
		\caption{Nemenyi test of SVQR, $\varepsilon$-SVQR, Online-SVQR, GPQR, and TSVQR.}
		\label{Nemenyi test}
		\begin{tabular}{@{}clcc@{}}
			\hline
			$\tau$&Comparison&Risk &CPU Time \\
			\hline
			0.10&TSVQR vs. SVQR                & $\bold{2.0000 }$          & $\bold{3.1818}$ \\
			    &TSVQR vs. $\varepsilon$-SVQR  & $\bold{2.3636}$          & 1.8181   \\
                &TSVQR vs. Online-SVQR         & $\bold{2.2727}$         & $\bold{2.9090}$\\
                &TSVQR vs. GPQR               & 0.1818                   & $\bold{2.0909}$\\
			\hline
			0.25&TSVQR vs. SVQR               & 1.7272              & $\bold{2.3636}$ \\
			    &TSVQR vs. $\varepsilon$-SVQR & $\bold{2.2727}$    & $\bold{2.7272}$\\
                &TSVQR vs. Online-SVQR       & $\bold{2.5454 }$    &$\bold{2.8181 }$\\
               &TSVQR vs. GPQR               & 1.6363              &$\bold{2.3636}$ \\
			\hline
			0.50&TSVQR vs. SVQR              & 0.9090               & $\bold{2.7272}$ \\
			&TSVQR vs. $\varepsilon$-SVQR    & 1.7272               &$\bold{2.0000}$\\
            &TSVQR vs. Online-SVQR           & 1.4545               &$\bold{2.4545}$\\
            &TSVQR vs. GPQR                  & $\bold{1.9090}$     &$\bold{2.9090}$ \\
			\hline
			0.75&TSVQR vs. SVQR              & 1.8181             & $\bold{2.4545}$ \\
			    &TSVQR vs. $\varepsilon$-SVQR & $\bold{2.0000}$    &$\bold{2.6363}$\\
            &TSVQR vs. Online-SVQR            & 1.8181             &$\bold{2.9090}$ \\
            &TSVQR vs. GPQR                  &$\bold{2.0909}$     &$\bold{2.0000}$ \\
			\hline
			0.90&TSVQR vs. SVQR              & $\bold{2.2727}$    & $\bold{2.3636}$ \\
			&TSVQR vs. $\varepsilon$-SVQR    &$\bold{2.2727 }$    &$\bold{2.4545}$\\
            &TSVQR vs. Online-SVQR           & $\bold{2.4545}$    &$\bold{2.8181}$ \\
            &TSVQR vs. GPQR                  & 0.7272            &$\bold{2.3636}$\\
			\hline
		\end{tabular}
	\end{center}
\end{table}

\begin{table}
 \footnotesize
 \begin{center}
  \caption{Nemenyi test of SVQR, $\varepsilon$-SVQR, Online-SVQR, GPQR, TSVQR, LS-LDMR, RQSVR and URALTSVR.}
  \label{Nemenyi test2}
  \begin{tabular}{@{}ll@{}}
  \hline
  Comparison & CPU Time \\
   \hline
   LS-LDMR vs.   SVQR                  & 5.4545  \\
   LS-LDMR vs. $\varepsilon$-SVQR      &  4.2727 \\
   LS-LDMR vs. Online-SVQR             & 5.2727  \\
   LS-LDMR vs. GPQR                    &  4.9090  \\
   LS-LDMR vs. RQSVR                   & 3.4545  \\
   LS-LDMR vs. URALTSVR                & 1.0909  \\
   LS-LDMR vs. TSVQR                   &  1.3636 \\
   \hline
  \end{tabular}
 \end{center}
\end{table}

\subsection {The quantile parameter and analysis}
\begin{figure}
\centering
\subfigure[Risk]{\includegraphics[width=0.35\textheight]{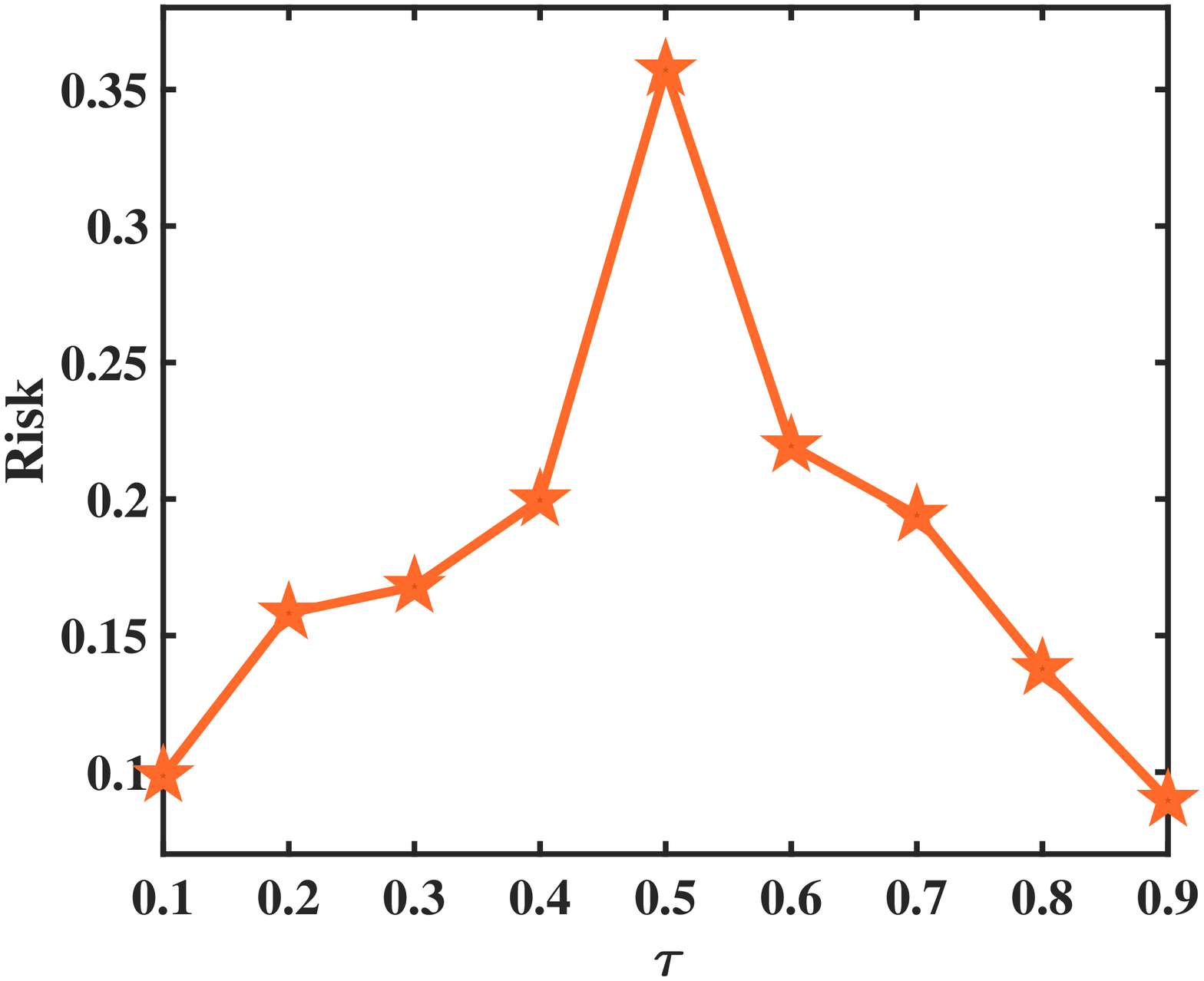}}
\subfigure[RMSE]{\includegraphics[width=0.35\textheight]{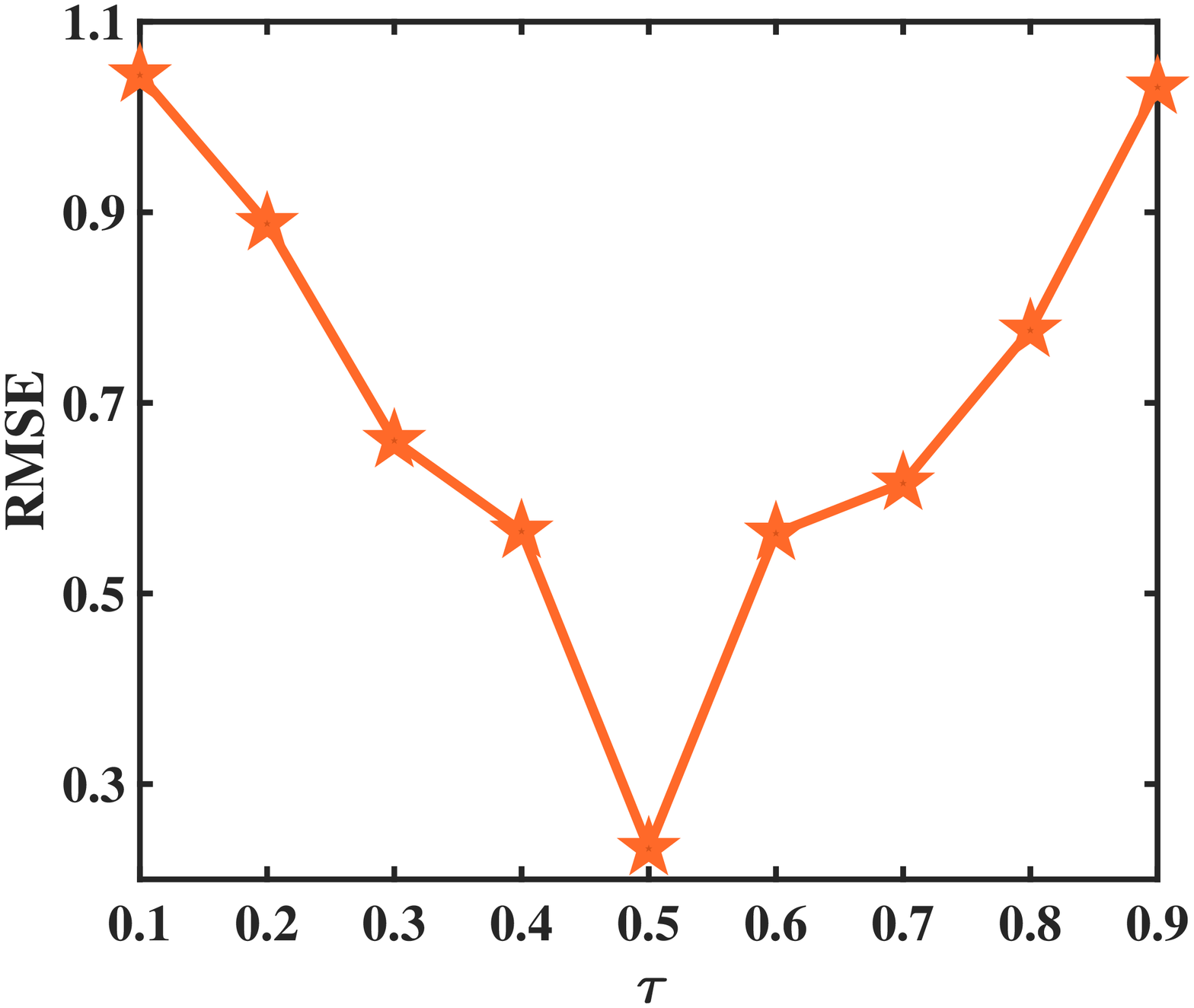}}
\subfigure[MAE]{\includegraphics[width=0.35\textheight]{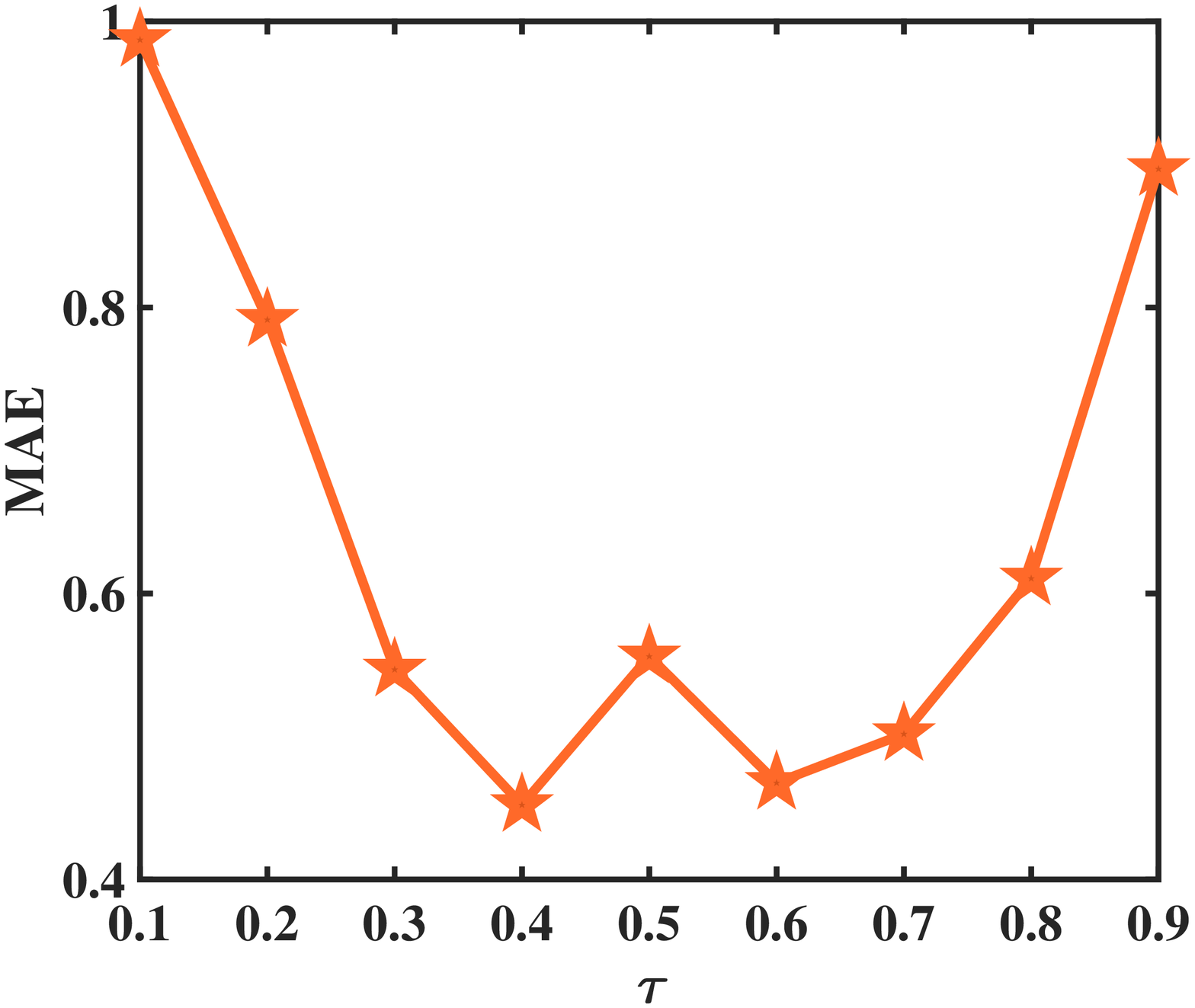}}
\subfigure[Time]{\includegraphics[width=0.35\textheight]{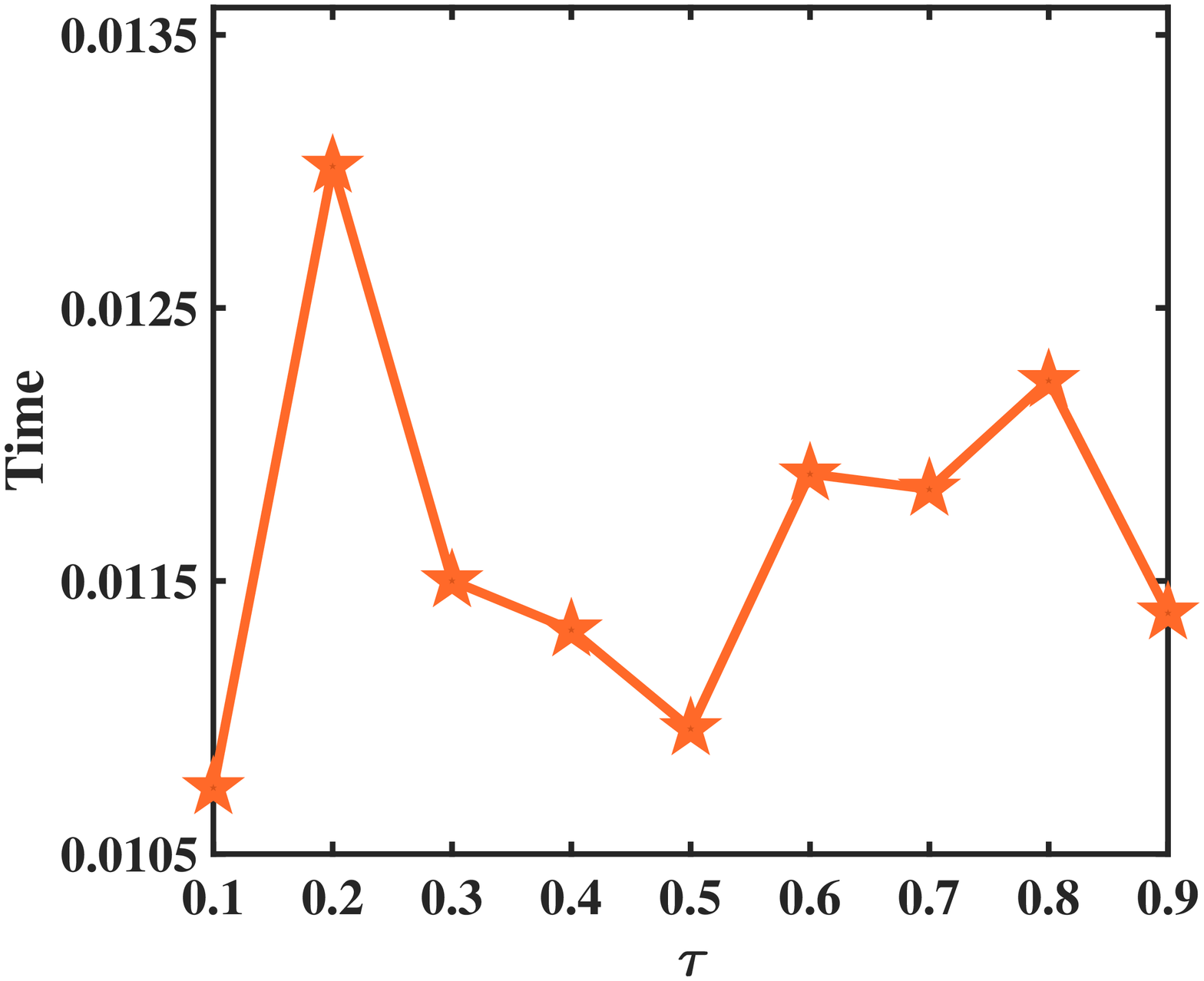}}
\caption{The influence of $\tau$ on the Risk, RMSE, MAE, and CPU Time. Parameters $C_1$, $C_2$, and $P$ are fixed as the optimal values.}\label{sensitivity}
\end{figure}

\begin{figure}
\centering
\subfigure[]{\includegraphics[width=0.6\textheight]{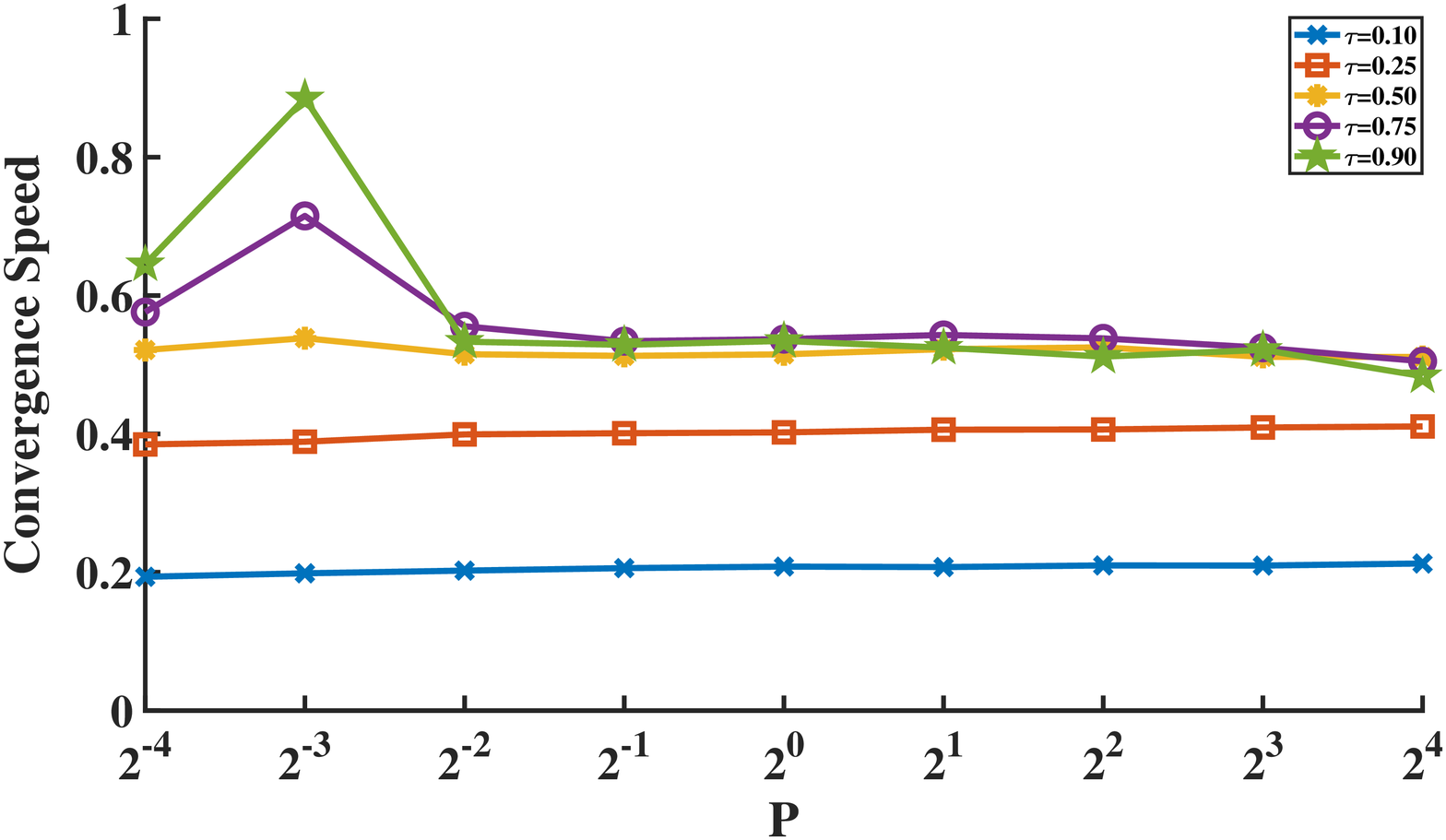}}
\subfigure[]{\includegraphics[width=0.6\textheight]{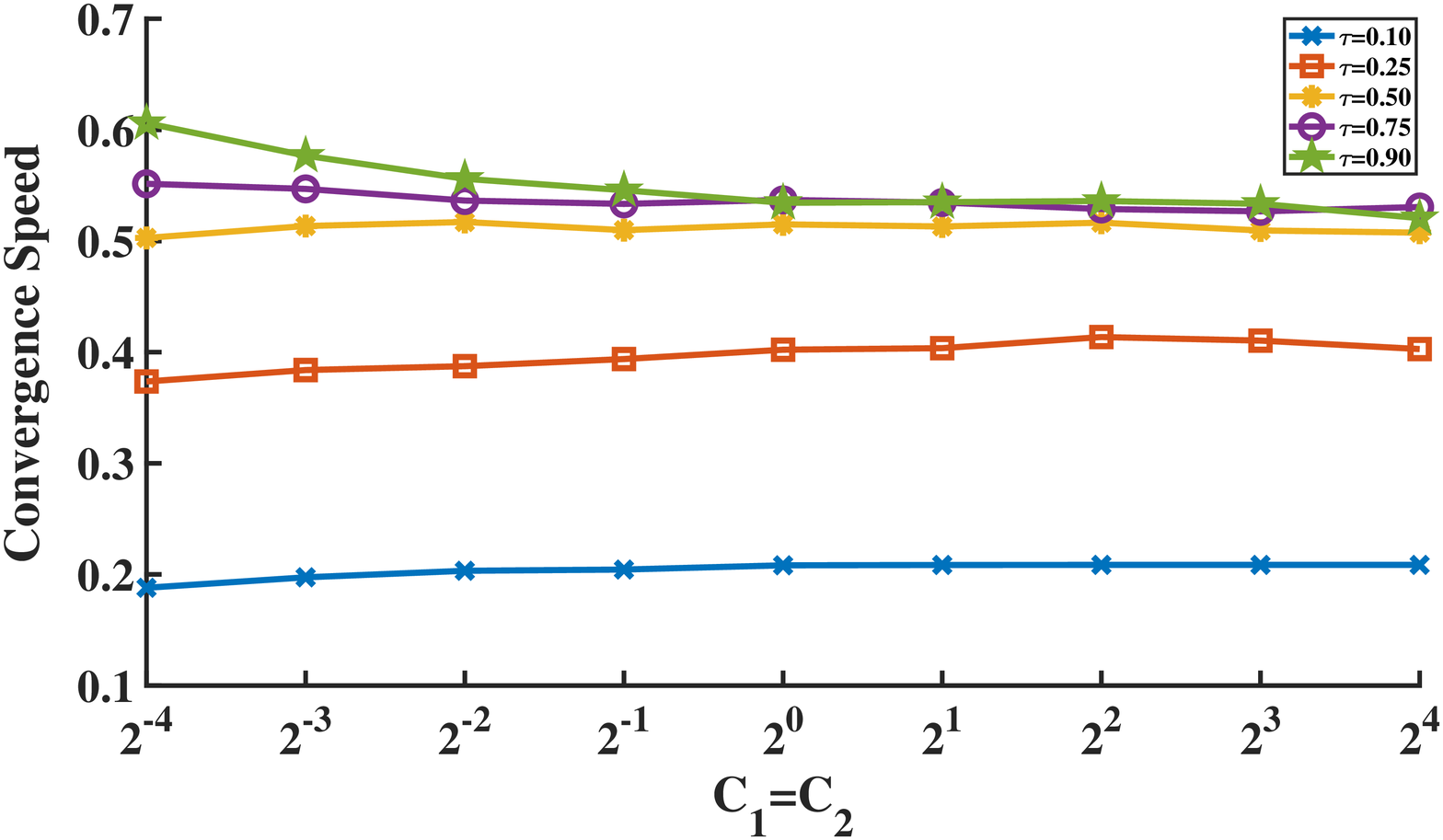}}
\caption{Convergence speed of TSVQR ($\tau=0.10, 0.25, 0.50, 0.75,$ and $0.90$).}\label{convergence}
\end{figure}

In this part, we adopt immunoglobulin G Data (\citeauthor{isaacs1983serum}, \citeyear{isaacs1983serum}) to analyze the effects of the quantile parameter $\tau$ on the regression results.
This data set comprises the serum concentration of immunoglobulin G in 298 children aged from 6 months to 6 years.
To test the influence of the parameter $\tau$ on the Risk, RMSE, MAE, and MAPE, we first fix the
parameters $C_1$, $C_2$, and $P$ as the optimal values selected by the grid search technique. Fig. \ref{sensitivity}
illustrates the influence of the parameter $\tau$ on the regression results. From Fig. \ref{sensitivity}, we observe that
as parameter $\tau$ increases, the value of Risk increases and then decreases,
which means that parameter $\tau$ has a strong influence on the regression results.
As $\tau$ increases, the value of RMSE decreases and then increases.
When $\tau=0.50$, Risk reaches a maximum value and RMSE reaches a minimum value.
From Fig. \ref{sensitivity}(d), we find that the curve of CPU time fluctuates slightly around approximate value of $0.0115$.
Fig. \ref{convergence} demonstrates the effects of parameters $C_1$, $C_2$, and $P$ on the convergence speed, when $\tau=0.10, 0.25, 0.50, 0.75,$ and $0.90$. From Fig. \ref{convergence}, we find that as parameter $\tau$ increases, the convergence speed decreases and then approximately remains the same. As parameters $P$ or $C_1$ and $C_2$ increase, all lines in Fig. \ref{convergence} fluctuate slightly.

\begin{figure}[htbp]
\centering
\subfigure[$\tau=0.10$]{\includegraphics[width=0.35\textheight]{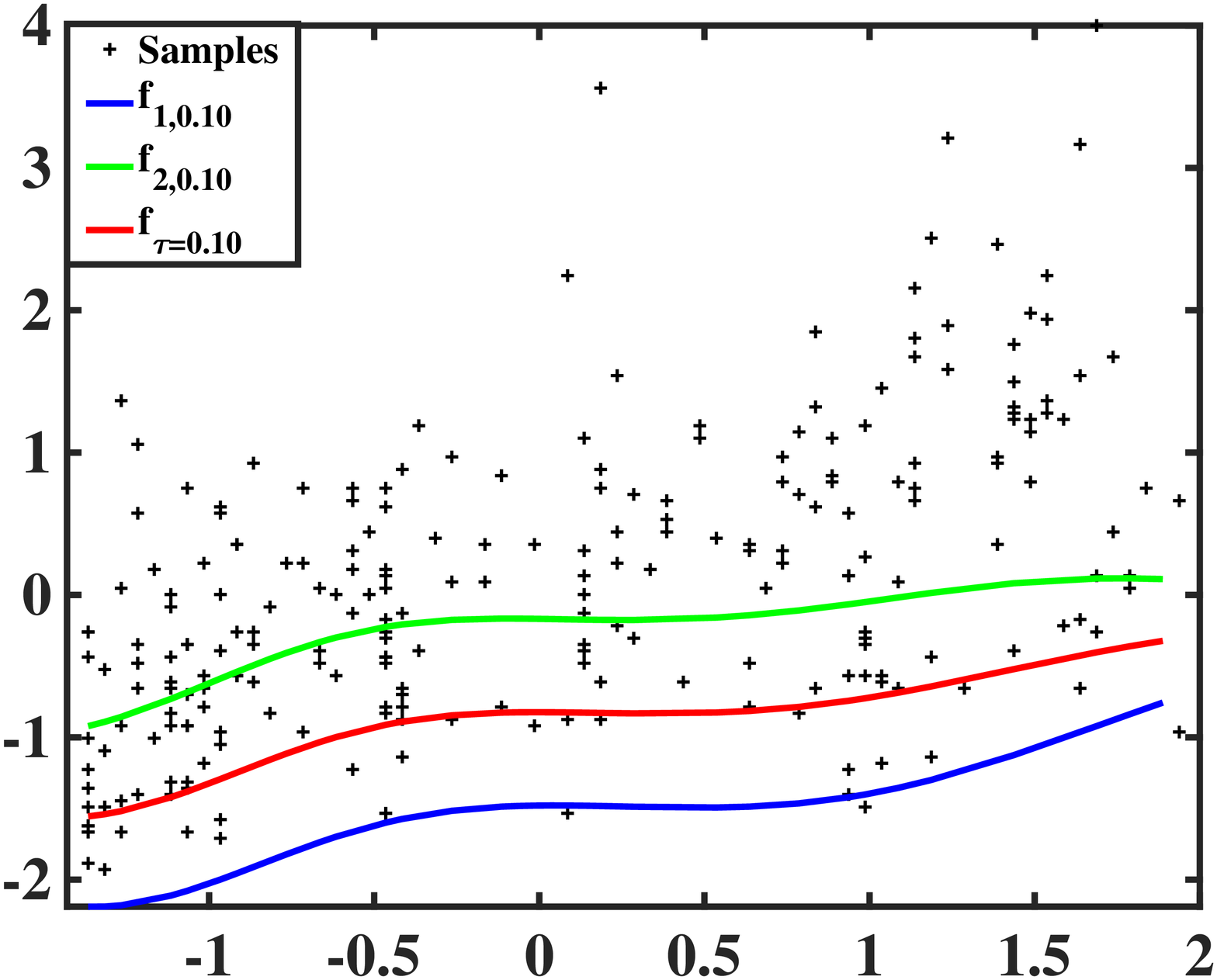}}
\subfigure[$\tau=0.25$]{\includegraphics[width=0.35\textheight]{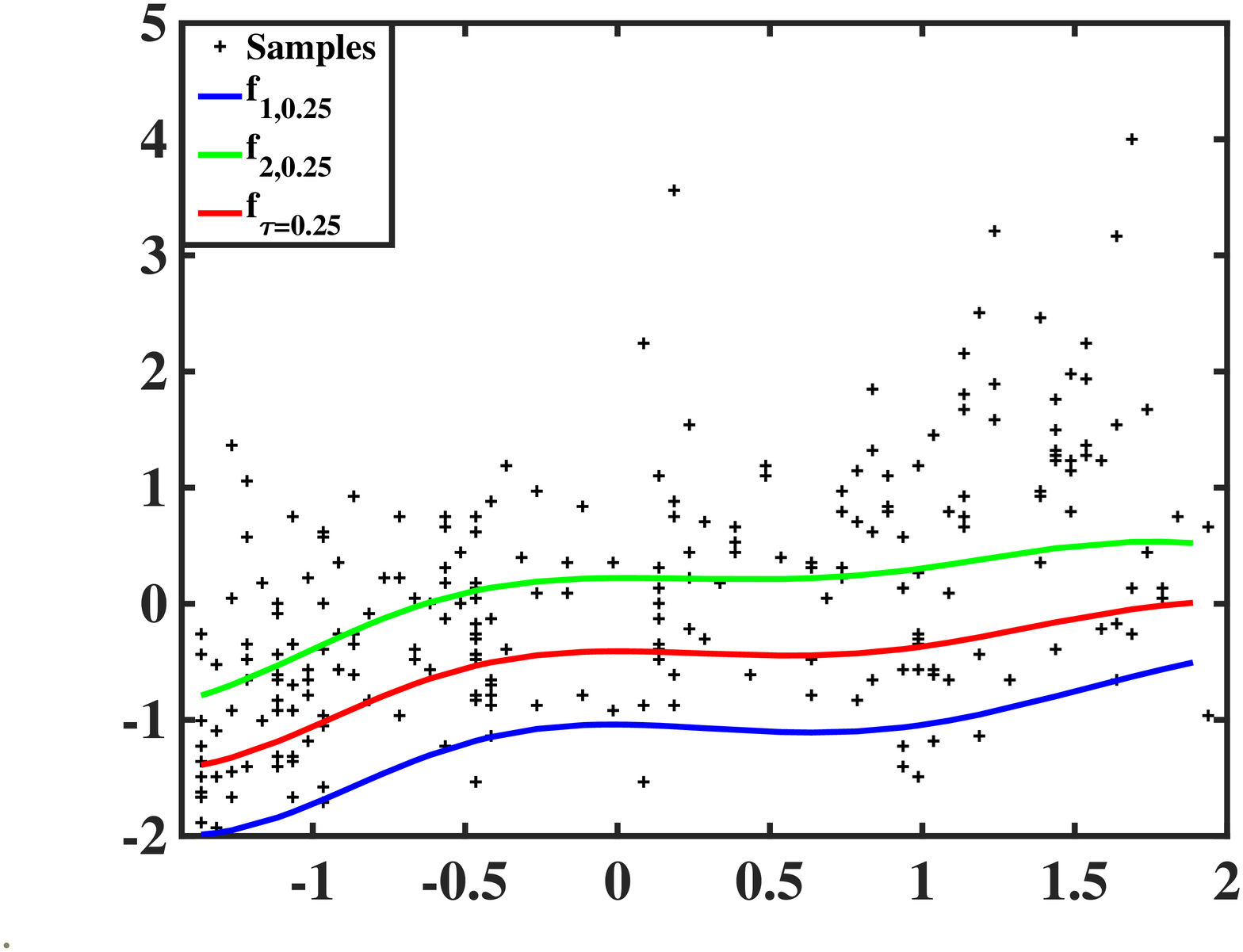}}
\subfigure[$\tau=0.50$]{\includegraphics[width=0.35\textheight]{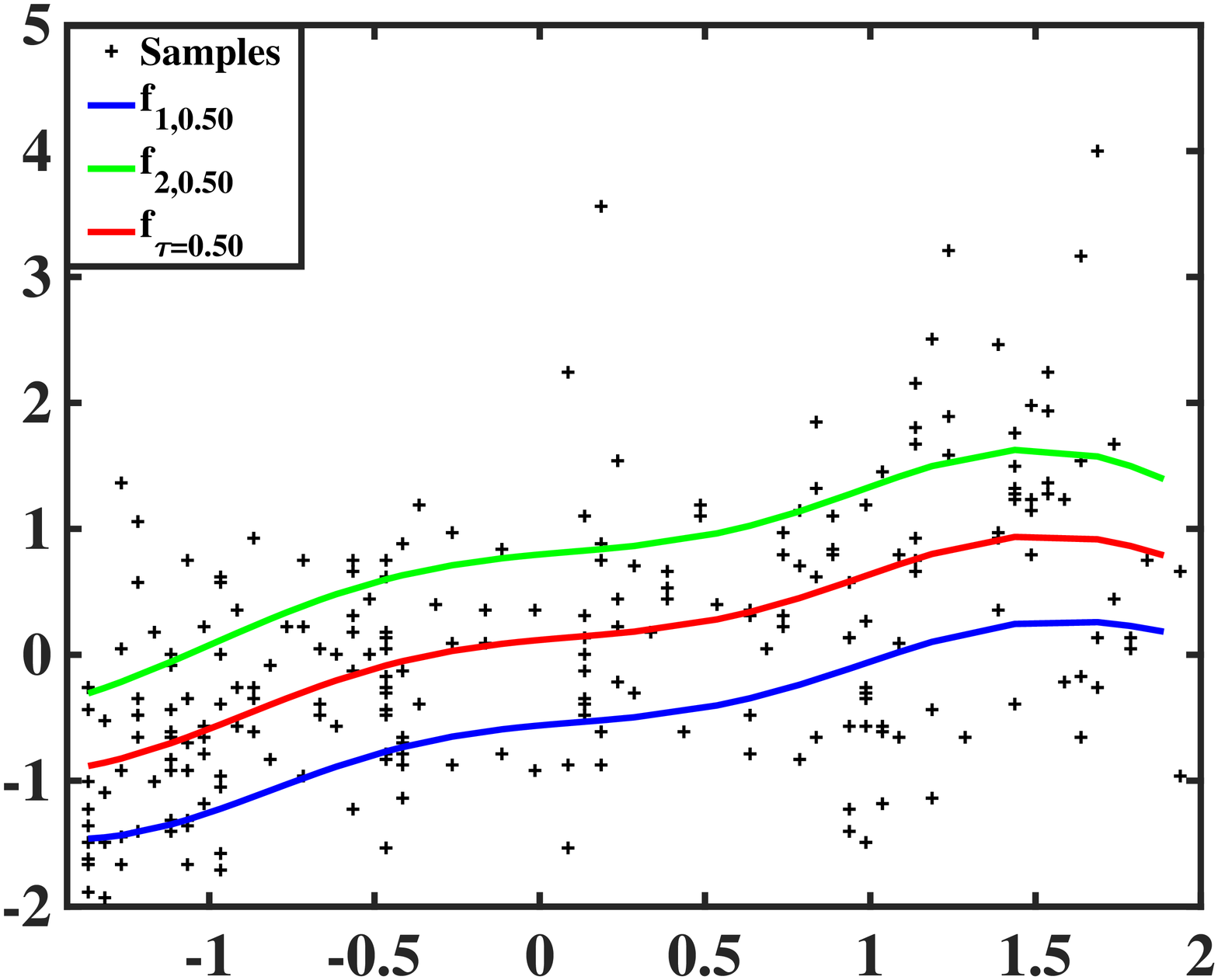}}
\subfigure[$\tau=0.75$]{\includegraphics[width=0.35\textheight]{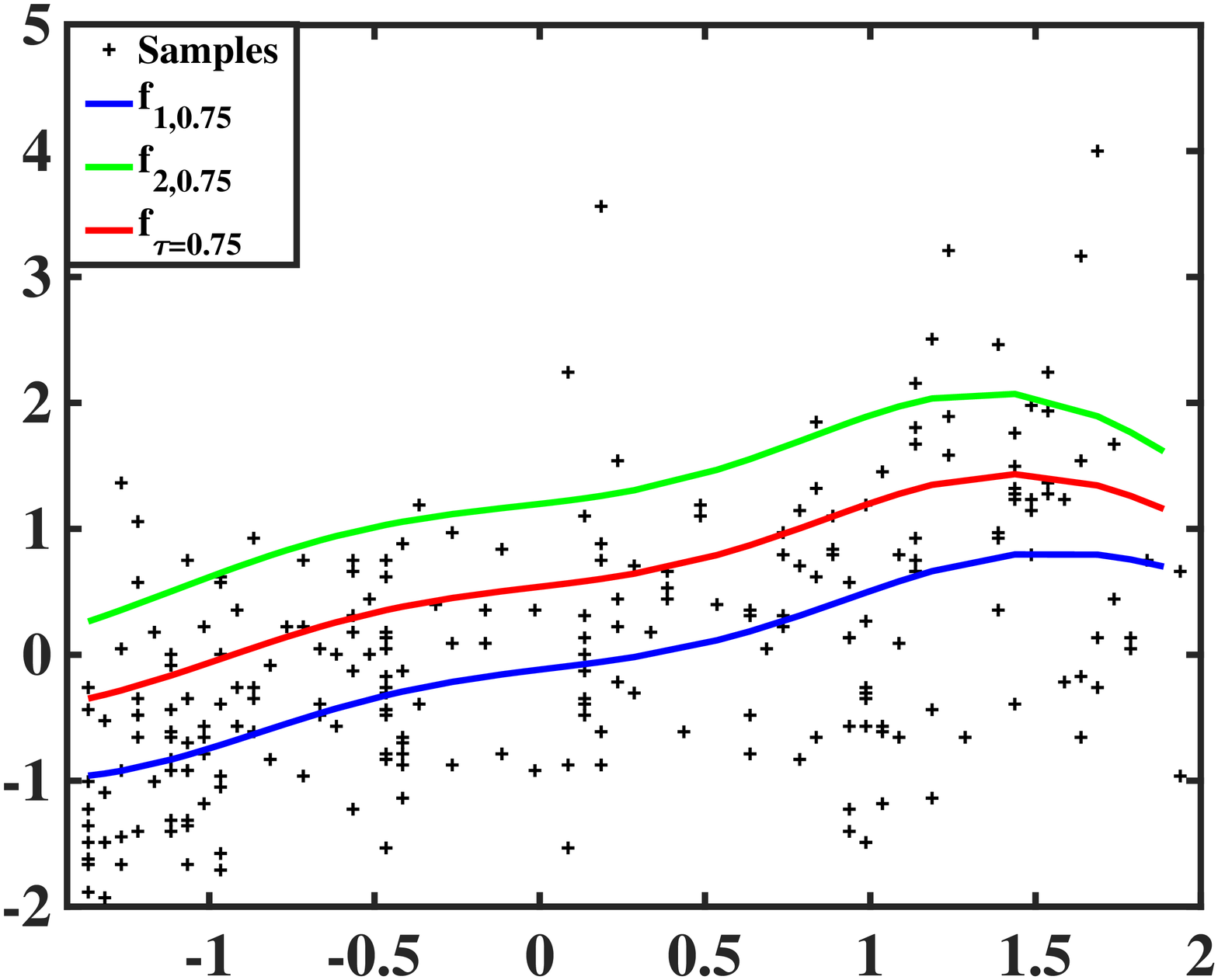}}
\subfigure[$\tau=0.90$]{\includegraphics[width=0.35\textheight]{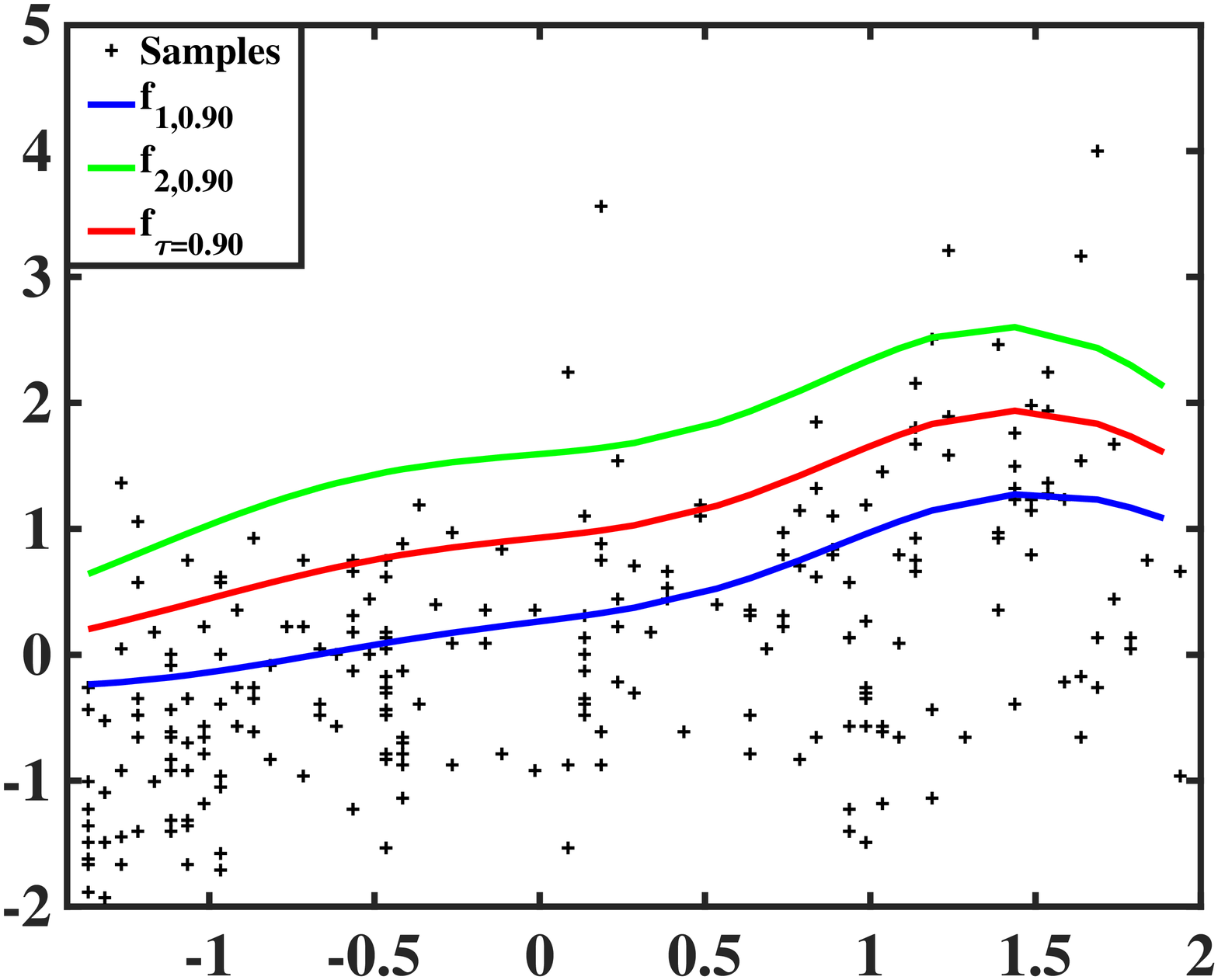}}
\caption{Regression results of TSVQR ($\tau=0.10, 0.25, 0.50, 0.75,$ and $0.90$).}\label{igm}
\end{figure}

Fig. \ref{igm} presents the regression results of the lower-bound function $f_{1}(x)$,
the upper-bound function $f_{2}(x)$, and the final decision function $f(x)$, when $\tau=0.10, 0.25, 0.50, 0.75,$ and $0.90$. As seen from Fig. \ref{igm},
we see that at each quantile level, $f_{1}(x)$ depicts the lower-bound distribution information,
$f_{2}(x)$ depicts the upper-bound distribution information, and $f_{1}(x)$ and $f_{2}(x)$ are nonparallel.
$f(x)$ combines the distribution information of $f_{1}(x)$ and $f_{2}(x)$, and then presents the information at each quantile location.
As the quantile parameter $\tau$ ranges from $0.1$ to $0.9$, we plot the corresponding final decision function $f(x)$ represented by the red solid curves in Fig. \ref{igm}.
Obviously, different quantile locations yield different red solid curves and these curves are nonparallel, indicating that the immunoglobulin G Data reflect potential heterogeneity and asymmetry.
Therefore, TSVQR completely captures the heterogeneous and asymmetric information at low and high quantiles of all data points.

\section{Applications of TSVQR}\label{Sec:5}
To validate the regression performance of TSVQR, we provide experiments on several real-world data sets, including large scale data, time series, and imbalanced data.
\subsection{Large scale data}
Heterogeneity and asymmetry have simultaneously emerged in environment field. We apply TSVQR to predicting the energy use of appliances (\citeauthor{candanedo2017data},\citeyear{candanedo2017data}) and assessing PM2.5 pollution in Beijing (\citeauthor{liang2015assessing}, \citeyear{liang2015assessing}). Appliances energy prediction data set involves 19735 instances and 29 attributes. Beijing PM2.5 data set contains 43824 instances and 13 attributes. We use the first half of each data set for training, and the rest for testing. For the nonlinear TSVQR, we employ a Gaussian kernel, and a wavelet kernel, respectively. All experiments are implemented
in the MATLAB R2021a environment on the Ubuntu 20.05 and 128 GB of RAM.

Tables \ref{energy} and \ref{PM2.5} list the regression results for TSVQR, when $\tau=0.10$, $0.25$, $0.50$, $0.75,$ and $0.90$. From both tables, we find that the values of Risk, RMSE, and MAE fluctuate slightly at each quantile level, which means that TSVQR fits the actual data set at every quantile location. Moreover, the regression results of both nonlinear kernel are similar, which is evident that TSVQR effectively captures the heterogeneous and asymmetric information in all data points. As for the training speed, the learning speed of TSVQR fluctuates slightly at each quantile level in Table \ref{PM2.5}. Therefore, TSVQR receives good regression performance with fast training speed for big data sets.

\begin{table}
	\footnotesize
	\begin{center}
		\caption{Evaluation indices of appliance energy prediction data set.}
		\label{energy}
		\begin{tabular}{@{}cccccc@{}}
			\hline
			$\tau$&0.10&0.25&0.50&0.75&0.90\\
			\hline
			Gaussian kernel &&&&& \\
			Risk&0.2774&0.2880&0.2918&0.2928&0.2921\\
			RMSE&0.9154&0.9551&0.9718&0.9799&0.9943\\
			MAE&0.5711&0.5803&0.5835&0.5847&0.5871\\
			CPU Time&51.8825&52.8494&84.0566&129.2691&142.7396\\
			\hline
			Wavelet kernel&&&&& \\
			Risk&0.2953&0.2952&0.2952&0.2949&0.2948\\
			RMSE&0.9966&0.9967&0.9967&0.9966&0.9967\\
			MAE&0.5901&0.5901&0.5901&0.5901&0.5901\\
			CPU Time&95.4110&59.2504&43.0104&76.8283&75.5245\\
			\hline
		\end{tabular}
	\end{center}
\end{table}

\begin{table}
	\footnotesize
	\begin{center}
		\caption{Evaluation indices of Beijing PM2.5 data sets}
		\label{PM2.5}
		\begin{tabular}{@{}cccccc@{}}
			\hline
			$\tau$&0.10&0.25&0.50&0.75&0.90\\
			\hline
			Gaussian kernel &&&&& \\
			Risk&0.1545&0.3369&0.5390&0.6605&0.7141\\
			RMSE&1.7663&1.5790&1.3864&1.3976&1.4737\\
			MAE&1.4656&1.3054&1.0780&0.9735&1.0461\\
			CPU Time&779.8702&728.8788&732.0485&728.1161&776.8404\\
	        \hline
            Wavelet kernel &&&&& \\
			Risk&0.2179&0.3682&0.5455&0.6655&0.7678\\
			RMSE&1.7392&1.5597&1.4250&1.5307&1.7420\\
			MAE&1.4357&1.2812&1.0910&1.1028&1.3615\\
			CPU Time&755.6335&671.0814&653.9155&653.5467&702.8267\\
			\hline
		\end{tabular}
	\end{center}
\end{table}

\subsection{Time series}
We apply TSVQR to ``MelTemp" time series and ``Gasprice" time series, which are available in R package. The ``MelTemp" data set illustrates daily maximum temperatures in Melbourne, involving $3650$ instances. The ``Gasprice" data set presents weekly US gasoline price, including 695 instances. We use the forward chaining method to validate the regression performance of TSVQR. Table \ref{timeseries} lists the regression results for TSVQR, when $\tau=0.10$, $0.25$, $0.50$, $0.75,$ and $0.90$. From Table \ref{timeseries}, we find that the values of Risk, RMSE, and MAE fluctuate slightly at each quantile level, which means that TSVQR fits the time series at every quantile location.  As for the training speed, the learning speed of TSVQR fluctuates slightly at each quantile level in Table \ref{timeseries}.

\begin{table}
	\footnotesize
	\begin{center}
		\caption{Evaluation indices of time series data sets.}
		\label{timeseries}
		\begin{tabular}{@{}cccccc@{}}
			\hline
			$\tau$&0.10&0.25&0.50&0.75&0.90\\
			\hline
			Gasprice&&&&& \\
			Risk&0.0006&0.0013&0.0019&0.0017&0.0008\\
			RMSE&0.0202&0.0248&0.0299&0.0284&0.0226\\
			MAE&0.0026&0.0031&0.0038&0.0036&0.0029\\
			CPU Time&0.0272&0.0232&0.0266&0.0332&0.0256\\
			\hline
			MelTemp&&&&& \\
			Risk&0.0002&0.0003&0.0004&0.0003&0.0001\\
			RMSE&0.0039&0.0089&0.0151&0.0190&0.0202\\
			MAE&0.0002&0.0005&0.0008&0.0010&0.0011\\
			CPU Time&0.9883&0.9836&1.0954&1.0476&0.9943\\
			\hline
		\end{tabular}
	\end{center}
\end{table}

\subsection{Imbalanced data}
To validate the asymmetric-information-capturing ability of TSVQR, we consider two imbalanced data sets: ``Dermatology" and ``Car Evaluation".
The ``Dermatology" data set contains $366$ instances and $34$ attributes. The ``Car Evaluation" data set contains $1728$ instances and $6$ attributes. Both data sets, where the total number of samples is not the same in the classes (\citeauthor{gupta2019fuzzy}, \citeyear{gupta2019fuzzy}), are very popular in the UC Irvine (UCI) Machine Learning Repository. We apply Monte Carlo cross validation (MCCV) with $1000$ iterations using the different ratios of $70\%:30\%$ and $60\%:40\%$. Tables \ref{Dermatology} and \ref{Car Evaluation} show the average results of Risk, RMSE, MAE, and CPU time, when $\tau=0.10$, $0.25$, $0.50$, $0.75,$ and $0.90$. From both data sets, we find that TSVQR gets small Risk at each quantile level, indicating that TSVQR captures the asymmetric information at different quantile location. Moreover, the average training speed of TSVQR fluctuates slightly at different quantile levels. As the number of training samples in the same data set increases, the average training speed decreases.
\begin{table}
	\footnotesize
	\begin{center}
		\caption{Evaluation indices of Dermatology data sets.}
		\label{Dermatology}
		\begin{tabular}{@{}lccccc@{}}
			\hline
			$\tau$&0.10&0.25&0.50&0.75&0.90\\
			\hline
			Ratio $70\%:30\%$ &&&&& \\
			Risk     &  0.1050  &  0.1206   & 0.1198   & 0.1094   & 0.0921 \\
			RMSE      & 0.5219  &  0.3860   & 0.3470   & 0.3548   & 0.4125  \\
			MAE      & 0.3418   & 0.2649    & 0.2397   & 0.2450    & 0.2937  \\
			CPU Time & 0.0141   & 0.0140    & 0.0139   &  0.0139   & 0.0141   \\
	        \hline
            Ratio $60\%:40\%$ &&&&& \\
			Risk   & 0.1103   & 0.1273   & 0.1261   & 0.1163   & 0.0993   \\
			RMSE  &  0.5482   & 0.4160   & 0.3649   & 0.3715   & 0.4317   \\
			MAE   & 0.3589   & 0.2823   & 0.2521    & 0.2577   & 0.3101    \\
			CPU Time & 0.0147    &0.0146    & 0.0147    &0.0147   & 0.0148 \\
			\hline
		\end{tabular}
	\end{center}
\end{table}

\begin{table}
	\footnotesize
	\begin{center}
		\caption{Evaluation indices of Car Evaluation data sets.}
		\label{Car Evaluation}
		\begin{tabular}{@{}lccccc@{}}
			\hline
			$\tau$&0.10&0.25&0.50&0.75&0.90\\
			\hline
			Ratio $70\%:30\%$ &&&&& \\
			Risk   & 0.0697  &  0.0924   & 0.1010    &0.0916   & 0.0768   \\
			RMSE   &  0.6413  &  0.4679  &  0.3534    &0.3291  &  0.3510   \\
			MAE    &  0.3027  &  0.2356  &  0.2021   & 0.1966   & 0.2250  \\
			CPU Time    & 0.1990   & 0.1982   & 0.1996  &  0.1982  &  0.2130\\
	        \hline
            Ratio $60\%:40\%$ &&&&& \\
			Risk   &0.0797    &0.1031   & 0.1108   & 0.1015  &  0.0871  \\
			RMSE   &0.6657   & 0.5035   & 0.3802   & 0.3505   & 0.3715  \\
			MAE    & 0.3247  &  0.2585   & 0.2217   & 0.2155    & 0.2479   \\
			CPU Time & 0.1449  &  0.1432  &  0.1441  &  0.1457 &   0.1547 \\
			\hline
		\end{tabular}
	\end{center}
\end{table}

\section{Conclusions}\label{Sec:6}
Heterogeneity and asymmetry are two major and common statistical features of modern data. How
to capture the heterogeneous and asymmetric information in data is a major challenge for regression technology. This paper proposed a twin support vector quantile regression to efficiently solve this problem.
Experimental results on both artificial data sets and real-world data sets demonstrated that
TSVQR effectively depicted the heterogeneous and asymmetric information and gave a more complete picture of the data set. The training speed
of TSVQR was significantly faster than that of SVQR, $\varepsilon$-SVQR, and Online-SVQR.

Compared with SVQR, $\varepsilon$-SVQR, Online-SVQR, and GPQR, the main
strengths of TSVQR are as follows. First, the quantile parameter is
adopted in TSVQR to completely measure the heterogeneous information in data points. Second,
TSVQR constructed two smaller-sized quadratic programming problems to obtain two nonparallel hyperplanes to depict
the asymmetric information at each quantile location. Third, two smaller-sized QPPs made TSVQR work quickly.
The dual coordinate descent algorithm to solve the QPPs accelerated the training speed of TSVQR.

Despite several contributions of this study, there are some limitations that should be taken into account
in future research. First, how to theoretically determine the optimal parameters of TSVQR for each quantile location is still a question.
Second, we construct more reasonable criteria to evaluate the regression results of TSVQR, as some traditional evaluation criteria
are only suitable for the conditional mean models.
Third, from the application perspective,
how to use the proposed method to deal with the heterogeneous problem or asymmetric problem
in the real world remains an open question. For example, heterogeneity and asymmetry have commonly emerged in medicine (\citeauthor{ncd2021heterogeneous}, \citeyear{ncd2021heterogeneous}) and environment (\citeauthor{ahmad2018empirics}, \citeyear{ahmad2018empirics}; \citeauthor{cheng2021does}, \citeyear{cheng2021does}). The application of TSVQR to both fields especially for big data, time series, or image data is our future work.

\section*{Data availability}
The code and artificial data that support the findings of this study are available in ``http://www.optimal-group.org/Resources/Code/TSVQR.html", the UCI benchmark data that support the findings of this study are available in ``http://archive.ics.uci.edu/ml/index.
php", and data sets of ``Engel", ``Bone density", ``US girls", ``MelTemp", and ``Gasprice" are available in ``https://www.r-project.org".

\section*{Acknowledgments}
This work is supported by the National Natural Science Foundation of
China (Nos. 12101552, 11871183, and 61866010),
the National Social Science Foundation of
China (No. 21BJY256), Philosophy and Social Sciences Leading Talent Training Project of Zhejiang Province (No. 21YJRC07-1YB),
the Natural Science Foundation of Zhejiang Province (No. LY21F030013), and the Natural Science Foundation of Hainan Province (No. 120RC449).

\bibliographystyle{model5-names}\biboptions{authoryear}

\bibliography{myreference}

\end{document}